%% file: PSSNet.tex
\pdfoutput=1
\documentclass{elsarticle}
\usepackage{lineno,hyperref}
\usepackage[font=footnotesize,labelfont=bf]{caption}
\usepackage[font=footnotesize,labelfont=bf]{subcaption}
\usepackage{amsmath}
\usepackage{float}
\usepackage{booktabs}
\usepackage{changepage}
\usepackage{xcolor}
\usepackage{adjustbox,lipsum}
\usepackage{multirow}
\usepackage{array}
\newcolumntype{L}[1]{>{\raggedright\let\newline\\\arraybackslash\hspace{0pt}}m{#1}}
\newcolumntype{C}[1]{>{\centering\let\newline\\\arraybackslash\hspace{0pt}}m{#1}}
\newcolumntype{R}[1]{>{\raggedleft\let\newline\\\arraybackslash\hspace{0pt}}m{#1}}
\usepackage[flushleft]{threeparttable} 
\usepackage{graphicx}
\usepackage{rotating}

\usepackage[capitalize]{cleveref}
\crefname{section}{Sec.}{Secs.}
\Crefname{section}{Section}{Sections}
\Crefname{table}{Table}{Tables}
\crefname{table}{Tab.}{Tabs.}

\usepackage{hyperref}
\usepackage{color} 
\usepackage{lineno} 
\usepackage{bbm} 
\usepackage{enumitem} 
\usepackage{adjustbox} 
\usepackage{multirow}
\usepackage[labelformat=simple]{subcaption} 

\graphicspath{{figures/pipeline/}{figures/pipeline/}{figures/ambigious_boundary/}{figures/appendix/}{figures/final_results/}{figures/graphs_2D/}{figures/pipeline/}{figures/plots/}{figures/segments/}{figures/segments/mrfpnp_config/}{figures/segments/row_same_op/}{figures/segments/row_same_seg_num/}{figures/observations/}}
\journal{ISPRS Journal of Photogrammetry and Remote Sensing}

\begin{document}\sloppy

\newpage

	\begin{frontmatter}
		
		\title{PSSNet: Planarity-sensible Semantic Segmentation of Large-scale Urban Meshes}

		\author[TUDelftaddress]{Weixiao GAO\corref{mycorrespondingauthor}}
		\cortext[mycorrespondingauthor]{Corresponding author}
		\ead{w.gao-1@tudelft.nl}
		
		\author[TUDelftaddress]{Liangliang Nan}
		\ead{liangliang.nan@tudelft.nl}
		
		\author[CMTaddress]{Bas Boom}
		\ead{bas.boom12@gmail.com}
		
		\author[TUDelftaddress]{Hugo Ledoux}
		\ead{h.ledoux@tudelft.nl}
		
		\address[TUDelftaddress]{3D Geoinformation Research Group, Faculty of Architecture and the Built Environment, Delft University of Technology, 2628 BL Delft, The Netherlands}
		\address[CMTaddress]{CycloMedia Technology, Zaltbommel, The Netherlands}
			
		\begin{abstract}
		We introduce a novel deep learning-based framework to interpret 3D urban scenes represented as textured meshes.
		Based on the observation that object boundaries typically align with the boundaries of planar regions, our framework achieves semantic segmentation in two steps: planarity-sensible over-segmentation followed by semantic classification.
		The over-segmentation step generates an initial set of mesh segments that capture the planar and non-planar regions of urban scenes.
		In the subsequent classification step, we construct a  graph that encodes the geometric and photometric features of the segments in its nodes and the multi-scale contextual features in its edges. 
		The final semantic segmentation is obtained by classifying the segments using a graph convolutional network.
		Experiments and comparisons on two semantic urban mesh benchmarks demonstrate that our approach outperforms the state-of-the-art methods in terms of boundary quality, mean IoU (intersection over union), and generalization ability.  
		We also introduce several new metrics for evaluating mesh over-segmentation methods dedicated to semantic segmentation, and our proposed over-segmentation approach outperforms state-of-the-art methods on all metrics. 
		Our source code is available at \hyperlink{https://github.com/WeixiaoGao/PSSNet}{https://github.com/WeixiaoGao/PSSNet}.
		\end{abstract}
	\end{frontmatter}

	\input{./source/introduction}

	\input{./source/related_work}
	
	\input{./source/methodology}

	\input{./source/experiments}

	\input{./source/conclusion}

	\bibliography{egbib}
	
\end{document}

%% file: source/introduction.tex
\section{Introduction}%
\label{sec:intro}

Recent advances in photogrammetry and 3D computer vision have enabled the generation of textured meshes of large-scale urban scenes that contain buildings, trees, vehicles, etc.~\cite{Helsinki3d,Google3d,gao2021sum}.
Deriving semantic information from the mesh models is critical to allowing the use of these meshes in diverse applications, e.g., energy estimate, noise modeling, and solar potential~\cite{biljecki2015applications,saran2015citygml,besuievsky2018skyline}.

\begin{figure*} [th]
	\centering
	\begin{subfigure}{0.32\linewidth}
		\includegraphics[width=\linewidth]{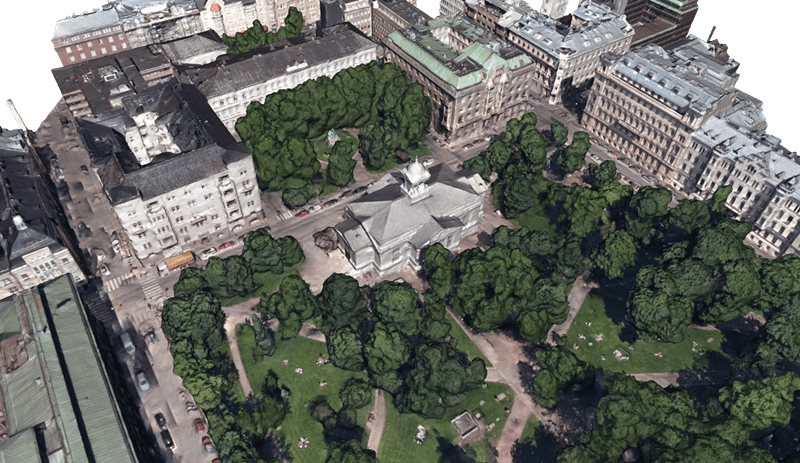}
		\caption{Input mesh}
		\label{fig:texture_meshes}
	\end{subfigure}
	\hfill
	\begin{subfigure}{0.32\linewidth}
		\includegraphics[width=\linewidth]{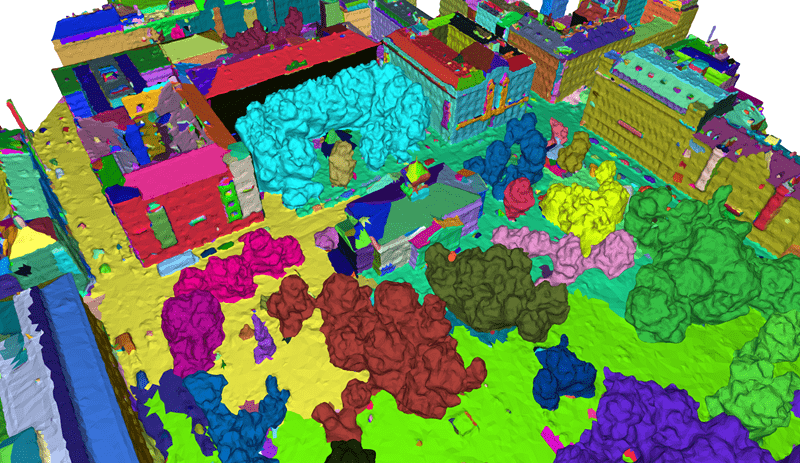}
		\caption{Over-segmented mesh}
		\label{fig:adaptive_segments}
	\end{subfigure}
	\hfill
	\begin{subfigure}{0.32\linewidth}
		\includegraphics[width=\linewidth]{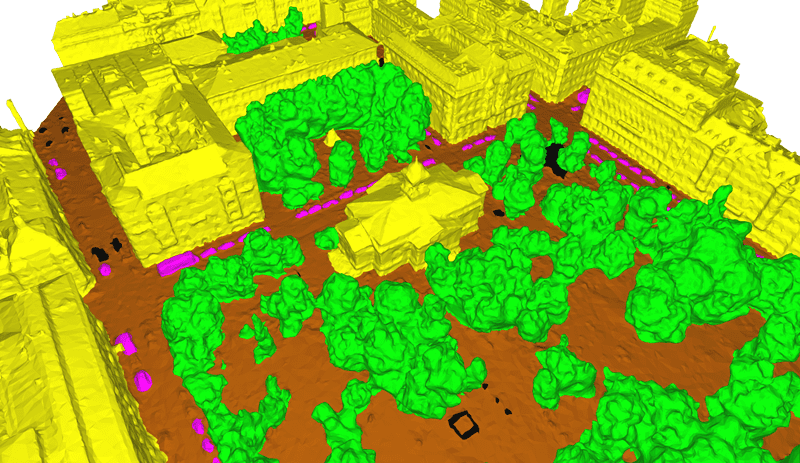}
		\caption{Semantic mesh}
		\label{fig:semantic_meshes}
	\end{subfigure}
	\caption{The workflow of our method. We first decompose the input mesh (a) into a set of \textit{planar} and \textit{non-planar} segments (b). Then we classify the segments using graph convolutional networks to obtain the results of semantic segmentation (c). In (b), the segments are randomly colorized. In (c), the colors are:
		\includegraphics[width=0.02\textwidth]{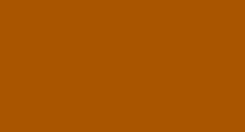} terrain,
		\includegraphics[width=0.02\textwidth]{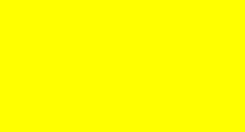} building,
		\includegraphics[width=0.02\textwidth]{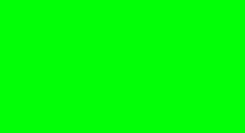} high vegetation,
		\includegraphics[width=0.02\textwidth]{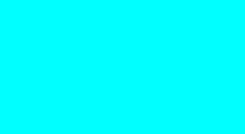} water,
		\includegraphics[width=0.02\textwidth]{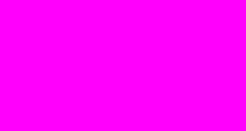} vehicle,
		\includegraphics[width=0.02\textwidth]{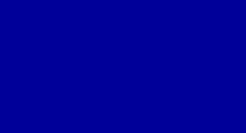} boat. 
	}%
	\label{fig:framework}
\end{figure*}

There exists a large volume of machine learning-based algorithms for the semantic segmentation of 3D data, and they are designed mainly for 3D point clouds~\cite{demantke2011dimensionality,hackel2016fast,qi2017pointnet,qi2017pointnet++,thomas2018semantic,thomas2019kpconv}.
A few recent works also address deep learning for surface meshes~\cite{hanocka2019meshcnn,gao2019sdm,selvaraju2021buildingnet,fu20213d} but are limited to individual objects or small indoor scenes (e.g., living room, kitchen).
Unlike point clouds that are usually obtained as the raw input from typical data acquisition devices, textured meshes (see~\Cref{fig:texture_meshes}) provide topological information, have continuous surfaces, yield better visualization, and are lightweight, which makes them an ideal representation for urban scenes.
Surprisingly, the semantic segmentation of urban meshes has rarely been investigated, ~\cite{verdie2015lod,rouhani2017semantic,gao2021sum} are exceptions. 

\begin{figure*} [th]
	\centering
	\begin{subfigure}{0.32\linewidth}
		\includegraphics[width=\linewidth]{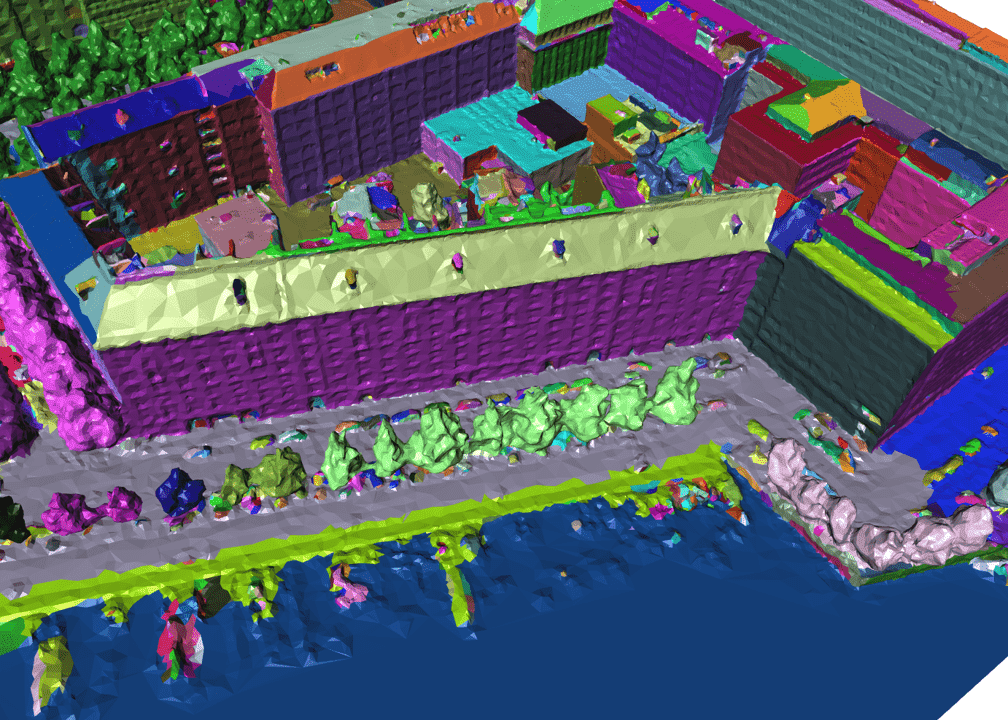}
	\end{subfigure}
	\hfill
	\begin{subfigure}{0.32\linewidth}
		\includegraphics[width=\linewidth]{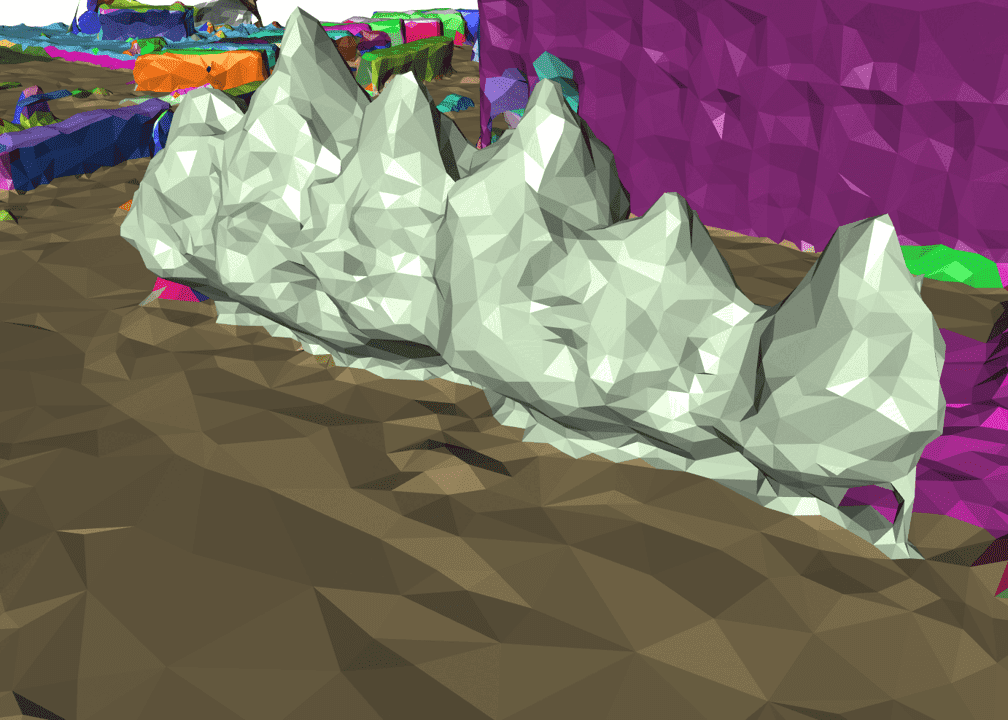}
	\end{subfigure}
	\hfill
	\begin{subfigure}{0.32\linewidth}
		\includegraphics[width=\linewidth]{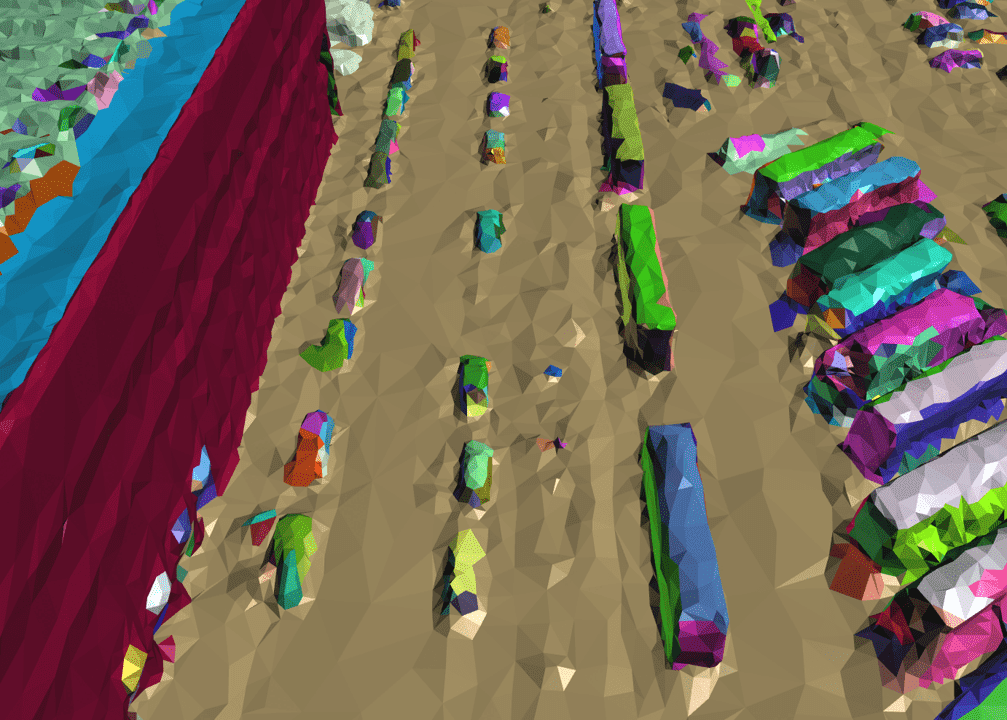}
	\end{subfigure}
	\begin{subfigure}{0.32\linewidth}
		\includegraphics[width=\linewidth]{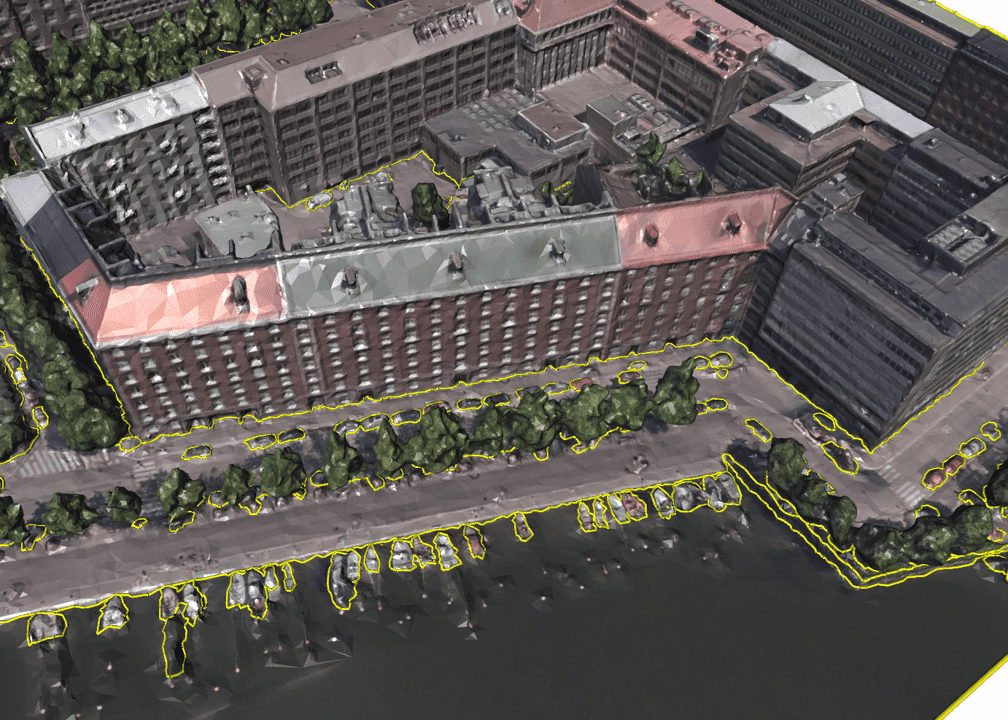}
	\end{subfigure}
	\hfill
	\begin{subfigure}{0.32\linewidth}
		\includegraphics[width=\linewidth]{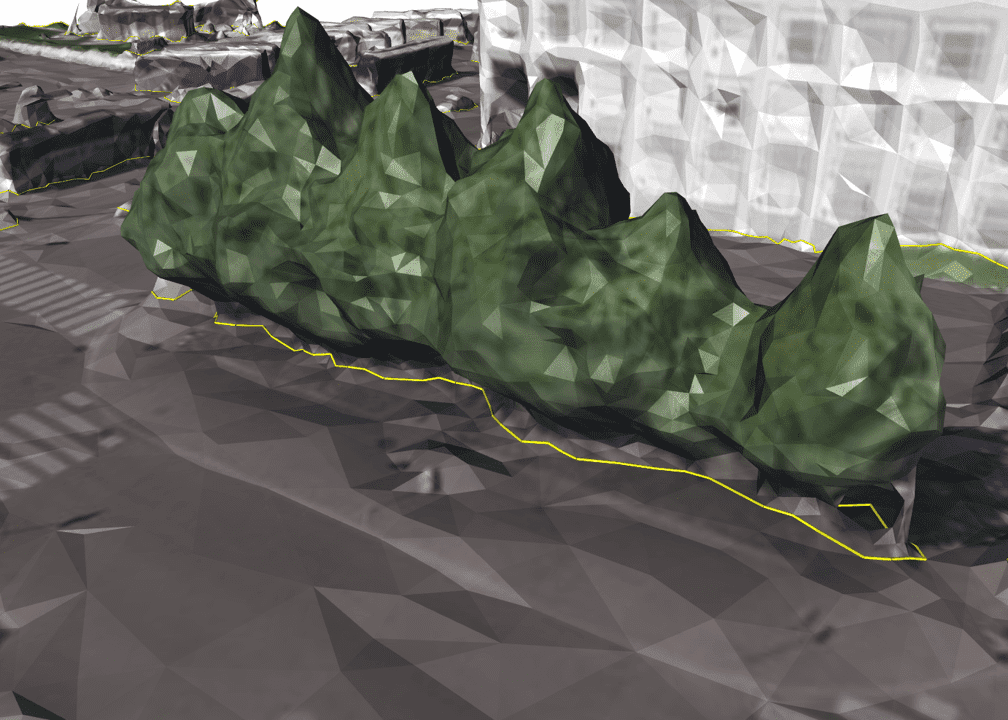}
	\end{subfigure}
	\hfill
	\begin{subfigure}{0.32\linewidth}
		\includegraphics[width=\linewidth]{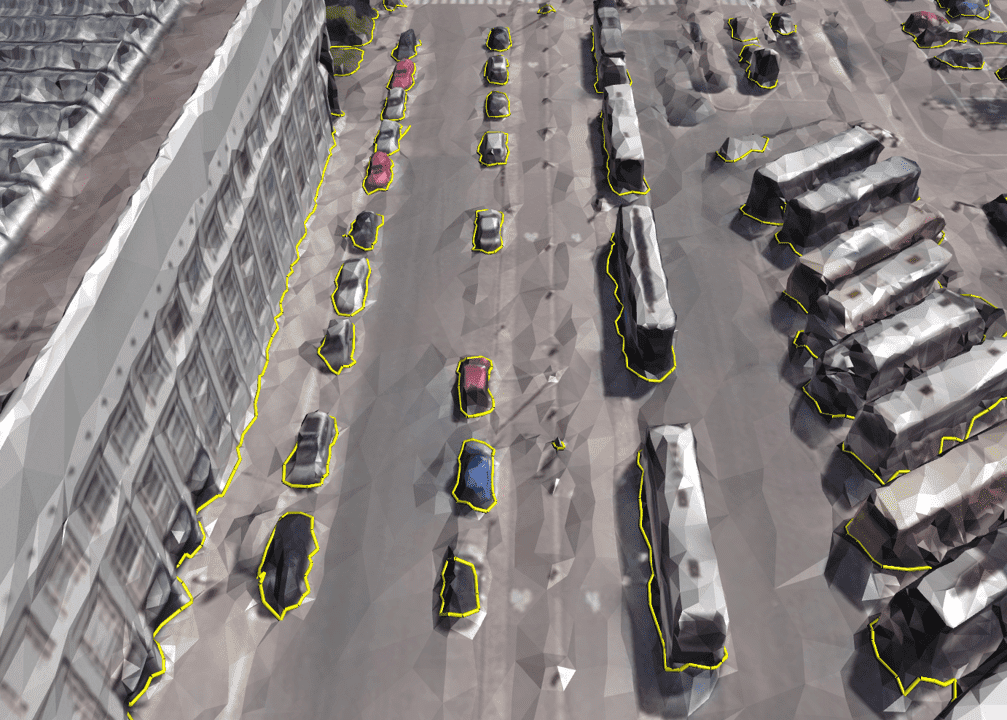}
	\end{subfigure}
	\caption{Object boundaries often align with the boundaries of planar regions. Top: planar segmentation results; Bottom: the corresponding groundtruth object boundaries (shown as yellow lines).}%
	\label{fig:observations}
\end{figure*}

In this work, we address the semantic segmentation of urban meshes by introducing a two-step framework using deep learning. 
Our framework is designed to improve the following three aspects of semantic segmentation:

\noindent \textbf{1) Segmentation quality}.
Urban scenes typically contain piecewise regions, which can already inspire the separation of man-made objects (e.g., roads, buildings) from organic objects (e.g., trees).
We observe that semantic segmentation algorithms usually perform well in the interior of large smooth surfaces (including planar surfaces), but that they perform poorly for the identification of object boundaries. 
Given the fact that object boundaries typically align with the boundaries of planar regions (see~\Cref{fig:observations}), our framework achieves semantic segmentation by first exploiting a planarity-sensible over-segmentation step that separates \textit{planar} and \textit{non-planar} surface patches.

\noindent \textbf{2) Descriptiveness of geometric features}. Existing methods for semantic segmentation of 3D data commonly rely on features defined on local primitives (i.e., points or triangles)~\cite{weinmann2013feature, weinmann2015contextual,qi2017pointnet, huang2019texturenet, li2019cross, schult2020dualconvmesh} or segments (i.e., a group of points or triangles)~\cite{lin2018toward,landrieu2018large, cohen2004variational,lafarge2012creating,verdie2015lod,rouhani2017semantic}. 
Features from local primitives are limited to a certain distance in the local neighborhood, while features used in existing segment-based approaches do not effectively capture the contextual relationships between segments. Thus, they are less descriptive in representing the complex shapes of diverse objects and in revealing the relationships between objects.
In our work, by initially decomposing a mesh model into \textit{planar} and \textit{non-planar} segments, both local geometric features of individual segments and global relationships between segments can be captured.

\noindent \textbf{3) Efficiency}. 
Existing deep learning-based methods for processing 3D data are limited by the data size, especially for large-scale urban scenes. 
This has already motivated over-segmentation for semantic segmentation~\cite{weinmann2015contextual,landrieu2018large,landrieu2019point,Hui_2021_ICCV}. 
Following the spirit of the previous work for improving efficiency, our over-segmentation facilitates better object boundaries and strengthens semantic segmentation by distinctive local and non-local features, which is suitable for the subsequent classification using graph convolutional networks (GCN).

Besides the two-step semantic segmentation framework, we also introduce several new metrics for evaluating mesh over-segmentation techniques. We believe the proposed metrics will further stimulate the improvement of over-segmentation for semantic segmentation.

Experiments on two benchmarks show that our approach outperforms recently developed methods in terms of boundary quality, mean IoU (intersection over union), and generalization ability.

In summary, our contributions are: 1) a novel mesh over-segmentation approach for extracting planarity-sensible segments that are dedicated for GCN-based semantic segmentation; 2) a new graph structure that encodes both local geometric and photometric features of segments, as well as global spatial relationships between segments; 3) several novel metrics for evaluating mesh over-segmentation techniques in the context of semantic segmentation.

%% file: source/related_work.tex
\section{Related Work}
\label{sec:relaw}

While there is a large volume of research on the over-segmentation and semantic segmentation of urban images~\cite{cordts2016cityscapes,yang2018denseaspp}, we focus in this sole section on methods designed to process large-scale 3D data, i.e., point clouds and meshes of urban scenes. Methods specially designed for handling individual objects or small scenes~\cite{nan2012search,hanocka2019meshcnn,gao2019sdm,selvaraju2021buildingnet,fu20213d} usually do not scale to large-scale urban scenes and thus are not covered.

\subsection{Over-segmentation of 3D data} 

Many methods for over-segmentation of 3D data are inspired by image over-segmentation algorithms~\cite{liu2011entropy} and can be divided into four categories: (1) primitive-based fitting~\cite{vosselman2004recognising,schnabel2007efficient,lafarge2012creating}, (2) graph-based partitioning~\cite{landrieu2018large,ben2018graph}, (3) local region expansion~\cite{cohen2004variational,melzer2007non,lafarge2012creating,papon2013voxel,rouhani2017semantic,vosselman2017contextual,papon2013voxel,lin2018toward}, and (4) learning-based methods~\cite{landrieu2019point,Hui_2021_ICCV}.
Over-segmentation often serves as pre-processing for tasks such as semantic segmentation, instance segmentation, or reconstruction, and aims at reducing the complexity of subsequent tasks by using fewer segments having local homogeneity.
Due to the complexity of real-world scenes and the irregularity of the data, it is challenging to obtain over-segmentation results with a desired number of segments and clear object boundaries.
The aforementioned methods are either limited by the primitive types (e.g., plane, sphere, and cylinder) or suffer from severe under-segmentation errors when the number of segments is reduced or by the type of available labels in the training data (e.g., a few methods require instance labels~\cite{landrieu2019point,Hui_2021_ICCV}).
We propose to partition the input meshes into a relatively small number of homogeneous regions with clear object boundaries based on both geometric and photometric characteristics (see~\Cref{sec:methods}), which is beneficial to semantic segmentation (see~\Cref{sec:experiments}).

\subsection{Semantic segmentation of 3D data} 

An important step in semantic segmentation is feature extraction.
Based on the methods used for feature extraction, semantic segmentation approaches can be roughly categorized into three groups: handcrafted-feature-based~\cite{demantke2011dimensionality,weinmann2013feature,verdie2015lod,weinmann2015contextual,hackel2016fast,rouhani2017semantic,vosselman2017contextual,thomas2018semantic}, learning-based~\cite{qi2017pointnet,qi2017pointnet++,thomas2019kpconv,hu2020randla,lei2021picasso}, and hybrid methods~\cite{landrieu2018large,wu2021scenegraphfusion}.
Handcrafted features are often effective when with limited training data. 
In contrast, deep-learning techniques are more effective when sufficient training data is available~\cite{guo2020deep}.
These methods usually require contextual information to compute or learn features.
However, it is difficult to capture effective global contextual features.
Inspired by SPG~\cite{landrieu2018large}, our graph structure encodes various local geometric, photometric, and contextual features, and we apply a GCN for semantic segmentation.
Our method exploits enriched spatial relationships in the graph at both local and global scales, which greatly facilitates the GCN model to capture contextual information and learn distinctive features for semantic segmentation.   

%% file: source/methodology.tex
\section{Methodology}\label{sec:methods}

Our framework for semantic segmentation of urban meshes has two steps (as shown in~\Cref{fig:framework}):

\noindent \textbf{Planarity-sensible over-segmentation.} This step decomposes the urban mesh into a set of \textit{planar} and \textit{non-planar} surface patches because object boundaries often align with the boundaries of planar regions. 
This step not only enhances the descriptiveness of the features learned through local context but also significantly reduces the number of segments to be classified.

\noindent \textbf{Segmentation classification.} We construct a graph with its nodes encoding the local geometric and photometric features of the segments and its edges encoding global contextual features. We achieve semantic segmentation of the mesh by classifying the segments using a graph convolutional network.

\subsection{Planarity-sensible over-segmentation}\label{sec:geo_classify}
\begin{figure}[!tb]
	\centering
	\begin{subfigure}{0.49\linewidth}
		\centering
		\captionsetup{justification=centering}
		\includegraphics[page=2, width=\linewidth]{semantic_graph_new_simplified.pdf}
		\caption{Planar segments}
		\label{fig:2D_planar}
	\end{subfigure}
	\hfill
	\begin{subfigure}{0.49\linewidth}
		\centering
		\captionsetup{justification=centering}
		\includegraphics[page=1, width=\linewidth]{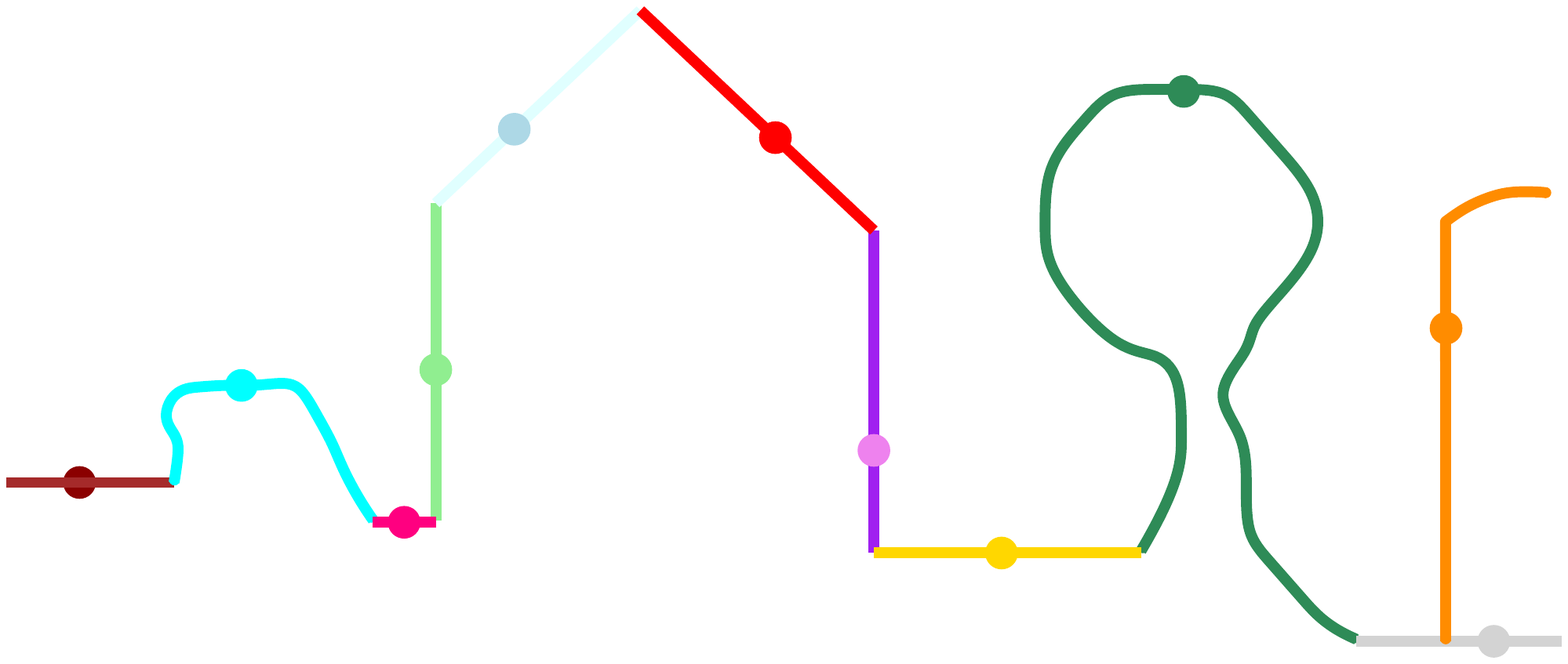}
		\caption{Planarity-sensible segments}
		\label{fig:2D_adaptive}
	\end{subfigure}
	\caption{2D illustrative comparison between planar and planarity-sensible segments. 
		Each dot and its line denote a segment.}
	\label{fig:planar_vs_adaptive}
\end{figure}

This step aims to decompose the urban mesh into a set of homogeneous segments in terms of geometric and photometric characteristics, see~\Cref{fig:planar_vs_adaptive}.
Compared with \textit{planar} segments generated by classical region growing methods~\cite{lafarge2012creating}, our segments can accommodate more complex surfaces (i.e., trees and vehicles).
Our over-segmentation, further detailed below, is achieved in two steps: (1) \textit{planar} and \textit{non-planar} classification, and (2) incremental segmentation.

\paragraph{Planar and non-planar classification} 
We classify the triangle faces of a mesh as either \textit{planar} or \textit{non-planar}. 
Following Gao et al.~\cite{gao2021sum}, we design a set of features including Eigen-based (i.e., \textit{linearity}, \textit{planarity}, \textit{sphericity}, \textit{curvature}, and \textit{verticality}), elevation-based (i.e., \textit{absolute}, \textit{relative}, and \textit{multi-scale}), scale-based (i.e, \textit{InMAT radius~\cite{ma20123d,peters2016robust}}: interior shrinking ball radius of 3D medial axis transformation), density-based (i.e., the number of vertices and the density of triangle faces), and color-based (i.e., \textit{greenness} and \textit{HSV histograms}) features, and we concatenate these features into a feature vector $\textbf{F}_{i}$. 
We then use random forest (RF)~\cite{geurts2006extremely} to learn the probability of a face being \textit{non-planar} as
\begin{equation}
\label{eq:adaptive_probability}
G_i(L)= \frac{1}{\left | \tau  \right |}\sum_{t\in \tau}\log \big( P_{t}  (l_i  \;| \; \textbf{F}_{i})\big) ,
\end{equation}
where $ \tau $ is a set of decision trees, and the predicted probability from decision tree $t$ is denoted by $P_{t} \in [0, 1] $.
$L = \left \{0, 1\right \}$ represents the potential labels of a face $i$ (i.e., $l_i = 0$ for \textit{planar} and $l_i = 1$ for \textit{non-planar}). We learn a probability map (instead of binary classification) for the subsequent segment aggregation.

\paragraph{Incremental segmentation}

\begin{figure*}[!tb]
	\centering
	\includegraphics[width=1.0\linewidth]{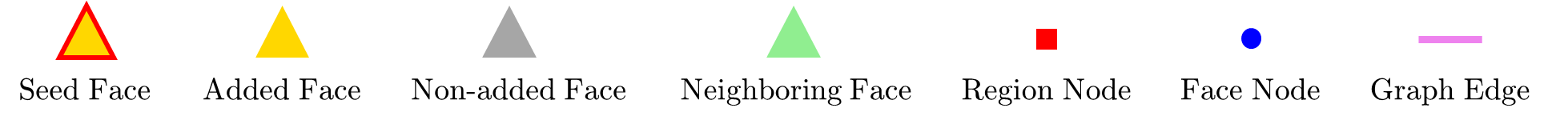}
	\begin{subfigure}{0.18\linewidth}
		\includegraphics[page=1,width=\linewidth]{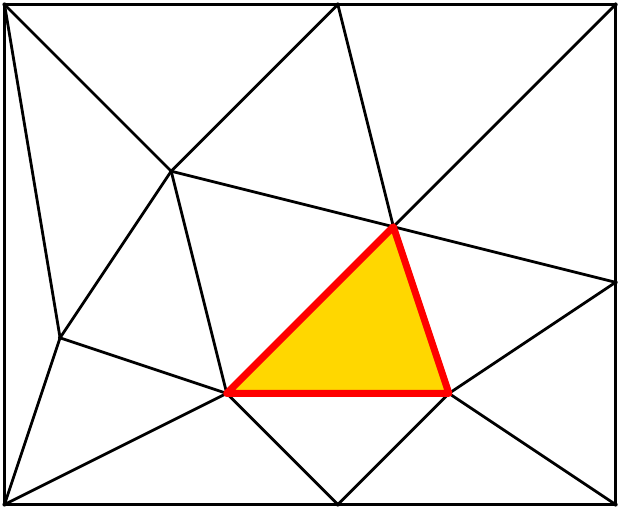}
		\caption{}
		\label{fig:overseg_process_1}
	\end{subfigure}
	\hfill
	\begin{subfigure}{0.18\linewidth}		
		\includegraphics[page=2,width=\linewidth]{overseg_process.pdf}
		\caption{}
		\label{fig:overseg_process_2}
	\end{subfigure}
	\hfill
	\begin{subfigure}{0.18\linewidth}		
		\includegraphics[page=3,width=\linewidth]{overseg_process.pdf}
		\caption{}
		\label{fig:overseg_process_3}
	\end{subfigure}
		\hfill
	\begin{subfigure}{0.18\linewidth}		
		\includegraphics[page=4,width=\linewidth]{overseg_process.pdf}
		\caption{}
		\label{fig:overseg_process_4}
	\end{subfigure}
		\hfill
	\begin{subfigure}{0.18\linewidth}		
		\includegraphics[page=5,width=\linewidth]{overseg_process.pdf}
		\caption{}
		\label{fig:overseg_process_5}
	\end{subfigure}
		\caption{An illustration of the first few steps of incremental segmentation. (a) The seed face with the highest \textit{planar} probability is shown in gold color. (b) The local graph is constructed on the seed face (represented by the region node) and its three neighboring faces (represented by the face node). (c) The labeling outcome of the Markov random field (MRF): one newly added face is used as a seed face, and two non-added faces will be labeled as visited faces for the current growth step. (d) A local graph is constructed based on the growing region (represented by the region node) and two neighboring faces (represented by the face node). (e) The new labeling outcome, where the newly added two faces will be used as seed faces for the next growing step.  }
	\label{fig:overseg_process}
\end{figure*}

\begin{figure}[!tb]
	\centering
	\begin{subfigure}{0.48\linewidth}
		\centering
		\captionsetup{justification=centering}
		\includegraphics[width=\linewidth]{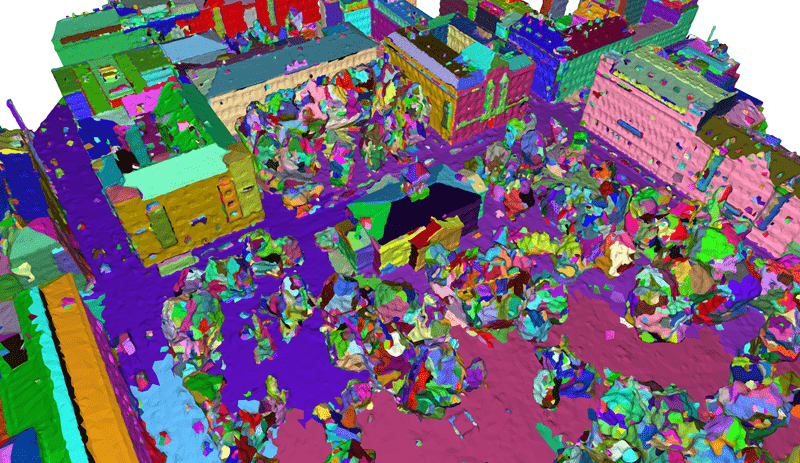}
		\caption{$\lambda_{d} = 1$, $\lambda_{m} = 0$, $\lambda_{g} = 0$}
		\label{fig:d1_m0_np0_po0}
	\end{subfigure}
	\hfill
	\begin{subfigure}{0.48\linewidth}
		\centering
		\captionsetup{justification=centering}
		\includegraphics[width=\linewidth]{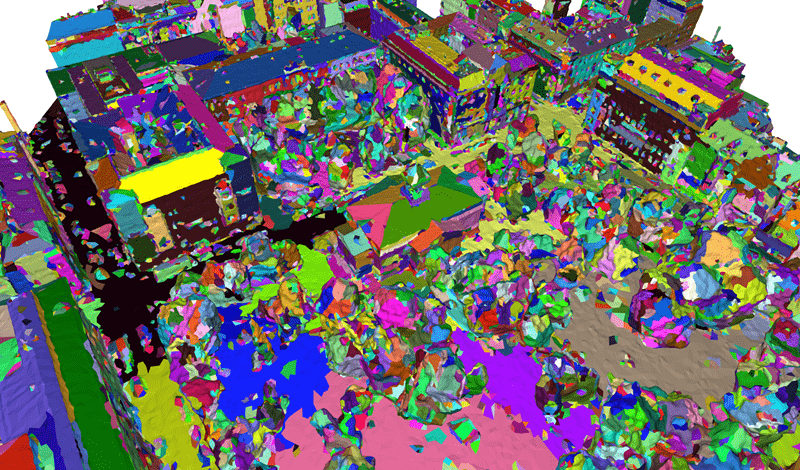}
		\caption{$\lambda_{d} = 2$, $\lambda_{m} = 0$, $\lambda_{g} = 0$}
		\label{fig:d2_m0_np0_po0}
	\end{subfigure}
	\begin{subfigure}{0.48\linewidth}
		\centering
		\captionsetup{justification=centering}
		\includegraphics[width=\linewidth]{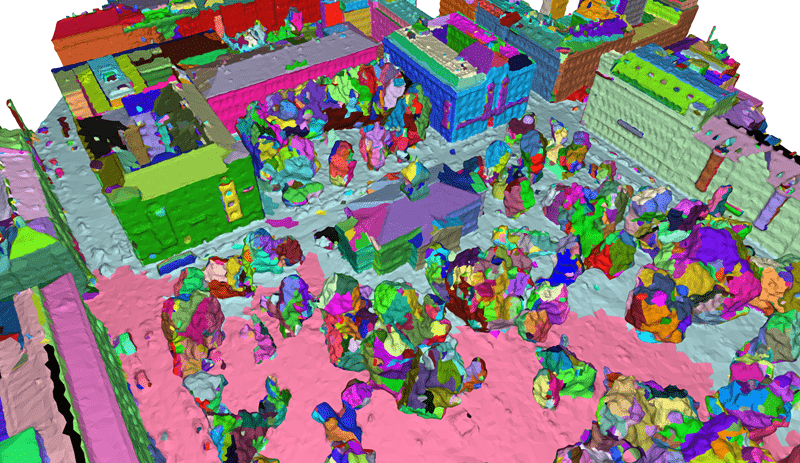}
		\caption{$\lambda_{d} = 1$, $\lambda_{m} = 1$, $\lambda_{g} = 0$}
		\label{fig:m1_np0_po0}
	\end{subfigure}
	\hfill
	\begin{subfigure}{0.48\linewidth}
		\centering
		\captionsetup{justification=centering}
		\includegraphics[width=\linewidth]{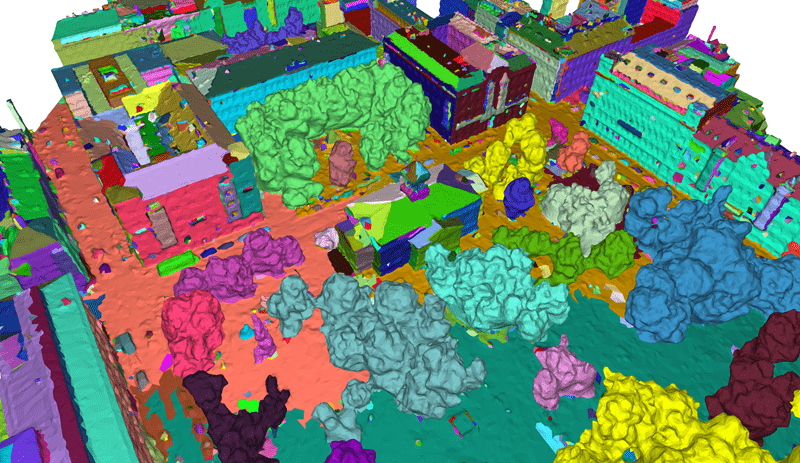}
		\caption{$\lambda_{d} = 1$, $\lambda_{m} = 0$, $\lambda_{g} = 1$}
		\label{fig:m0_np1_po0}
	\end{subfigure}
	\caption{The effect of the parameters $\lambda_{d}$, $\lambda_{m}$, and $\lambda_{g}$ on the over-segmentation. 
		These parameters provide control over the size and boundary smoothness of the segments.}
	\label{fig:adpativeconfig}
\end{figure}

Grouping all triangles into segments in one step using graph cuts would require the total number of segments, which is often not a priori.
Therefore, we use the learned \textit{planar} and \textit{non-planar} probability maps to incrementally aggregate the mesh faces into a set of locally homogeneous segments.
Inspired by Lafarge and Mallet~\cite{lafarge2012creating}, we accumulate faces for a segment by solving a binary labeling problem.
Starting from the face with the highest \textit{planar} probability (i.e., the current region $r$ has only a starting face at the beginning), we incrementally gather its neighboring face $i$ to the current region $r$ based on the labeling outcome of face $i$. The growing process is illustrated in~\Cref{fig:overseg_process}. 
Our idea is to grow a region ${r}$ if its neighboring face $i$ receives the same label.
We exploit a Markov Random Field (MRF) formulation to select the most suitable face for the aggregation in each growing iteration. 
The energy function $U(X)$ is defined as the sum of a unary term $\psi_{i}(x_i)$ and a pairwise term $\varphi_{i, r}(x_i,x_r)$, i.e.,

\begin{equation}\label{eq:MRF_local}
U(X)= \lambda_{d} \cdot\sum_{i\in A} \psi_{i}(x_i) + \lambda_{m} \cdot \sum_{\substack{i\in A}} \varphi_{i, r}(x_i, x_r)  ,
\end{equation}
where $A$ denotes the neighboring faces of the current growing region (i.e., the faces directly connected to $r$). $x_i$ and $x_r$ denote the binary labels that will be received by face $i$ and region $r$, respectively. 
A neighboring face can be added to the current region only if it receives the same label as the current region.
In our implementation, we fix the label of the current region to 0 (i.e., $x_r \equiv 0$) before minimizing the energy function. The face $i$ is added to $r$ only when $x_i = 0$  after the optimization.
$\lambda_{d} \ge 0$ and $\lambda_{m} \ge 0$ are the weights balancing the unary and pairwise terms.
A larger $\lambda_{d}$ can lead to an excessive number of segments with smaller under-segmentation errors (see~\Cref{fig:d1_m0_np0_po0} and~\Cref{fig:d2_m0_np0_po0}). 
In contrast, a larger $\lambda_{m}$ can result in fewer segments but may introduce larger under-segmentation errors (see~\Cref{fig:m1_np0_po0}). 

The \textbf{unary term} $\psi_{i}(x_i)$ measures the penalty of assigning a label $x_i$ to a face $i$. To define this term, we consider the geometric distance (for \textit{planar} regions) and the probability map (for \textit{non-planar} regions), which is formulated as

\begin{equation}\label{eq:unary}
	\begin{aligned}
			\psi(x_i) = \begin{cases}
				\begin{aligned}
					&\textup{min}\bigl\{\textup{d} (f_i, p_{r}), C_i \bigr\} , \hspace{1.2cm} \text{if  } \;  x_i = 0\\
					&1 - \textup{min}\bigl\{\textup{d} (f_i, p_{r}), C_i \bigr\} , \hspace{0.6cm} \text{if  } \;  x_i = 1
				\end{aligned}
			\end{cases}
		\end{aligned},
\end{equation}

\begin{equation}\label{eq:geo_term}
	\begin{aligned}
		C_i = \begin{cases}
			\begin{aligned}
				& 1 - \lambda_{g} \cdot G_i , \hspace{0.6cm} \text{if  } \;  l_i = 1 \wedge \; l_r = 1\\
				& \infty , \hspace{2.6cm} \text{otherwise} 
			\end{aligned}
		\end{cases}
	\end{aligned},
\end{equation}
where $\textup{d} (f_i, p_{r})$ measures the Euclidean distance between the farthest vertex of face $i$ and the fitted plane $p_{r}$ of the region $r$. 
The plane is obtained by linear least squares fitting using all the vertices of the region and dynamically updated when a new face has been added. 
During growing, when $x_i = 0$, $\textup{min}\bigl\{\textup{d} (f_i, p_{r}), C_i \bigr\}$ measures the cost of assigning face $i$ the same label as the current region $r$ (i.e., the cost of adding face $i$ to the current region $r$).
On the contrary, the cost is measured by $1 - \textup{min}\bigl\{\textup{d} (v_i, p_{r}), C_i \bigr\}$ when $x_i = 1$.
In particular, for the \textit{planar} case, since $C_{i}=\infty$, the geometric distance $\textup{d} (f_i, p_{r})$ is actually used as the cost measure.
For the \textit{non-planar} case, the prior term $G_{i}$ (see \Cref{eq:adaptive_probability}) is considered to define the cost $C_i$.
$\lambda_{g} \ge 0$ is a weight that controls the relative numbers of \textit{planar} and \textit{non-planar} segments (see~\Cref{fig:d1_m0_np0_po0} and~\Cref{fig:m0_np1_po0}).

The \textbf{pairwise term} $\varphi_{i, r}(x_i,x_r)$ is designed to control the smoothness degree during the growing process,
\begin{equation}\label{eq:pairwise}
\varphi_{i, r}(x_i, x_r)= \angle (\mathbf{n}_i, \mathbf{n}_{r})\cdot \mathbbm{1}(x_i\neq x_r),
\end{equation}
where $ \mathbf{n}_i $ and $ \mathbf{n}_{r} $ denote the normals of a neighboring triangle face $i$ and the region $r$, respectively. 
This term encodes the angle between these normal vectors, to reduce the normal deviation within the local neighborhood in the segmentation. 
$\mathbbm{1}(x_i\neq x_{r})$ is an indicator function that measures the coherence between $x_i$ and $x_r$.

The energy $U(X)$ is minimized using the $\alpha - \beta$ swap graph cut algorithm~\cite{boykov2001fast} to accumulate a face for the current segment.
The growth of a segment stops if no more faces can be accumulated. 
We then restart growing a new segment from the face with the highest \textit{planar} probability in the remaining set of faces.
The growing of segments is repeated until all mesh faces have been processed.

\subsection{Classification}
We construct a graph whose nodes encode features of the segments and edges encode interactions between segments. 
With this graph, the semantic segmentation of the mesh is achieved by classifying the segments using GCN.

\paragraph{Node feature embedding}
	In our graph, each node represents a segment and it encodes two types of features generated based on the vertices and face centroids of the segment: 
	1) $F_{l}(s_{k})_{256}$ is learned using PointNet~\cite{qi2017pointnet}, and the input to it is a point cloud of randomly sub-sampled points from mesh vertices and face centroids. The size of the feature vector is $128 \times 6$ and consists of XYZ and RGB;
	2) $F_{h}(s_{k})_{48}$ is generated from the handcrafted feature generator (HFG), and it contains the same type of features used for \textit{planar and non-planar classification} (see in~\Cref{sec:geo_classify}) and four additional shape-based features capturing local geometric differences (see~\Cref{tab:shapefeas}).

\begin{table}[!tb]
	\centering
	\noindent\adjustbox{max width=1.0\linewidth}
	{
		\begin{tabular}{lc|lc}
			\toprule
			\textbf{Compactness} & $CP_k = $\scalebox{1.4}{$\frac{4\cdot \pi \cdot area(s_k) }{C(s_k)^{2}}$} & \textbf{Shape Index} & $SI_k = $\scalebox{1.4}{$\frac{C(s_k)}{\sqrt[4]{area(s_k)}} $}\\ 
			\midrule
			\textbf{Straightness} & $SD_k = $\scalebox{1.4}{$\frac{\sum_{i=1}^{m} \frac{\lambda_{2}}{\lambda_{3}}}{m} $} &
			\textbf{Avg Distance} & $D_k = $\scalebox{1.4}{$\frac{\sum_{n}^{i=1} dist(p_i, P_k) }{n} $}\\ 
			
			\bottomrule
		\end{tabular}
	}
	\caption{Shape-based features defined on segments. 
		$C(s_k)$ is the circumference of a segment $s_k$. 
		$\lambda_{2}$ and $\lambda_{3}$ are the eigenvalues derived from the linear fitting line of $m$ boundary points in 3D~\cite{pearson1901liii}. 
		\textit{Avg Distance} measures the average distance from $n$ mesh vertices $p_i$ to the supporting plane $P_k$ of the segment.}
	\label{tab:shapefeas}
\end{table}

\paragraph{Edge feature embedding}
In contrast to graphs defined on 3D points or triangle faces, graphs defined on certain segmentation of the data can better capture global contextual relationships than simple adjacency connections.
The idea behind the designed graph is to give more prominence to the segment differences that are usually present in urban scenarios. 
To this end, we intend to include global features that fulfill two conditions. First, the global features should be generalizable, i.e., they can be captured in different scenarios. Second, the global features should be established between graph nodes that have large feature differences.
To make full use of the \textit{planar} and \textit{non-planar} segments and establish meaningful relationships between the segments, we propose a graph consisting of the following four types of edges (see~\Cref{fig:graphs2D}):

\begin{figure}[!tb]
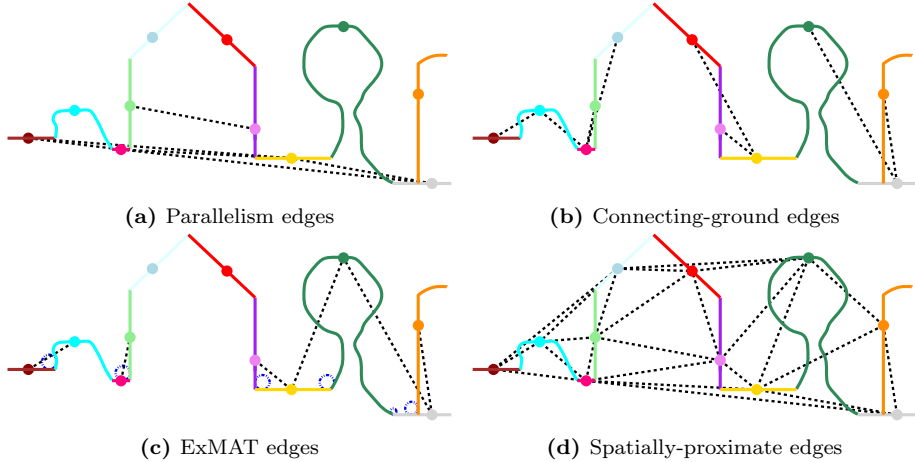

	\begin{subfigure}{0.49\linewidth}
		\centering
		\captionsetup{justification=centering}
		\includegraphics[page=4,width=\linewidth]{semantic_graph_new_simplified.pdf}
		\caption{Parallelism edges}
		\label{fig:parallelism_graph}
	\end{subfigure}
	\hfill
	\begin{subfigure}{0.49\linewidth}	
		\centering
		\captionsetup{justification=centering}	
		\includegraphics[page=3,width=\linewidth]{semantic_graph_new_simplified.pdf}
		\caption{Connecting-ground edges}
		\label{fig:local_ground_graph}
	\end{subfigure}
	\hfill
	\begin{subfigure}{0.49\linewidth}
		\centering
		\captionsetup{justification=centering}
		\includegraphics[page=5,width=\linewidth]{semantic_graph_new_simplified.pdf}
		\caption{ExMAT edges}
		\label{fig:exmat_graph}
	\end{subfigure}
	\hfill
	\begin{subfigure}{0.49\linewidth}
		\centering
		\captionsetup{justification=centering}
		\includegraphics[page=6,width=\linewidth]{semantic_graph_new_simplified.pdf}
		\caption{Spatially-proximate edges}
		\label{fig:delaunay_graph}
	\end{subfigure}
	
	\caption{An illustration of the four types of edges in our graph. 
		Each color indicates a segment encoded as a node (i.e., the colored dot on each segment) in the graph.
		The dash lines denote the graph edges. 
		In (c), the blue circles represent the exterior shrinking balls.}
	\label{fig:graphs2D}
\end{figure}

\begin{enumerate}[label={\arabic*)}]
	\item \textit{parallelism edges}: edges connecting parallel \textit{planar} segments. Two \textit{planar} segments are considered parallel if the angle between their supporting planes is smaller than a threshold ($5^{\circ}$ in our experiments). These edges mainly connect the \textit{planar} segments belonging to man-made objects.
	
	\item \textit{connecting-ground edges}: edges connecting segments and their local ground planes. A local ground plane is identified as the lowest and largest \textit{planar} segment in a cylindrical neighborhood (30 $m$ in our experiments) around the boundary vertices of the segment. These edges primarily capture the relationship between the ground and all non-ground objects.
	
	\item \textit{exterior medial axis transform (ExMAT) edges}. We first build the ExMAT (i.e., exterior shrinking ball radius of 3D medial axis transformation)~\cite{ma20123d,peters2016robust} on the segments, and we introduce graph edges that link the segments connected by the exterior shrinking ball (see~\Cref{fig:exmat_graph}). Since the external skeleton usually corresponds to the joints between objects, ExMAT edges allow connecting segments that are adjacent but belong to different objects.
	
	\item \textit{spatially-proximate edges}. We first build a 3D Delaunay triangulation~\cite{jaromczyk1992relative} with the input mesh vertices and the centroids of mesh faces.
	Two segments are connected by an edge if at least one pair of points from the two segments are connected by a Delaunay edge. 
	This type of edges allow the encoding of contextual information on different scales.  
	Particularly, these edges contribute to capturing the relationships between urban objects from short-range (for objects that are close to the ground) to long-range (for objects that are far away from the ground).
\end{enumerate}

With the above graph edges, we define the edge feature $F_{h}(e_{k,k+1}) = log (F_{h}(s_{k}) / F_{h}(s_{k+1}))$, where $s_{k}$ and $s_{k+1}$ are the two segments connected by an edge $e_{k,k+1}$. 
We also introduced two additional edge features defined as the mean and standard deviation of the vertex offsets of the segment boundaries, in which the offset is defined using the closest point pair between two segments.

\begin{figure}[!tb]
	\centering
	\includegraphics[width=0.7\linewidth]{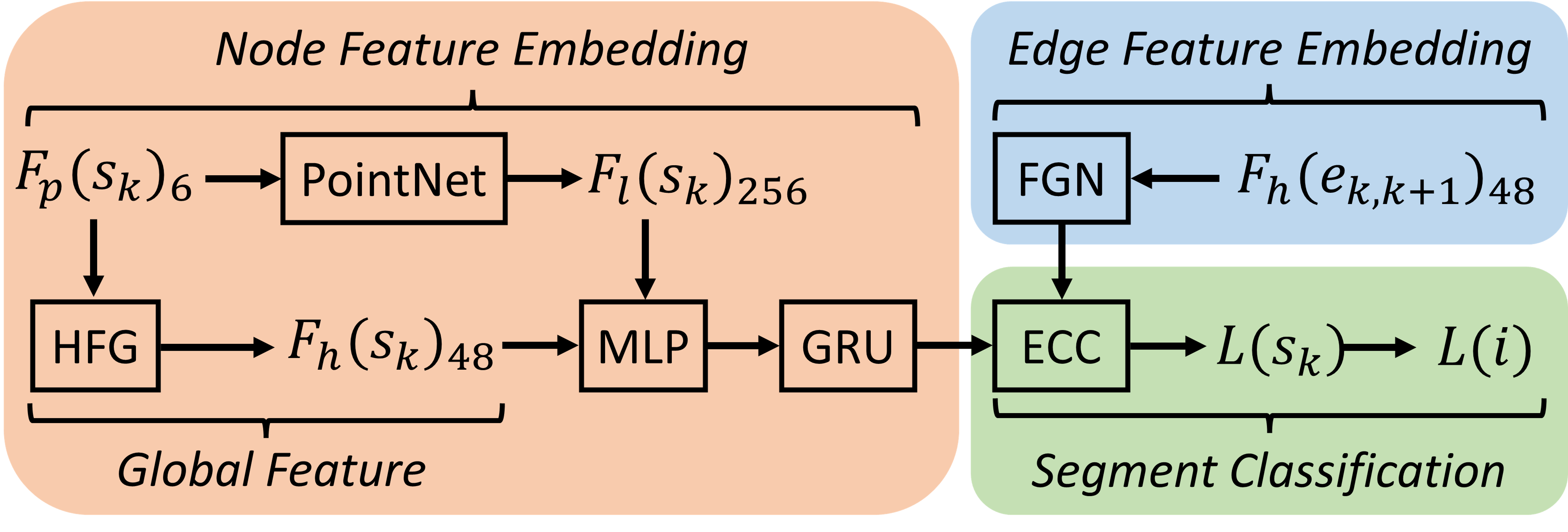}
	\caption{The feature embedding components for segment classification.
		Our network takes mesh vertices and face centers (with XYZ and RGB) of each segment, denoted as $F_{p}(s_{k})_{6}$, as input.
		PointNet~\cite{qi2017pointnet} is used to learn features $F_{l}(s_{k})_{256}$ for each segment. 
		The handcrafted features $F_{h}(s_{k})_{48}$ are computed using the feature generator (HFG).
		These two types of features are then processed jointly by the MLP and then refined in the GRU. 
		The $F_{h}(e_{k, k+1})_{48}$ are handcrafted edge features and input to a filter generating network (FGN).
		The final classification is obtained using edge-conditioned convolution (ECC) that takes both node and edge features as input.
		The output segment labels $L(s_k)$ are then transferred to face labels $L(i)$. 	  
	}
	\label{fig:semantic_seg_pipeline}
\end{figure}

\paragraph{Segment classification}
Based on the graph and feature embedding (see~\Cref{fig:semantic_seg_pipeline}), we exploit a GCN~\cite{li2015gated} to classify the segments. 
The node features learned from PointNet~\cite{qi2017pointnet} and the handcrafted features are concatenated and fed to MLP (Multilayer perceptron) to output a 64D feature vector that serves as the hidden state of the Gated Recurrent Unit (GRU: a gating mechanism in recurrent neural networks for updating and resetting hidden states to capture short-term and long-term dependencies in sequence)~\cite{cho2014learning}.
We apply ReLU activation~\cite{nair2010rectified} and batch normalization~\cite{ioffe2015batch} for each hidden layer of all MLPs.
The computed edge features are used as the input to the Filter Generating Network (FGN: a sequence of MLPs with widths of 32, 128, and 64 to output a 64D edge feature vector)~\cite{landrieu2018large}.
The output edge weights are then used to update the hidden state and refine the GRUs via Edge-Conditioned Convolution (ECC: a dynamic edge-conditioned filter that computes element-wise vector-vector multiplication for each edge and averages the results over respective nodes)~\cite{simonovsky2017dynamic}.
To alleviate class imbalance, we apply a standard cross-entropy loss~\cite{szegedy2016rethinking} $l_n = -w_{y_n} \log \frac{\exp(x_{n,y_n})}{\sum_{c=1}^C \exp(x_{n,c})}$ weighted by $w_{y_n} =\sqrt{N/n_{c}}$, where $x$ is the input, and $y$ is the target. $N$ denotes the total number of segments, and $n_{c}$ represents the number of segments in each class $c$.
The final output is per-segment labels that are then transferred to the faces of the input mesh.

%% file: source/experiments.tex
\section{Evaluation}  \label{sec:experiments}
\subsection{Data Split}

We have implemented our mesh over-segmentation with CGAL~\cite{cgal:eb-20b} and Easy3D~\cite{easy3d2021}, and the semantic classification with PyTorch~\cite{pytorch2019}. 
All experiments were carried out on a desktop PC with a 3.5GHz CPU and a GTX 1080Ti GPU.

We have used the SUM dataset~\cite{gao2021sum} and H3D dataset~\cite{kolle2021h3d} to evaluate our method. To the best of our knowledge, SUM is the largest benchmark dataset for semantic urban meshes, which covers about 4 $km^2$ of Helsinki (Finland) with six object classes: \textit{terrain} (\textit{terra.}), \textit{high vegetation} (\textit{h-veg.}), \textit{building} (\textit{build.}), \textit{water}, \textit{vehicle} (\textit{vehic.}), and \textit{boat}. The whole dataset contains 64 tiles each covering an 250 $m$ $\times $ 250 $m$ area. Following the SUM baseline, we used 40 tiles (62.5\% of the whole dataset) for training, 12 tiles (18.75\%) for the test, and 12 tiles for validation. 
The H3D dataset covers about 0.19 $km^2$ area of the village of
Hessigheim (Germany) with 11 classes: \textit{Low Vegetation}, \textit{	Impervious Surface}, \textit{Vehicle}, \textit{Urban Furniture}, \textit{	Roof}, \textit{Facade}, \textit{Shrub}, \textit{Tree}, \textit{	Soil/Gravel}, \textit{Vertical Surface}, and \textit{Chimney}. We follow the data splits in H3D~\cite{kolle2021h3d}, and we further merge small mesh tiles into a large one to obtain more contextual information.

\subsection{Evaluation metrics}
\label{sec:metric_eval}

\paragraph{Metrics for over-segmentation} 
Our over-segmentation aims to produce homogeneous segments to better facilitate semantic segmentation. We propose three novel evaluation metrics focusing on the impact of the over-segmentation on the final semantic segmentation: \textit{object purity (OP)}, \textit{boundary precision (BP)}, and \textit{boundary recall (BR)}. 

\begin{figure}[!tb]
	\begin{subfigure}{0.23\textwidth}
		\includegraphics[page=7,width=\linewidth]{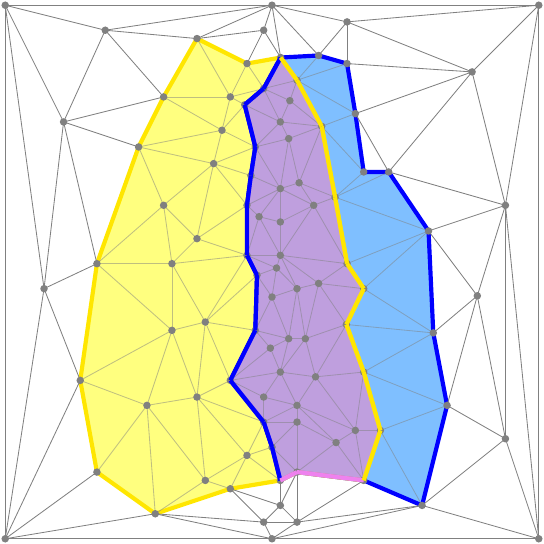}
		\caption{}
	\end{subfigure}
	\hspace*{\fill}
	\begin{subfigure}{0.23\textwidth}
		\includegraphics[page=4,width=\linewidth]{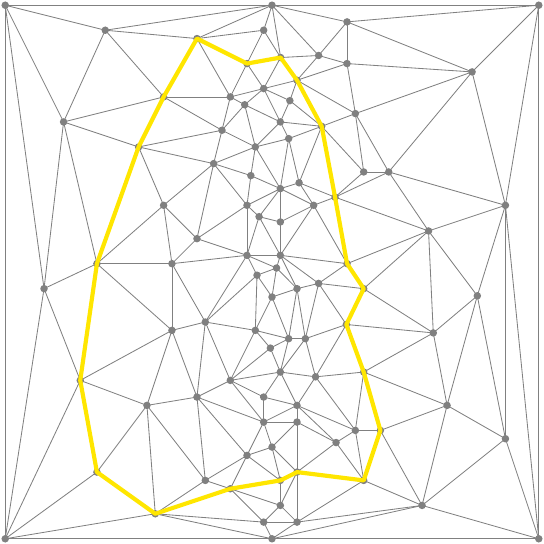}
		\caption{}
	\end{subfigure}
	\hspace*{\fill}
	\begin{subfigure}{0.23\textwidth}
		\includegraphics[page=5,width=\linewidth]{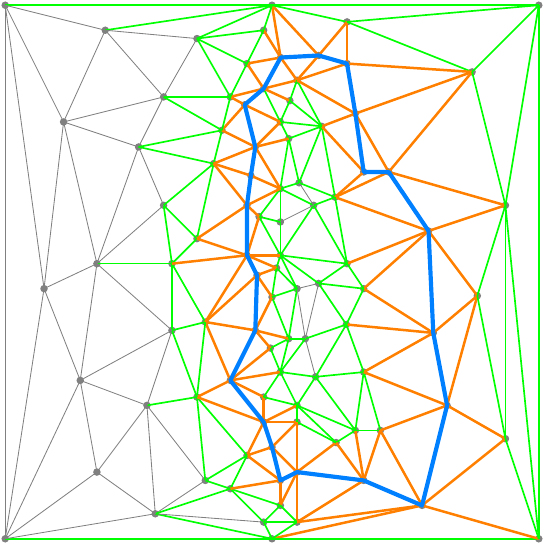}
		\caption{}
	\end{subfigure}
	\hspace*{\fill}
	\begin{subfigure}{0.23\textwidth}
		\includegraphics[page=6,width=\linewidth]{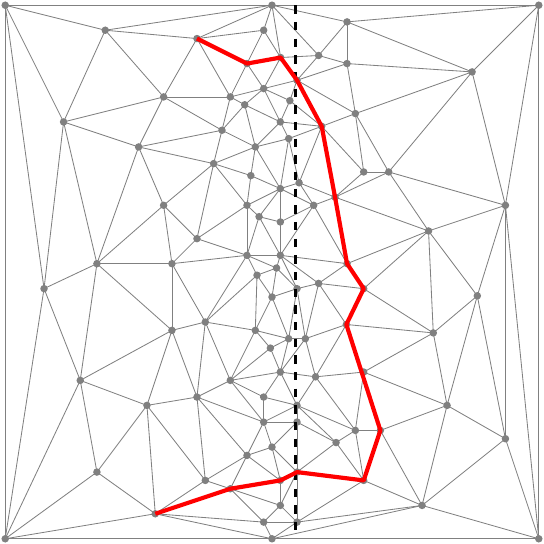}
		\caption{}
	\end{subfigure}
	\caption{
		2D schematic of \textit{object purity}, \textit{boundary precision}, and \textit{boundary recall}. 
		In (a), the yellow region represents the ground truth segment $G$.
		The blue region represents the generated segment $S$.
		The pink region represents the largest overlapping region between a segment and the ground-truth segments.
		In (b), the yellow edges represent the border $B_G$ of the ground truth segment.
		In (c), the blue edges represent the border $B_S$ of the generated segment. The orange edges represent the first ring of $B_S$. The green edges represent the second ring of $B_S$.
		In (d), the red edges are the intersection of $B_G$ and $B_S$ (as well as its first two rings). The black dashed line represents the true border between objects.
	}
	\label{fig:opbrbp}
\end{figure}

Since our goal is semantic segmentation, the best achievable over-segmentation is identical to the ground truth semantic segmentation. 
In this ideal situation, each segment covers exactly an individual object, and its boundaries perfectly align with the object boundaries. Thus, similar to \textit{intersection over union}, we define \textit{object purity} as 
\begin{equation}\label{eq:aoa}
OP(S, G) = \frac{\sum_{k} purity(s_k, G)}{area(G)},
\end{equation}
where $S = \{s_k\}$ denotes the set of segments in our over-segmentation, and $G = \{g_k\}$ are the segments extracted as connected components from the ground-truth semantic segmentation. 
$purity(s_k, G)$ measures the surface area of the largest overlapping region between a segment and the ground-truth segments (see~\Cref{fig:opbrbp}). 

\textit{Boundary precision} measures the correctness of the segment boundaries. Thus, it is defined to quantify how much the segment boundaries overlap with the boundaries of the ground-truth semantic segmentation,
\begin{equation}\label{eq:border_precission}
BP(B_S, B_G) = \frac{length(B_S \cap B_G)}{length(B_S)},
\end{equation}
where $B_S$ and $B_G$ denote the boundaries of the over-segmentation and those of the ground truth of the semantic segmentation, respectively. 
The function $length(\cdot)$ quantifies the total length of a set of segment boundaries.
To handle noisy and dense meshes, we allow a tolerance when looking for overlapping boundary edges. 
Specifically, two edges $e_1$ and $e_2$ are considered overlapping if the two endpoints of $e_2$ fall within the 2-ring neighborhood of the endpoints of $e_1$ (see~\Cref{fig:opbrbp}).	

\textit{Boundary recall} measures the completeness of the segment boundaries, defined as
\begin{equation}\label{eq:border_recall}
BR(B_S, B_G) = \frac{length(B_S \cap B_G)}{length(B_G)}.
\end{equation}

\paragraph{Metrics for semantic segmentation} 
To evaluate semantic segmentation results, we measure the precision, recall, F1 score, and intersection over union (IoU) for each object class, and we also record the overall accuracy (OA), mean accuracy (mAcc), and mean intersection over union (mIoU) of all object classes. 

\subsection{Evaluation of over-segmentation}
\begin{figure*}[!th]
	\begin{subfigure}[t]{0.23\linewidth}
		\includegraphics[width=\linewidth]{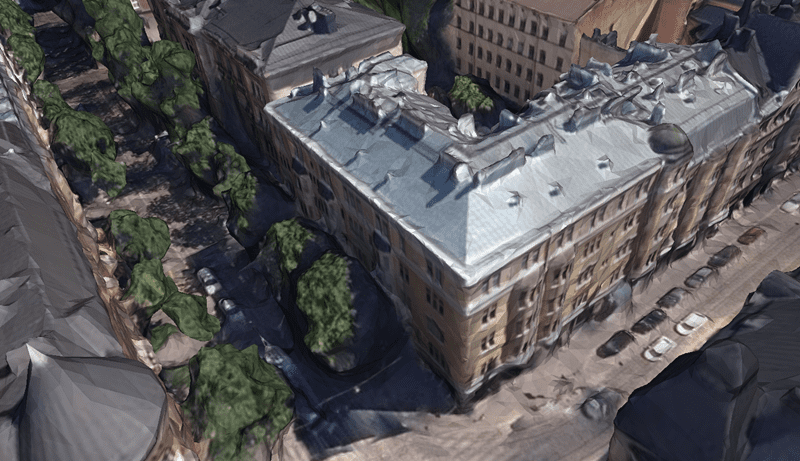}
		\caption{Input}
		\label{fig:adap_tex_loc}
	\end{subfigure}
	\hfill
	\begin{subfigure}[t]{0.23\linewidth}
		\includegraphics[width=\linewidth]{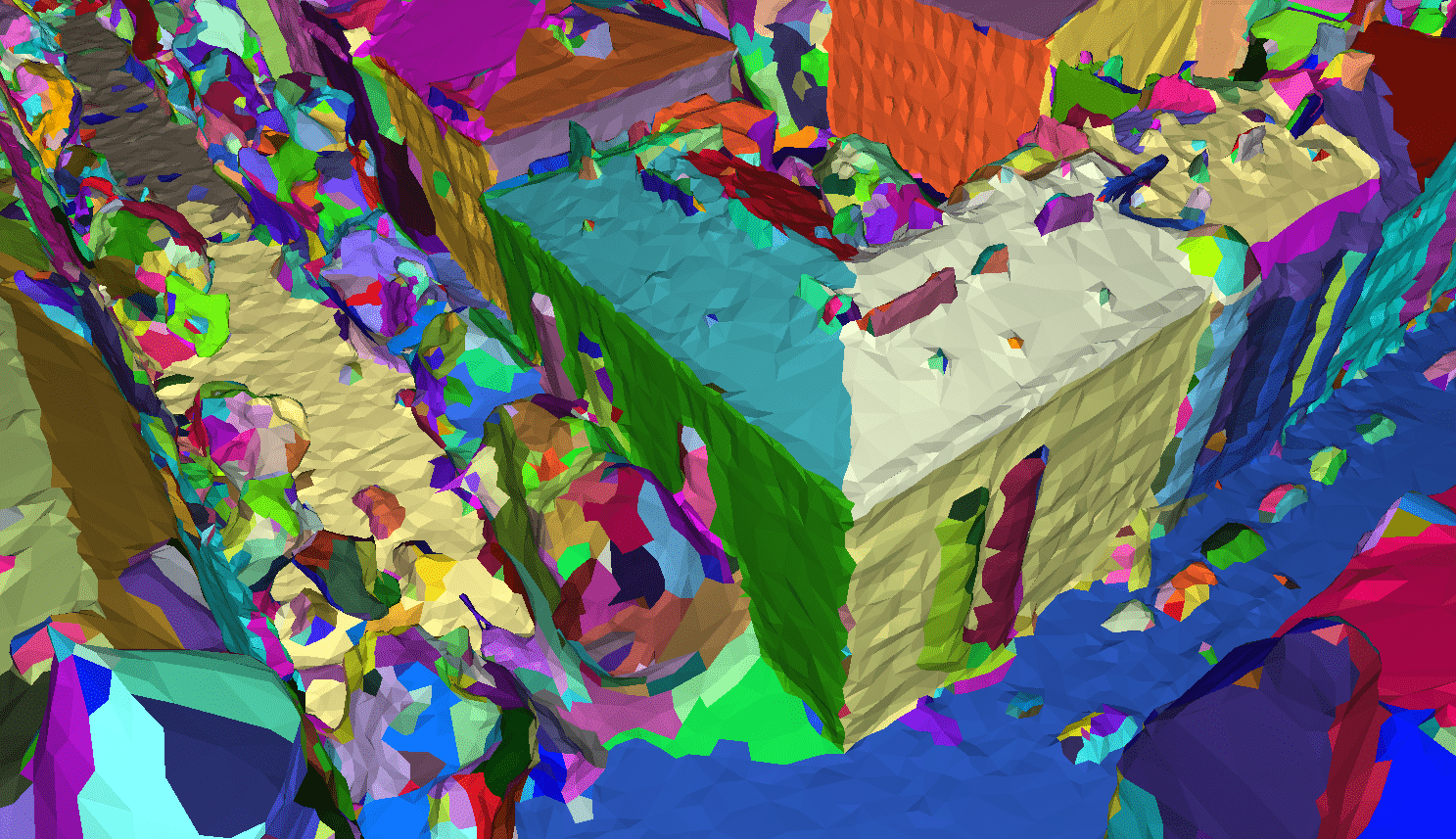}
		\caption{RG~\cite{lafarge2012creating}. 13656}
		\label{fig:rg_loc_seg_num}	
	\end{subfigure}
	\hfill
	\begin{subfigure}[t]{0.23\linewidth}
		\includegraphics[width=\linewidth]{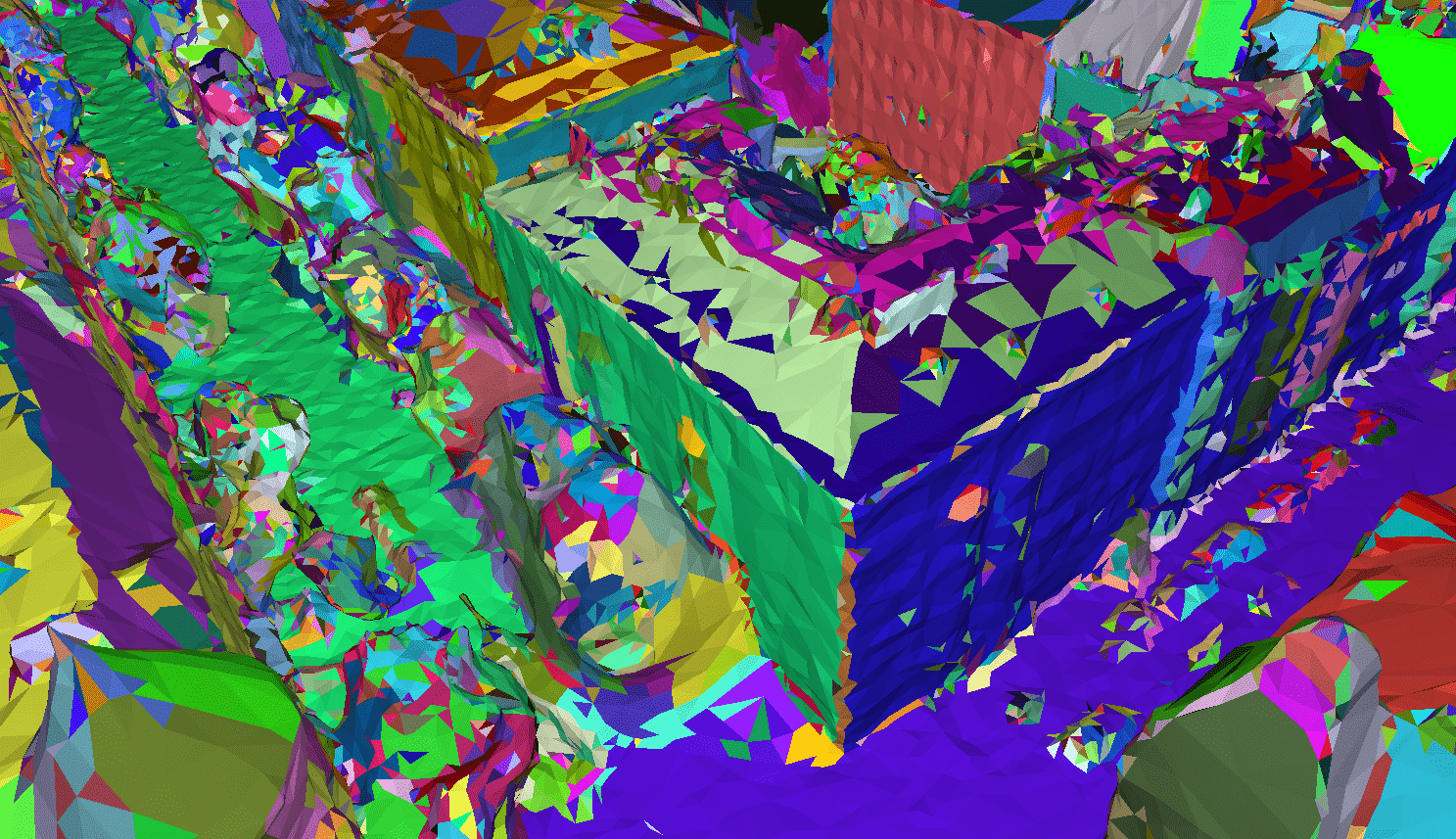}
		\caption{RA~\cite{schnabel2007efficient}. 446424}
		\label{fig:ransac_seg_num}
	\end{subfigure}
	\hfill
	\begin{subfigure}[t]{0.23\linewidth}
		\includegraphics[width=\linewidth]{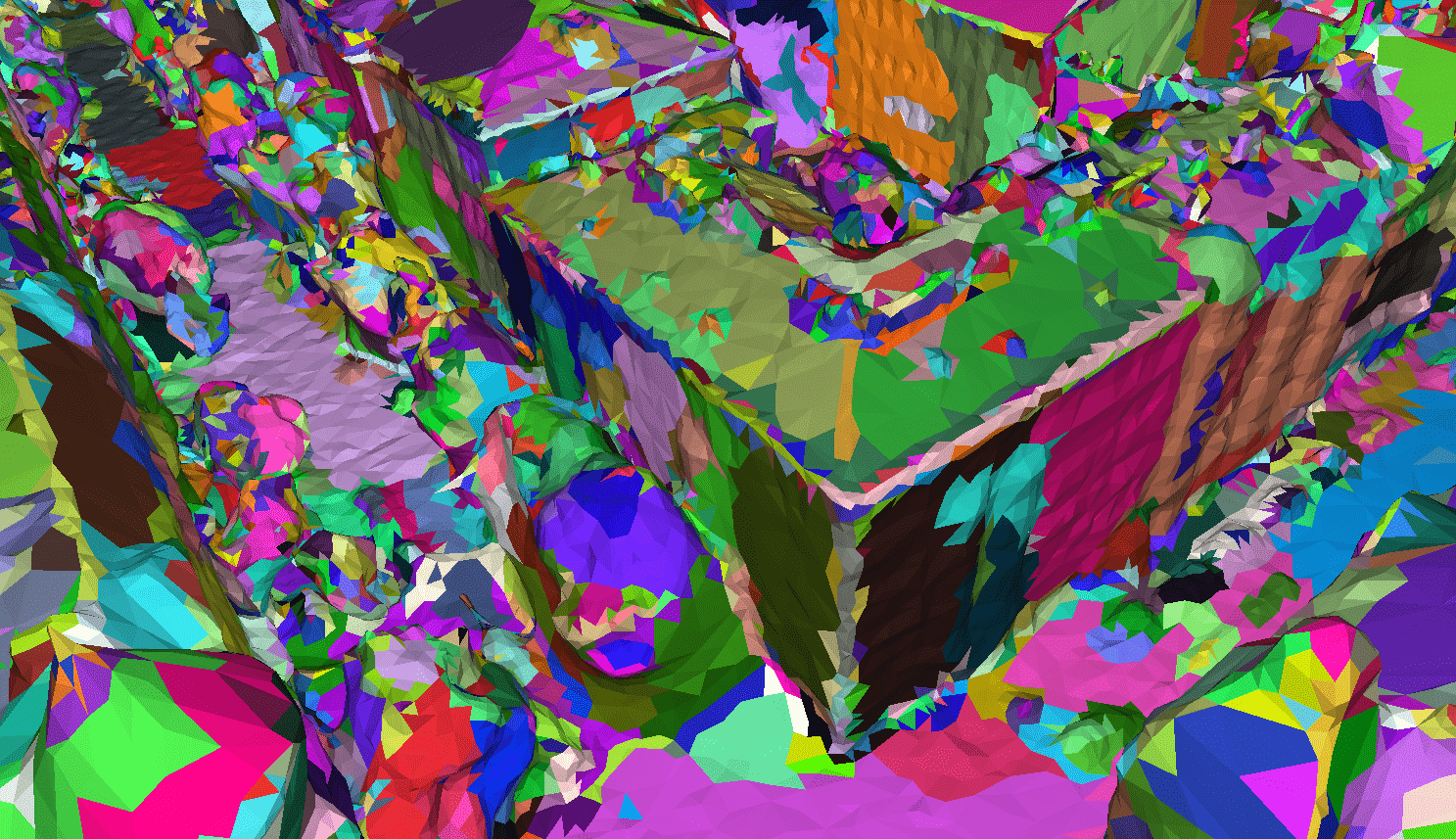}
		\caption{GP~\cite{landrieu2018large}. 30586}
		\label{fig:gp_loc_seg_num}
	\end{subfigure}
	\begin{subfigure}[t]{0.23\linewidth}
		\includegraphics[width=\linewidth]{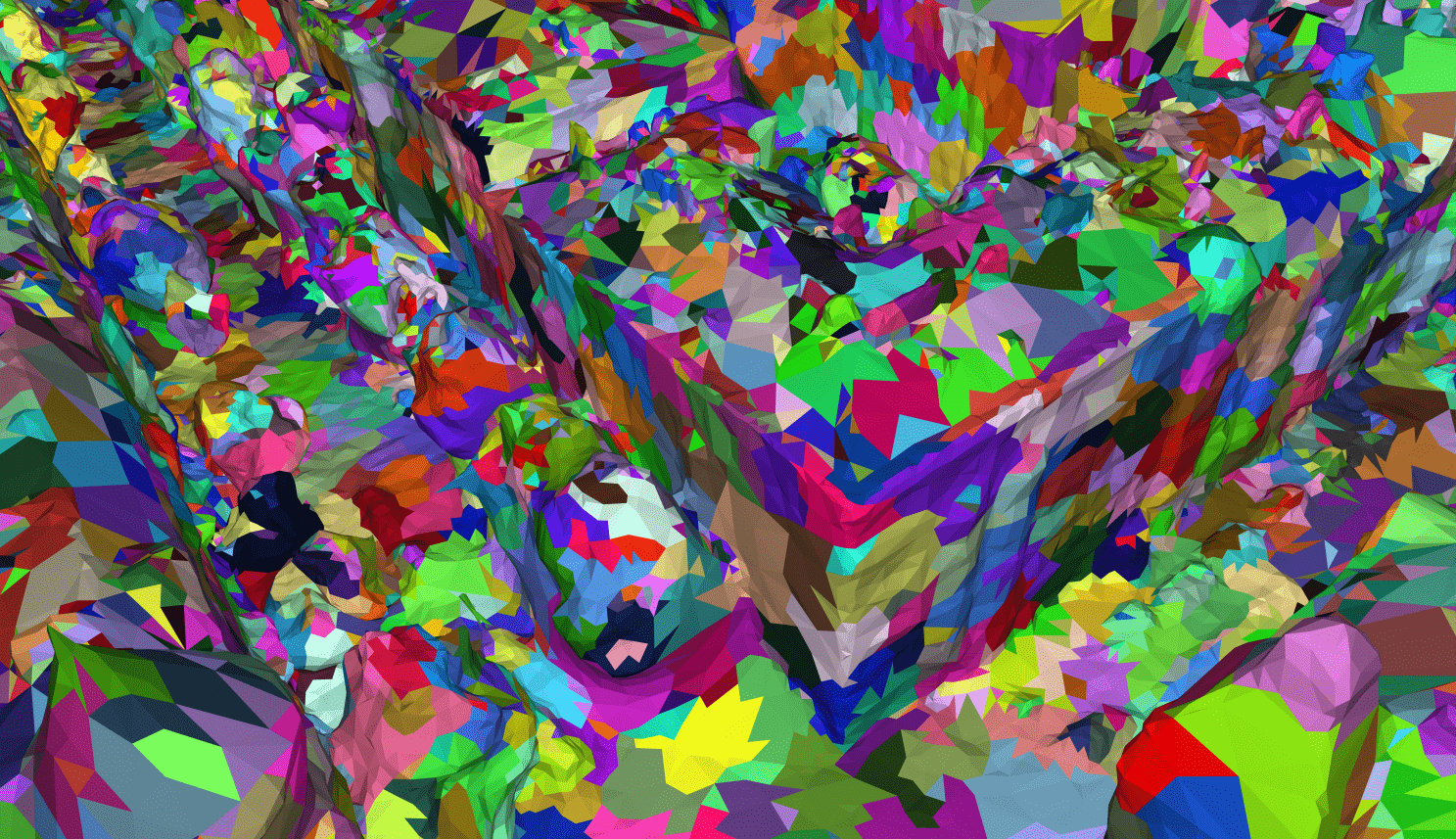}
		\caption{SSP~\cite{landrieu2019point}. 30375}
		\label{fig:ssp_loc_seg_num}
	\end{subfigure}
	\hfill
	\begin{subfigure}[t]{0.23\linewidth}		
		\includegraphics[width=\linewidth]{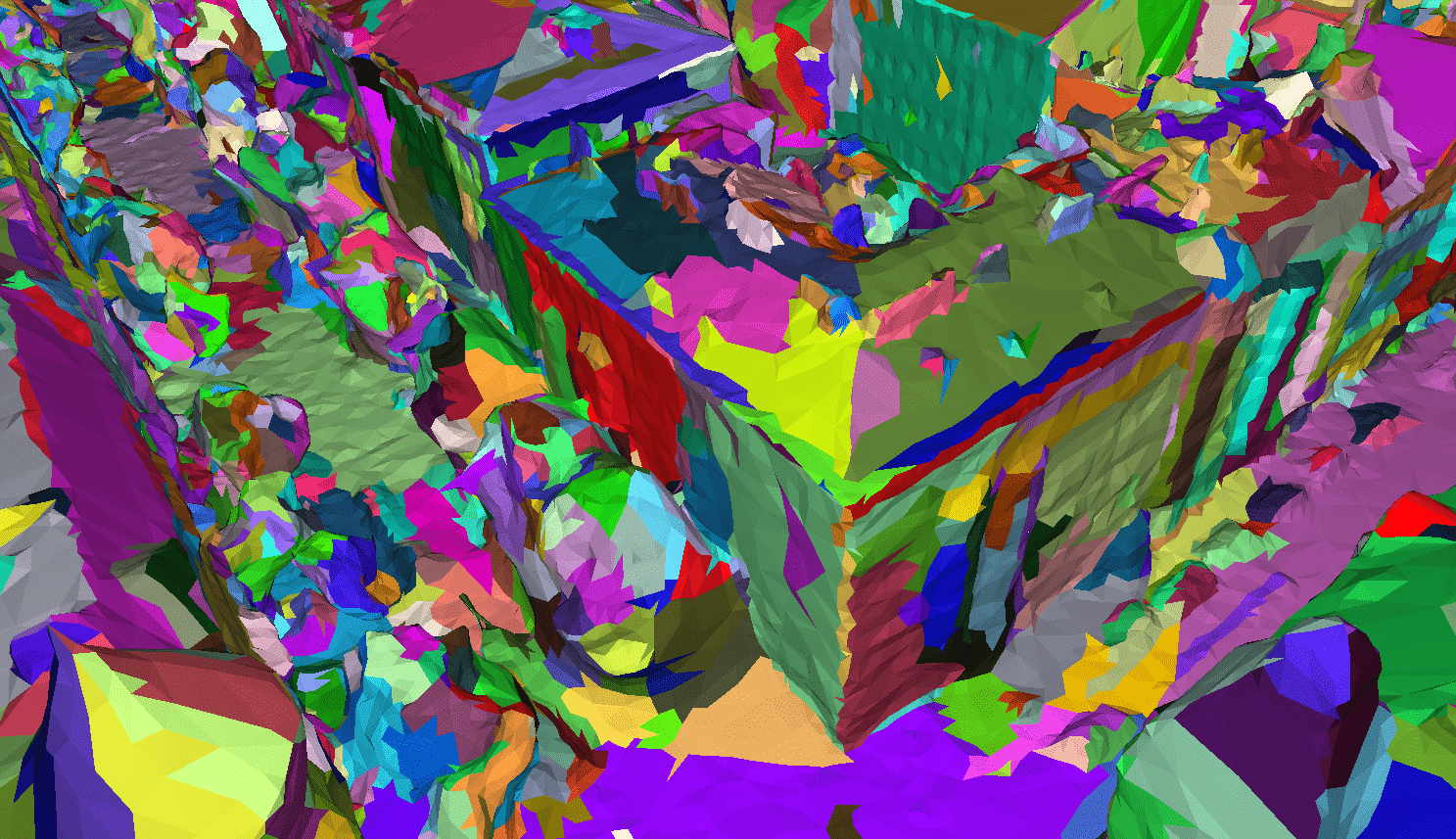}
		\caption{VSA~\cite{cohen2004variational}. 14177}
		\label{fig:vsa_loc_seg_num}
	\end{subfigure}
	\hfill
	\begin{subfigure}[t]{0.23\linewidth}
		\includegraphics[width=\linewidth]{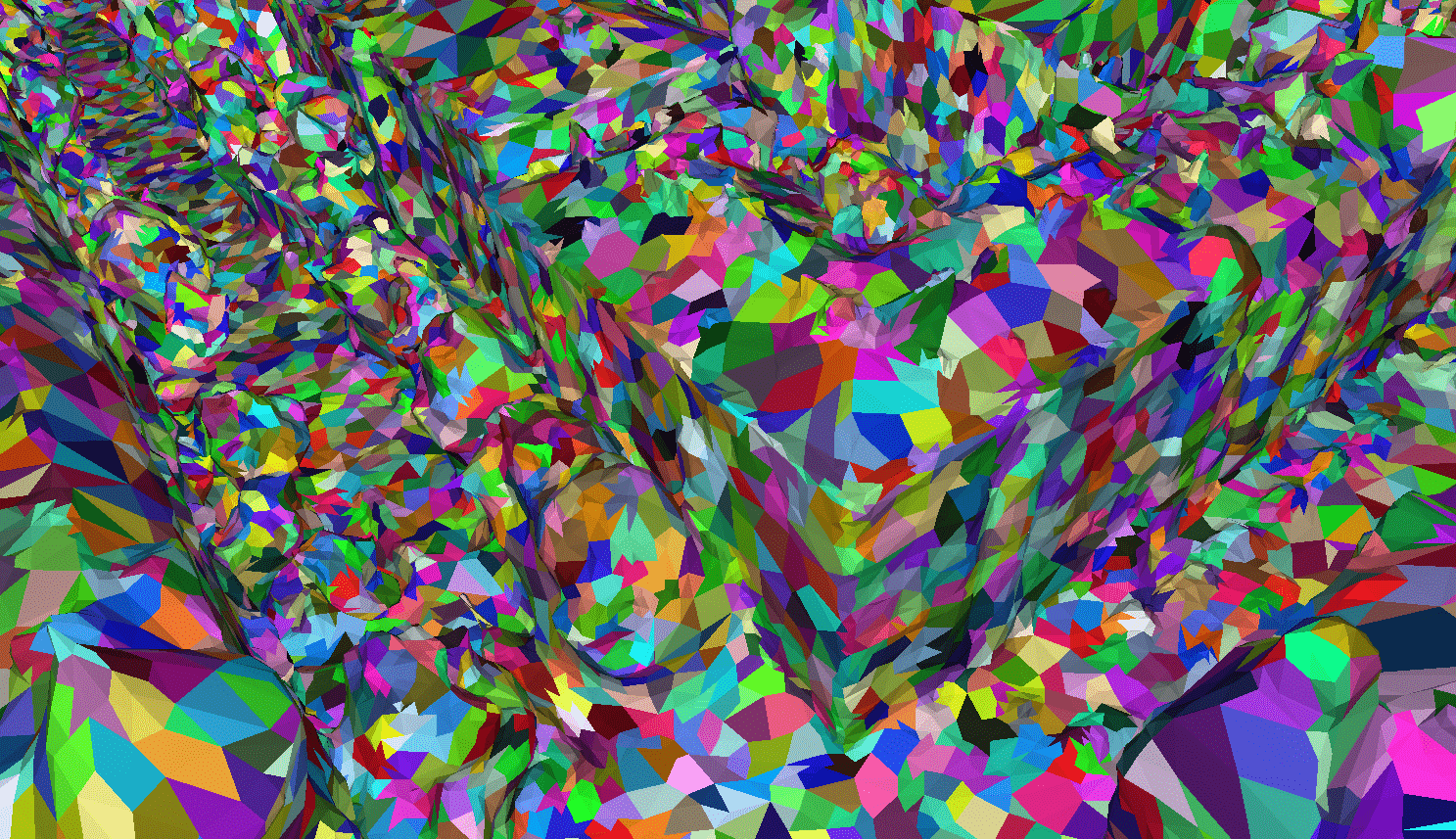}
		\caption{SPV~\cite{lin2018toward}. 53554}
		\label{fig:spvox_loc_seg_num}
	\end{subfigure}
	\hfill
	\begin{subfigure}[t]{0.23\linewidth}
		\includegraphics[width=\linewidth]{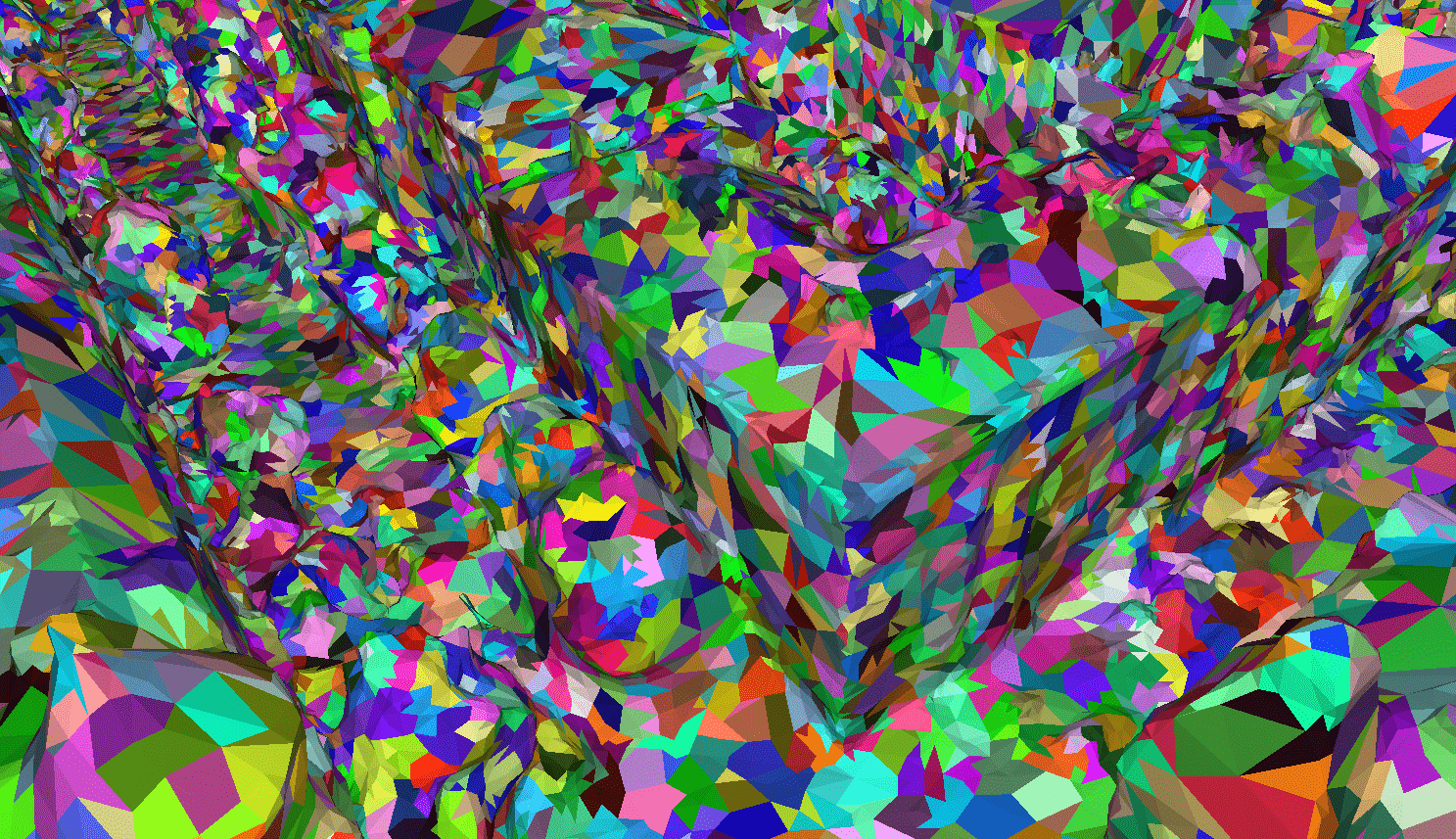}
		\caption{Vccs~\cite{papon2013voxel}. 51199}
		\label{fig:vccs_loc_seg_num}
	\end{subfigure}
	\begin{subfigure}[t]{0.23\linewidth}
		\includegraphics[width=\linewidth]{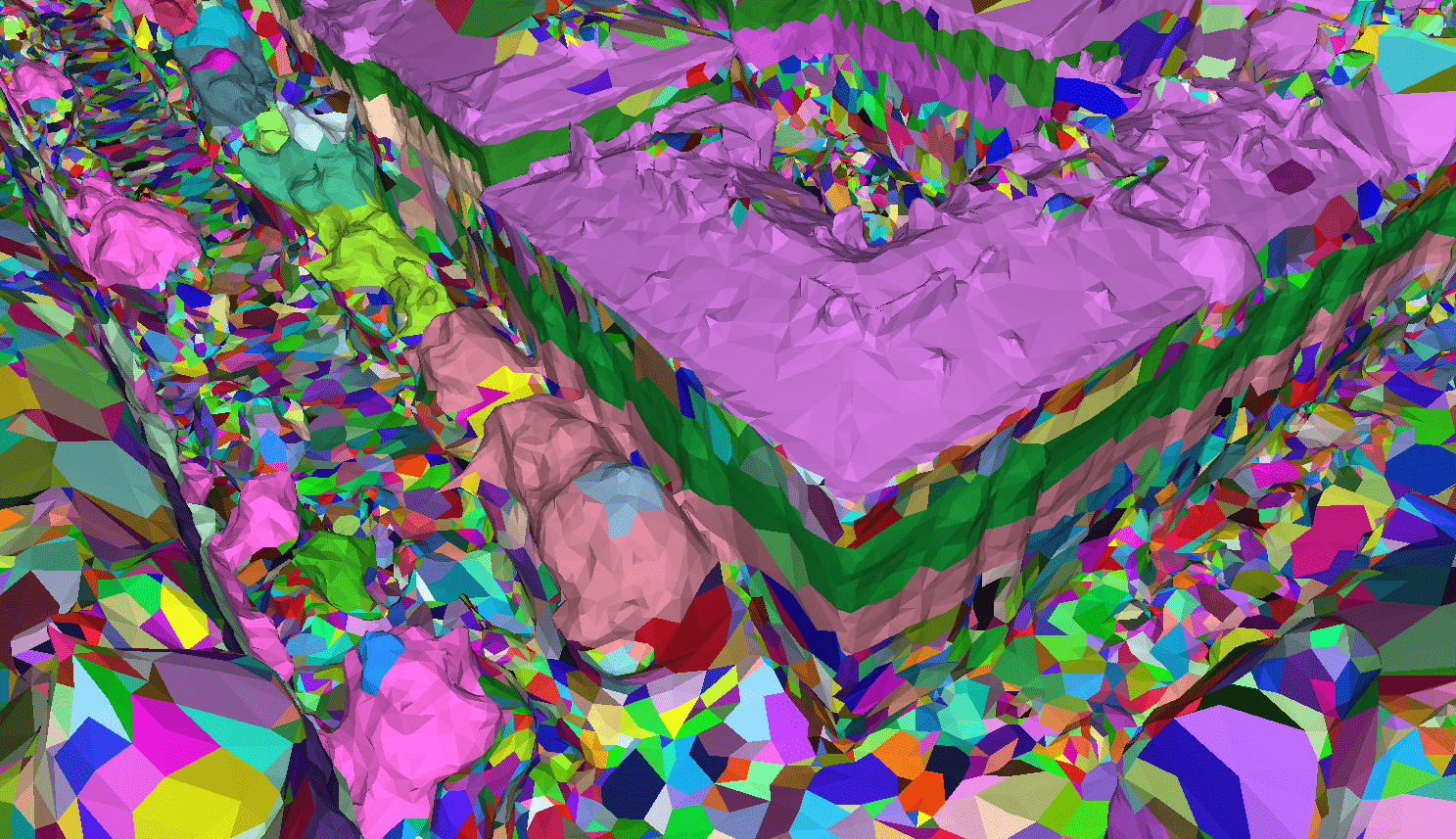}
		\caption{SC~\cite{verdie2015lod}. 85126}
		\label{fig:sc_loc_seg_num}
	\end{subfigure}
	\hfill
	\begin{subfigure}[t]{0.23\linewidth}
		\includegraphics[width=\linewidth]{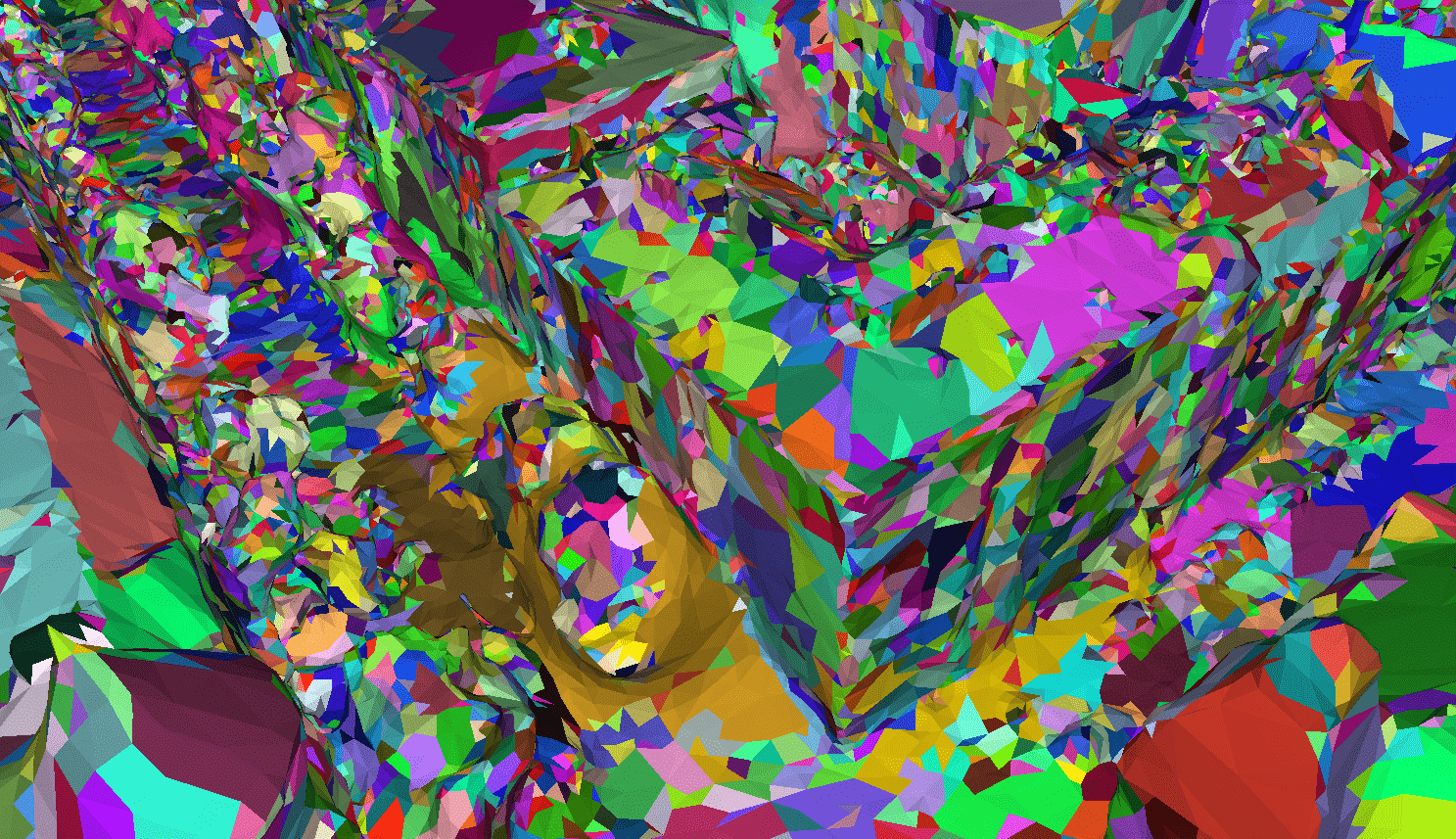}
		\caption{SP~\cite{rouhani2017semantic}. 92010}
		\label{fig:sp_loc_seg_num}
	\end{subfigure}
	\hfill
	\begin{subfigure}[t]{0.23\linewidth}
		\includegraphics[width=\linewidth]{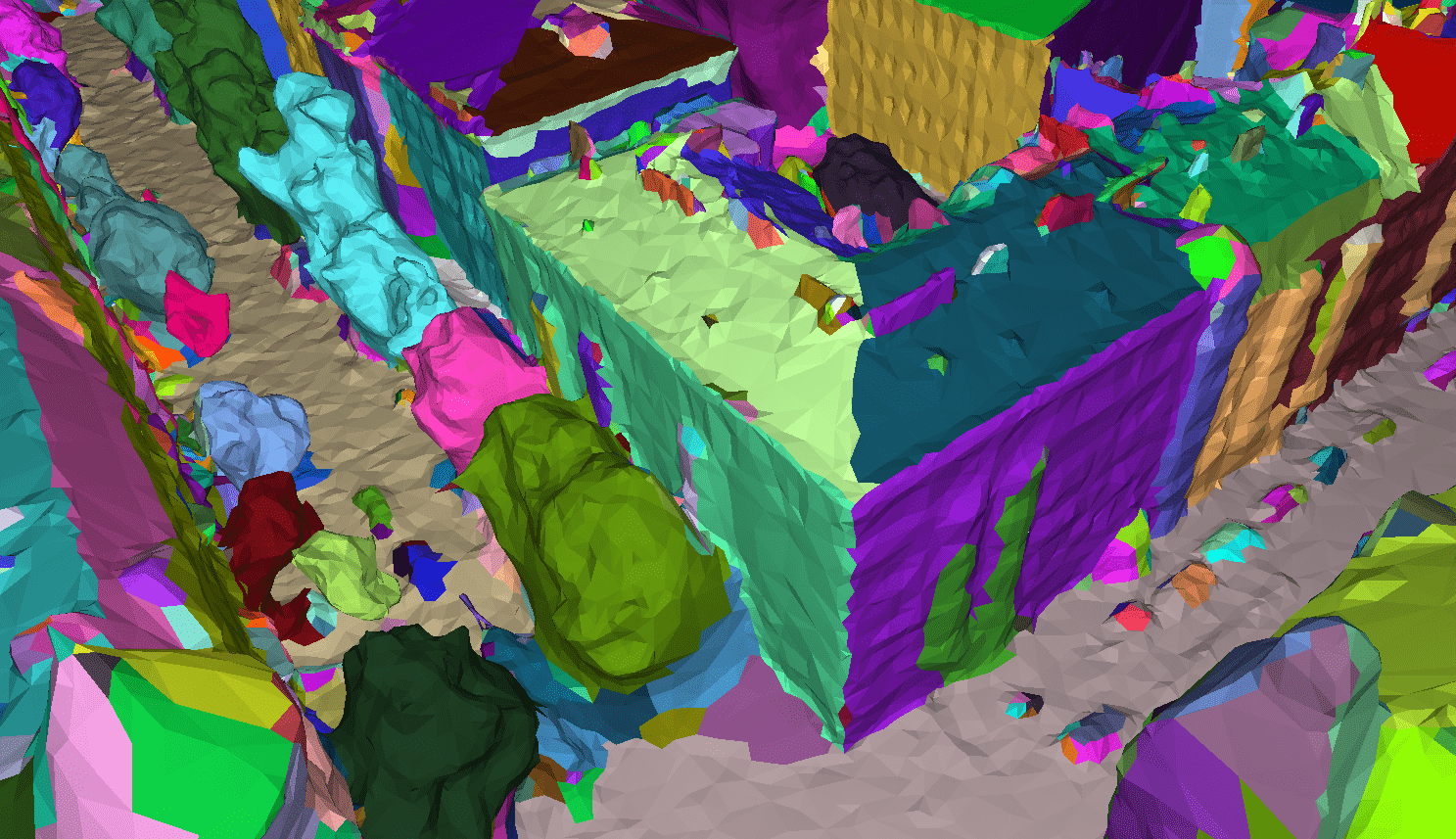}
		\caption{Ours. 8637}
		\label{fig:adap_mrf_loc_seg_num}
	\end{subfigure}
	\hfill
	\begin{subfigure}[t]{0.23\linewidth}
		\includegraphics[width=\linewidth]{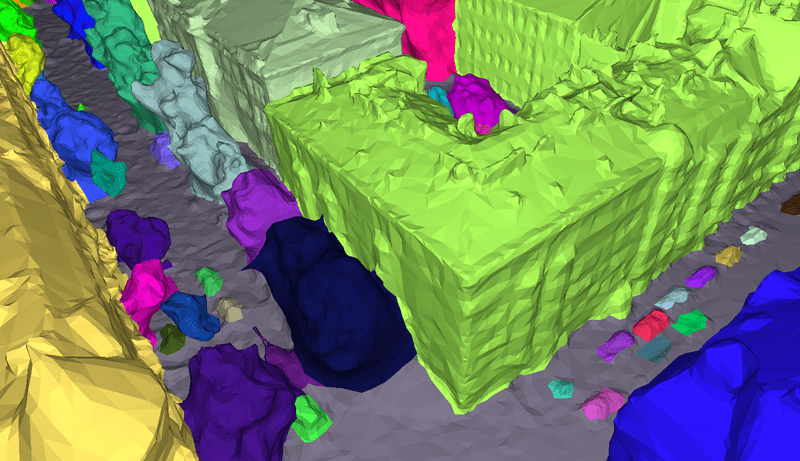}
		\caption{Truth. 371}
		\label{fig:adap_seg_truth_loc}
	\end{subfigure}
	\caption{Comparison of mesh over-segmentation methods on a tile of the SUM dataset~\cite{gao2021sum}. 
			(b) to (k) show the over-segmentation results with the same \textit{object purity} (around $92\%$) for all methods. The number below each result denotes the number of segments required to achieve the desired \textit{object purity}.
			(l) shows the connected components extracted from the ground truth semantic segmentation.
	}
	\label{fig:adpative_seg_loc}
\end{figure*}

We evaluate over-segmentation on SUM dataset~\cite{gao2021sum} because it covers a larger area and contains fewer unlabeled areas compared to H3D dataset~\cite{kolle2021h3d}.	
\Cref{fig:adpative_seg_loc} presents our planarity-sensible over-segmentation result and comparison with seven other commonly used over-segmentation techniques, namely region growing (RG)~\cite{lafarge2012creating}, efficient RANSAC (RA)~\cite{schnabel2007efficient}, geometric partition (GP)~\cite{landrieu2018large}, supervized superpoint generation (SSP)~\cite{landrieu2019point}, variational shape approximation (VSA)~\cite{cohen2004variational}, supervoxel generation (SPV)~\cite{lin2018toward}, voxel cloud connectivity segmentation (Vccs)~\cite{papon2013voxel}, superface clustering (SC)~\cite{verdie2015lod}, and superface partitioning (SP)~\cite{rouhani2017semantic}. 
RG, VSA, SC, SP, and our method use meshes as input, and the other methods (originally developed for point clouds) perform over-segmentation on points that we densely sampled (10 $pts/m^2$) from the input mesh. 
We can see from~\Cref{fig:adpative_seg_loc} that the segment boundaries of our method are largely aligned with object boundaries. 
RG and VSA perform similarly but generate excessive segments for non-planar objects such as trees. Our over-segmentation generates segments that are closer to semantically meaningful objects.
In terms of the number of segments, RA, SC, SP, Vccs, and SPV generate relatively large numbers of segments, which are unfavorable to the subsequent classification using GCN. 
In contrast, our method generates the least segments, which is a strong advantage for the subsequent classification step in terms of efficiency.

\begin{figure*}[!tb]
	\centering
	\includegraphics[width=0.7\linewidth]{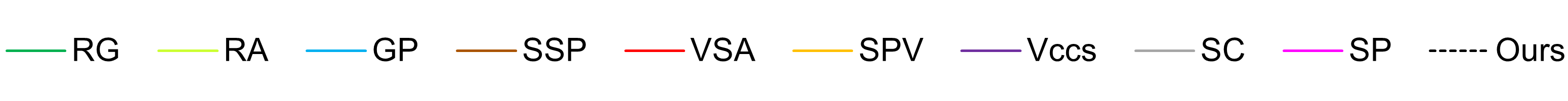}
	\begin{subfigure}{0.32\linewidth}
		\includegraphics[page=1,width=1.09\linewidth, height=0.13\textheight]{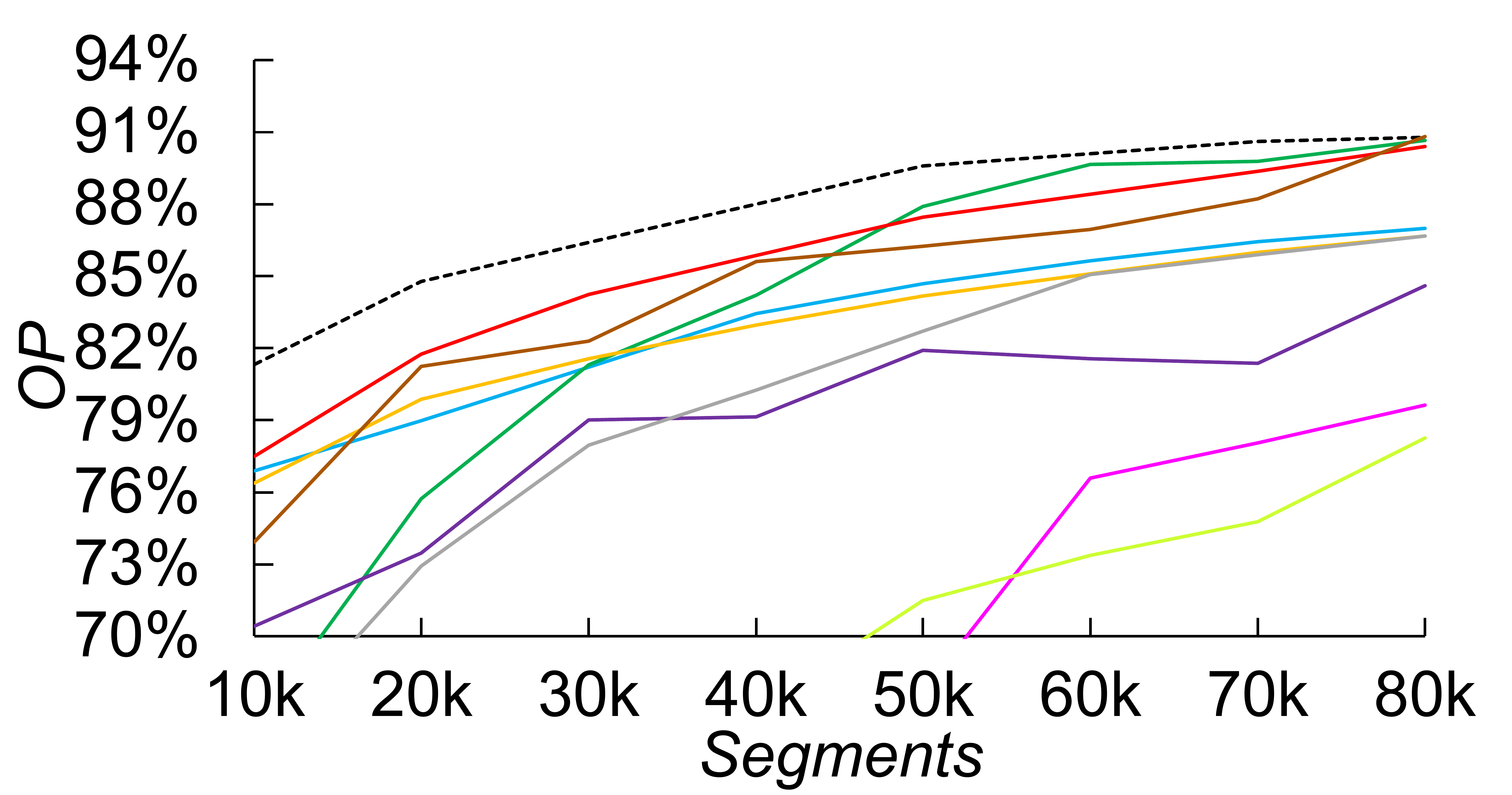}
		\caption{Object purity}
		\label{fig:asa}
	\end{subfigure}
	\begin{subfigure}{0.32\linewidth}
		\includegraphics[page=1,width=1.09\linewidth, height=0.13\textheight]{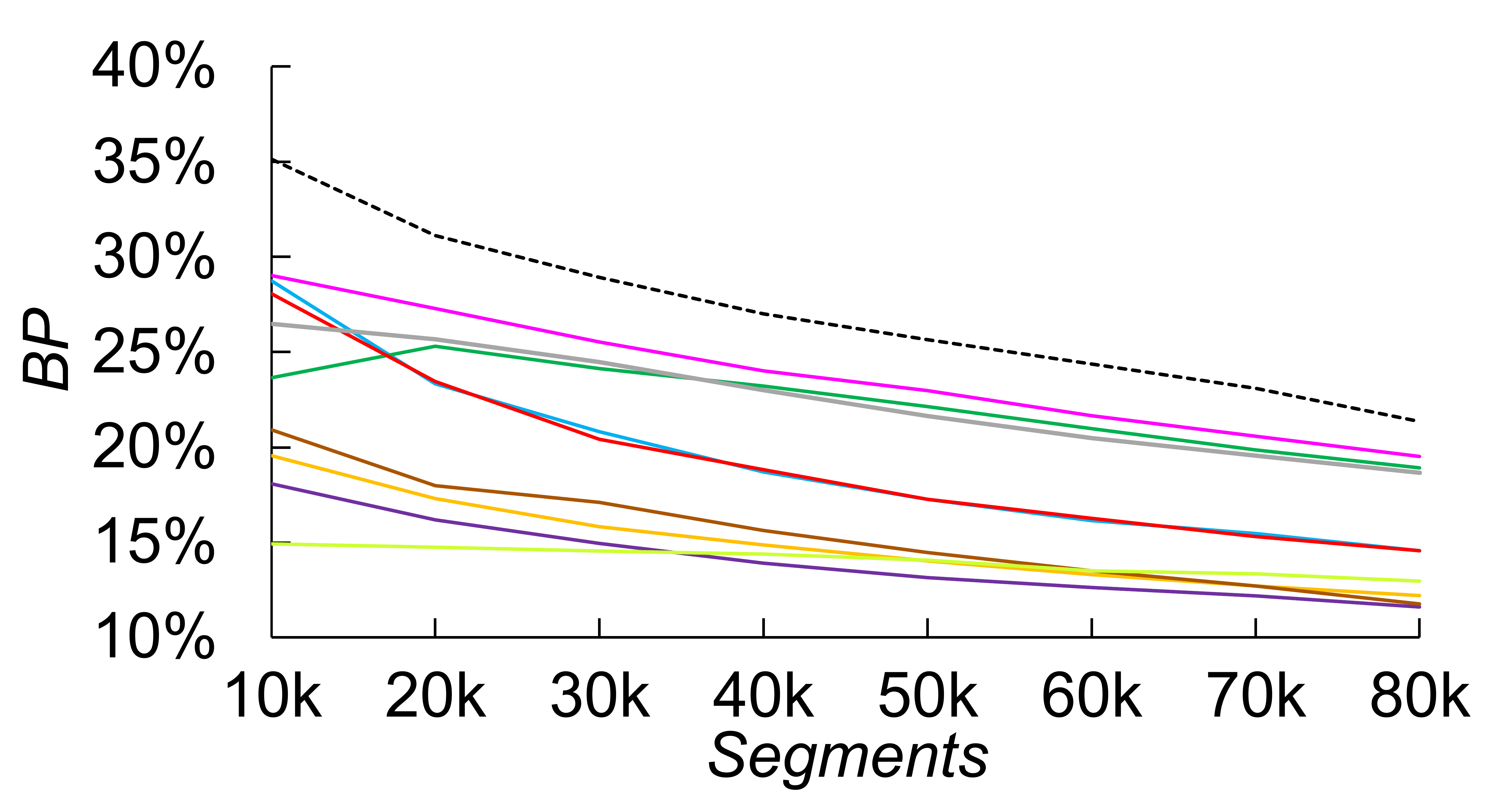}
		\caption{Boundary precision}
		\label{fig:bp}
	\end{subfigure}
	\begin{subfigure}{0.32\linewidth}		
		\includegraphics[page=1,width=1.09\linewidth, height=0.13\textheight]{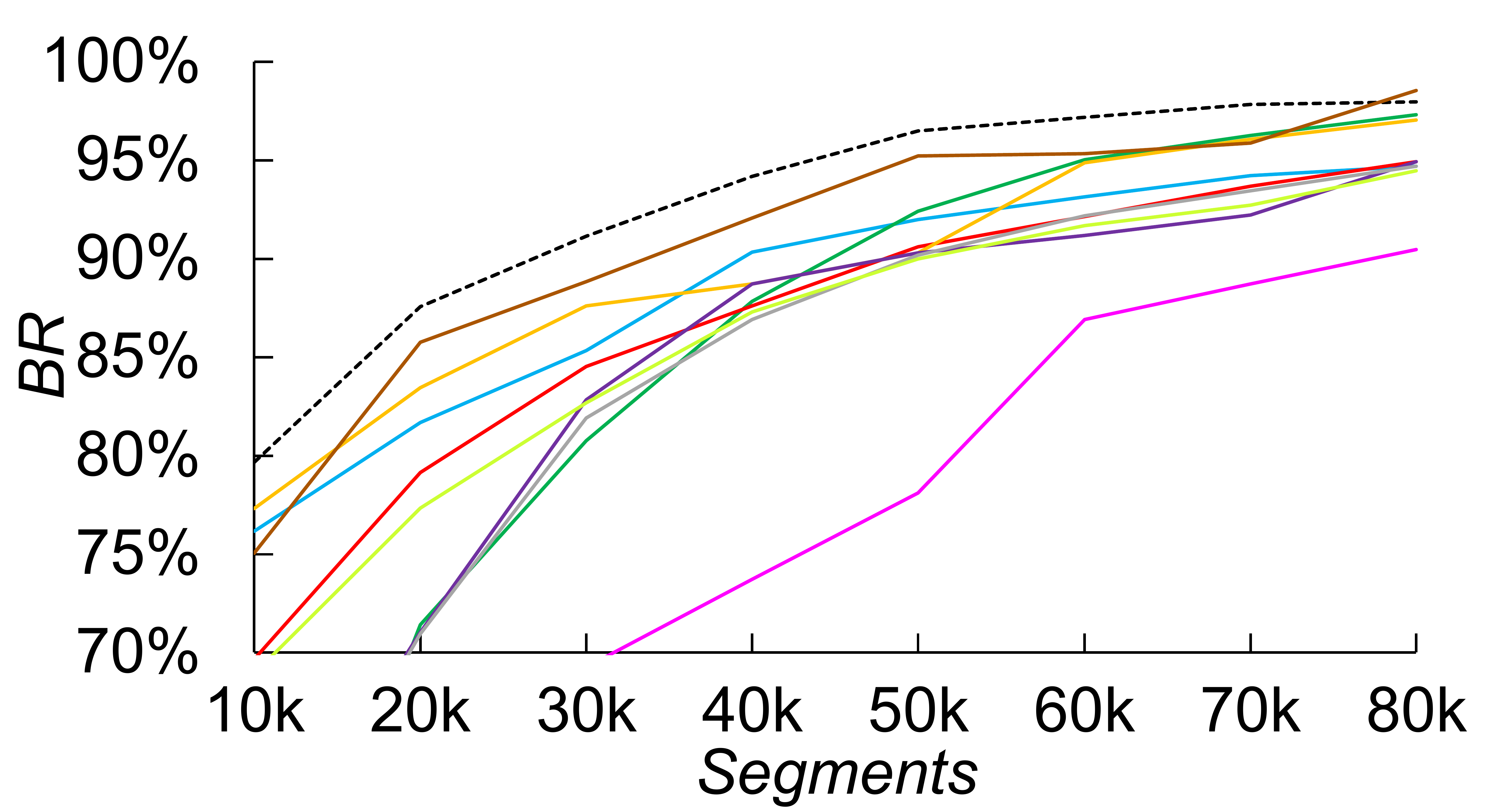}
		\caption{Boundary recall}
		\label{fig:br}
	\end{subfigure}
	\caption{Comparison of different over-segmentation method in terms of \textit{OP}, \textit{BP}, and \textit{BR} on the SUM dataset~\cite{gao2021sum}. 
	For each method, we tuned their parameters such that all methods generated a similar number of segments. The data was recorded at different numbers of segments for all methods.
	}
	\label{fig:evaluation_overseg}
\end{figure*}

We have used the entire SUM dataset~\cite{gao2021sum} to evaluate our over-segmentation method in terms \textit{OP}, \textit{BP}, and \textit{BR}, and we have compared them with those of the other seven over-segmentation methods. 
In the comparison, we tuned the parameters of each method such that all methods generated a similar number of segments, and we then computed the \textit{OP}, \textit{BP}, and \textit{BR} for each method. 
We recorded the performance of all methods for different numbers of segments, and the results are shown in~\Cref{fig:evaluation_overseg}.
It can be observed that our method outperforms the others for all three metrics. 
Specifically, as the number of segments increases, the \textit{OP} of VSA, RG, and SSP get closer to ours. However, VSA underperforms our method in terms of \textit{BP} and \textit{BR}, which indicates that our method generated segments with better boundary qualities.
For RG and SSP, its \textit{OP}, \textit{BP}, and \textit{BR} are rather low when the number of segments is small, indicating that our method is more robust with a relatively small number of segments.
Other methods like RA, SP, GP, SPV, Vccs, and SC also require a larger number of segments to produce satisfactory results.

To understand the potential of each method,~\Cref{tab:upper_bound} provides the maximum achievable performance of semantic segmentation for each over-segmentation method. The maximum achievable performance is measured by the maximum IoU and mIoU that can be achieved in theory.
We can see that our method significantly outperforms the other methods with a considerable margin ranging from $4.8\%$ to $30.1\%$ in terms of mIoU (which reflects the under-segmentation errors).
It is also worth noting that VSA has slightly better results on very few object classes (e.g., \textit{water} and \textit{boat}) while our method can better distinguish small non-planar objects such as vehicles which are very common in urban textured meshes.

\begin{figure*}[!tb]
	\centering
	\includegraphics[width=0.7\linewidth]{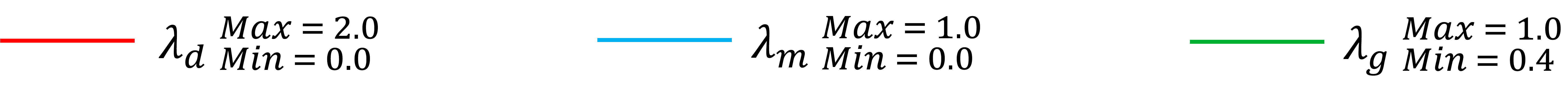}
	\begin{subfigure}{0.49\linewidth}
		\includegraphics[page=1,width=\linewidth]{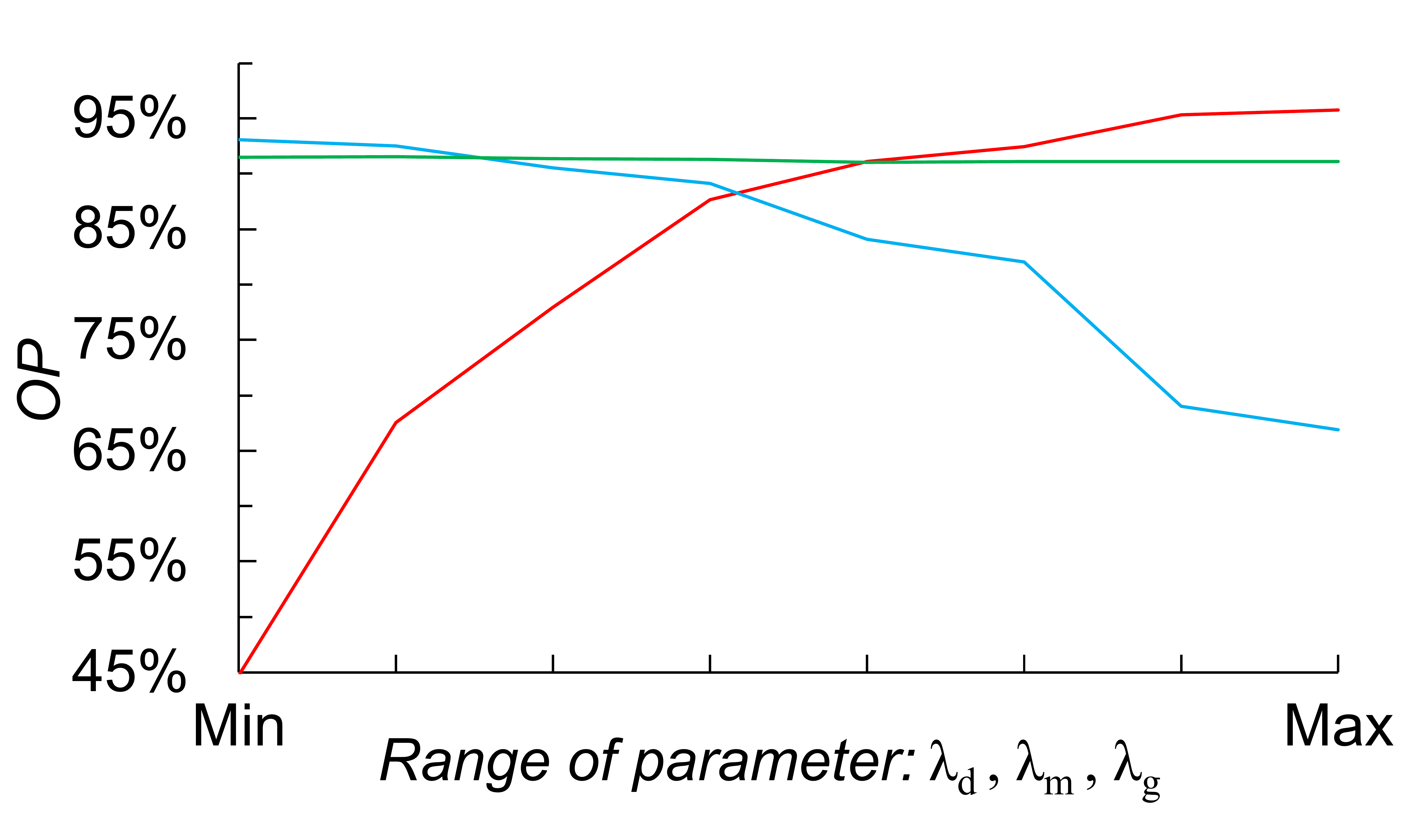}
		\caption{Object purity}
		\label{fig:op_params}
	\end{subfigure}
	\hfill
	\begin{subfigure}{0.49\linewidth}		
		\includegraphics[page=1,width=\linewidth]{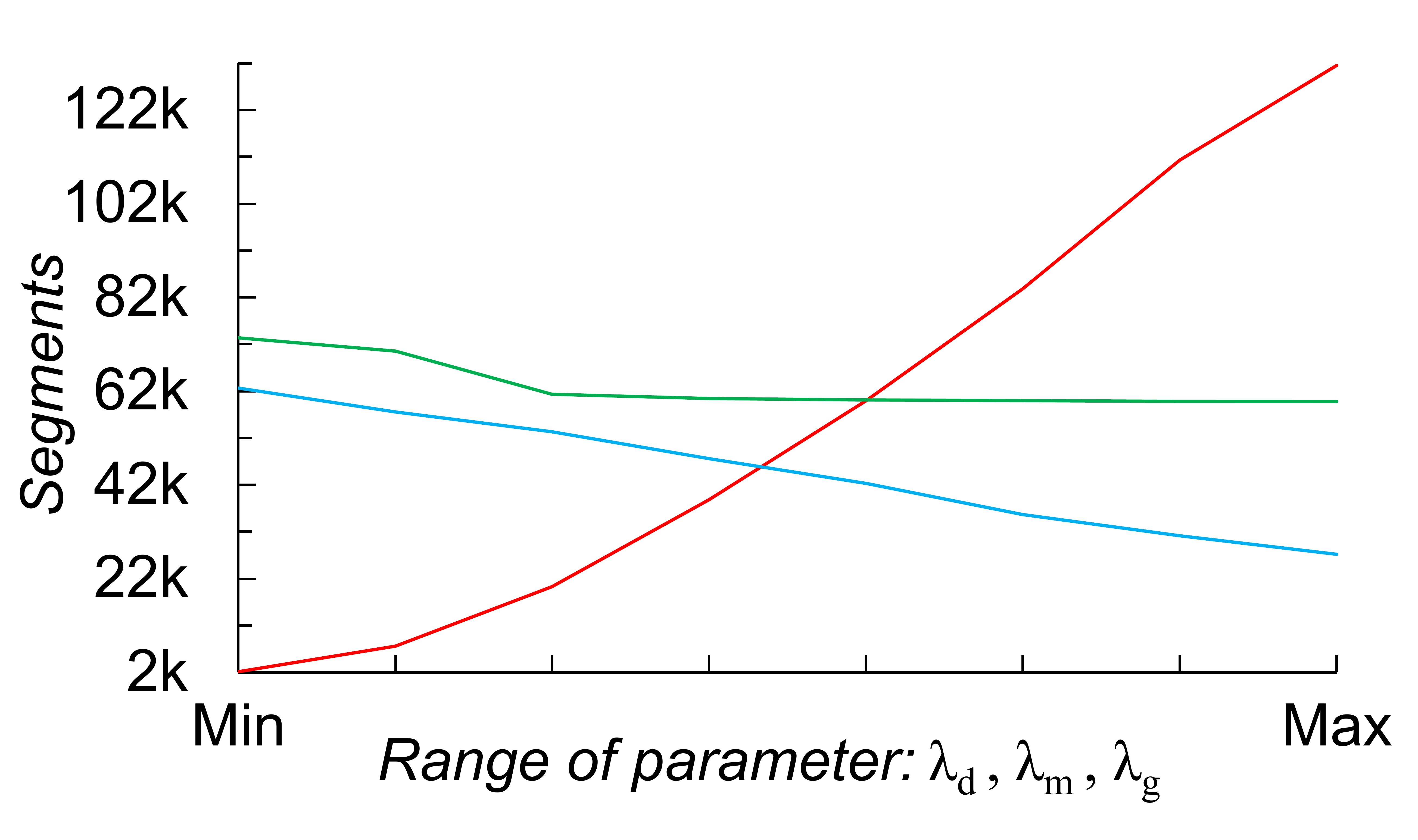}
		\caption{Number of Segments}
		\label{fig:segnum_params}
	\end{subfigure}
	\caption{Comparison of over-segmentation with different parameter configurations in terms of object purity and the number of segments on the SUM dataset~\cite{gao2021sum}. Note that the range of parameters depends on the quality of the input data.
	}
	\label{fig:evaluation_params}
\end{figure*}

\paragraph{Performance analysis}
For the impact of planarity-sensible over-segmentation in different configurations, we experimented with different settings for each weight term based on the default parameters (i.e., $\lambda_{d} = 1.2$,  $\lambda_{m} = 0.1$, and $\lambda_{g} = 0.9$).
In each experiment, only one weight is tuned while the others remain unchanged.
~\Cref{fig:evaluation_params} shows its results in terms of \textit{OP} and the number of \textit{segments}.
We can observe that increasing $\lambda_{d}$ leads to a larger \textit{OP} but also encourages the splitting of segments into smaller planar ones. 
In contrast, increasing $\lambda_{m}$ results in over-smoothed segments (i.e., the smaller segments are merged into a larger one). 
However, $\lambda_{g}$ performs differently from the other two since our aim is to use $\lambda_{g}$ to reduce the number of \textit{segments} without decreasing \textit{OP} (as demonstrated in~\Cref{fig:evaluation_params}).
The performance of applying $\lambda_{g}$ is limited to the results of \textit{planar} and \textit{non-planar} classification, while the performance of the other two terms is limited to the quality of the mesh.

\begin{table}[!tb]
	\centering
	\noindent\adjustbox{max width=1.0\linewidth}
	{
		\begin{tabular}{l|cccccc|c}
			\hline
			Methods & Terra. & H-veg. & Build. & Water & Vehic. & Boat & mIoU \\
			\hline
			SP~\cite{rouhani2017semantic} & 73.4 & 88.1 & 96.8 & 12.5 & 15.7 & 68.1 & 59.1 \\
			RANSAC~\cite{schnabel2007efficient} & 82.3 & 88.6 & 97.0 & 57.3 & 29.8 & 69.7 & 70.8 \\
			Vccs~\cite{papon2013voxel} & 77.4 & 86.3 & 94.3 & 87.4 & 35.9 & 84.5 & 77.6 \\
			SC~\cite{verdie2015lod} & 86.8 & 92.5 & 97.8 & 75.5 & 46.1 & 79.3 & 79.7 \\
			SPV~\cite{lin2018toward} & 83.7 & 91.5 & 97.0 & 87.1 & 37.5 & 84.5 & 80.2 \\
			GP~\cite{landrieu2018large} & 86.5 & 91.2 & 96.6 & 87.1 & 46.9 & 84.6 & 82.1 \\
			RG~\cite{lafarge2012creating} & 90.9 & 93.9 & 98.4 & 84.5 & 53.9 & 75.6 & 82.9 \\
			SSP~\cite{landrieu2019point} & 85.3 & 91.2 & 96.1 & 91.1 & 49.6 & 88.3  & 83.6 \\
			VSA~\cite{cohen2004variational} & 91.1 & 95.1 & 98.7 & \textbf{93.4} & 39.3 & \textbf{88.9} & 84.4 \\
			Ours & \textbf{93.3} & \textbf{95.6} & \textbf{98.9} & 91.6 & \textbf{67.1} & 88.8 & \textbf{89.2} \\
			\hline
		\end{tabular}
	}
	\caption{Comparison of different over-segmentation methods in terms of maximum achievable performance of semantic segmentation on the test data from the SUM dataset~\cite{gao2021sum}, with 50,000 segments. The evaluation metric is reported in per-class IoU (\%) and mean IoU (mIoU, \%).}
	\label{tab:upper_bound}
\end{table}

\subsection{Evaluation of semantic classification}

\begin{figure*}[!th]
	\centering
	\begin{subfigure}[t]{0.18\linewidth}
		\includegraphics[width=\linewidth]{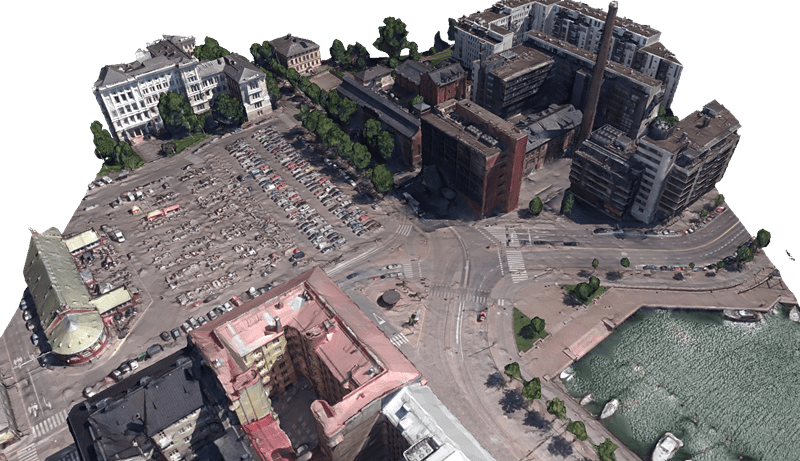}\vspace{0.1em}
		\includegraphics[width=\linewidth]{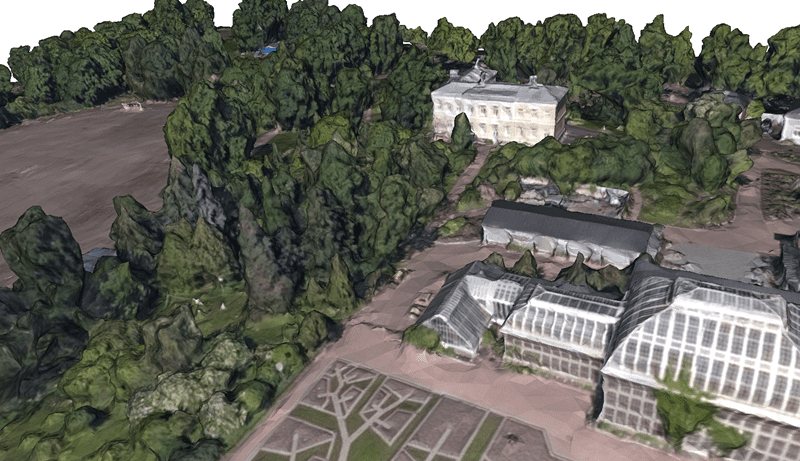}
		\caption{Input}
		\label{fig:tex_mesh}
	\end{subfigure}
	\hfill
	\begin{subfigure}[t]{0.18\linewidth}		
		\includegraphics[width=\linewidth]{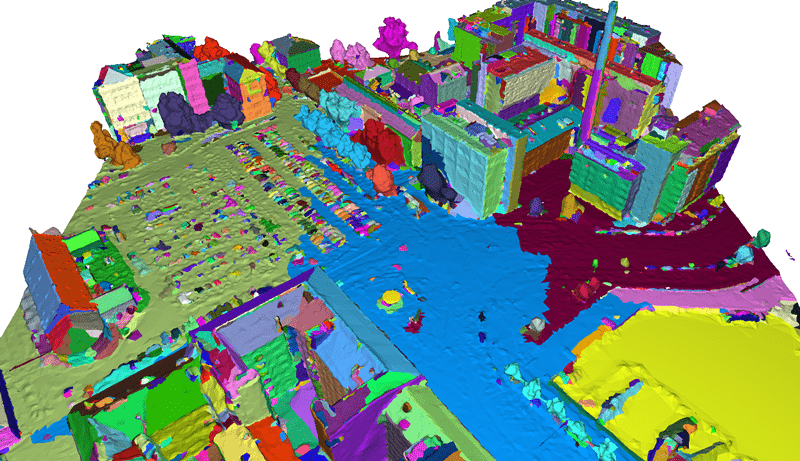}\vspace{0.1em}
		\includegraphics[width=\linewidth]{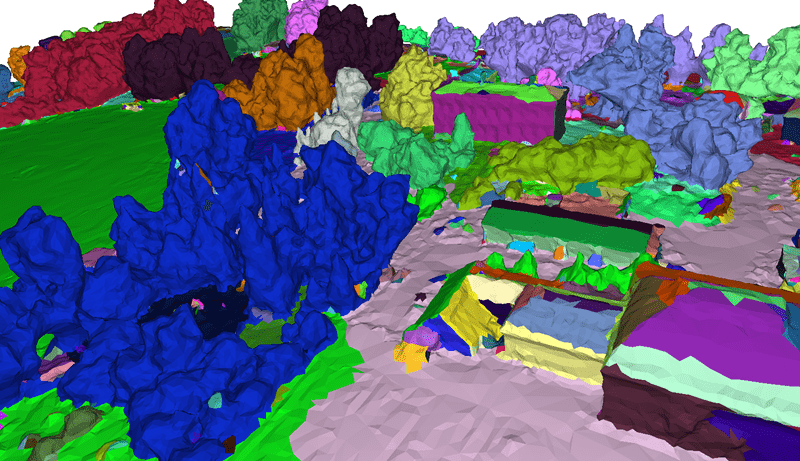}
		\caption{Segments}
		\label{fig:segments}
	\end{subfigure}
	\hfill
	\begin{subfigure}[t]{0.18\linewidth}
		\includegraphics[width=\linewidth]{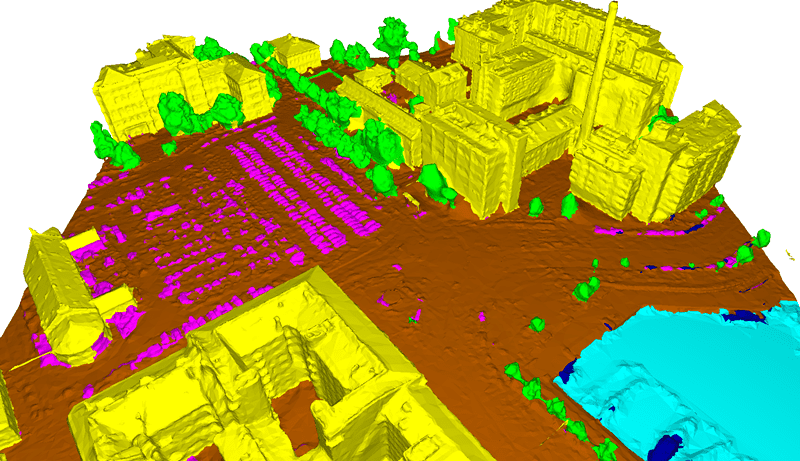}\vspace{0.1em}
		\includegraphics[width=\linewidth]{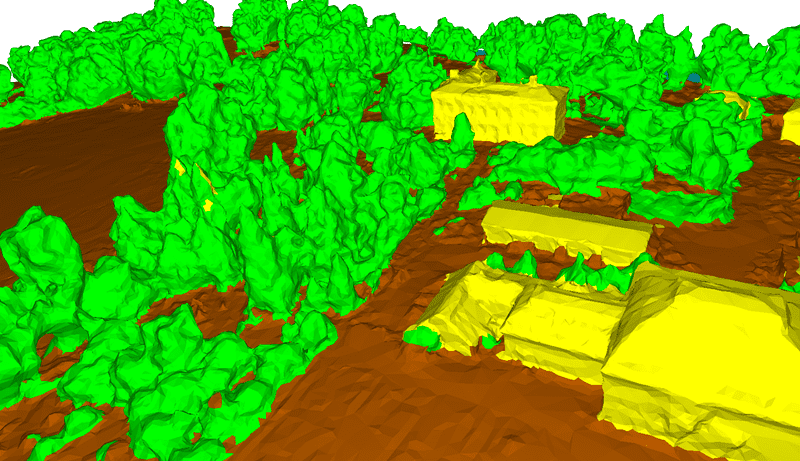}
		\caption{Semantic}
		\label{fig:predict}
	\end{subfigure}
	\hfill
	\begin{subfigure}[t]{0.18\linewidth}
		\includegraphics[width=\linewidth]{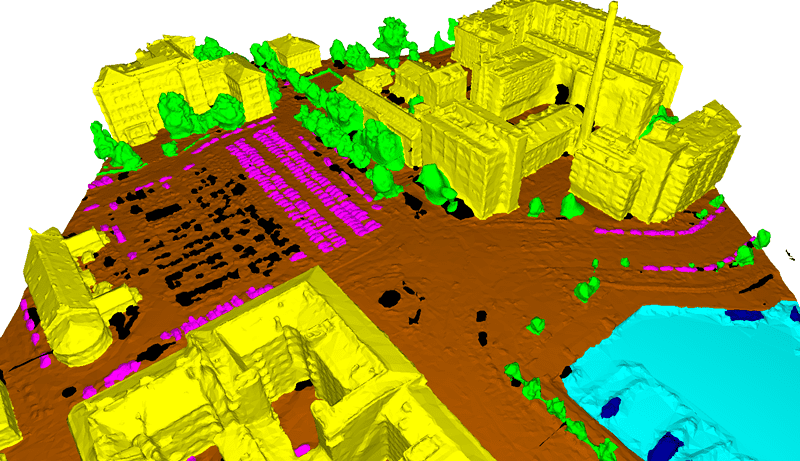}\vspace{0.1em}
		\includegraphics[width=\linewidth]{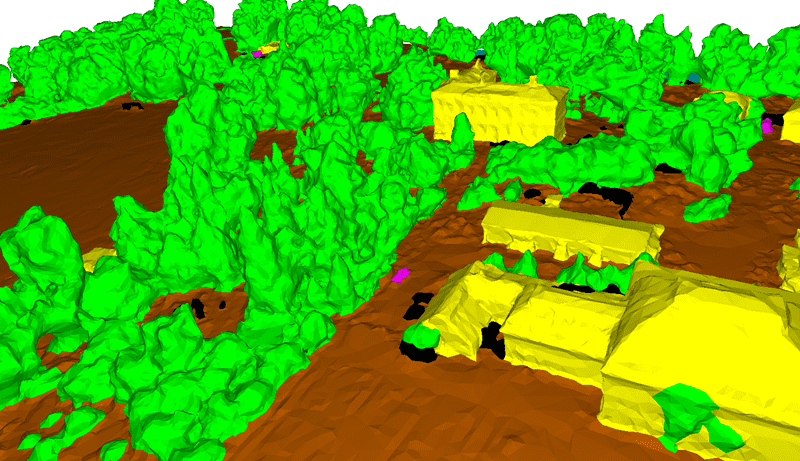}
		\caption{Truth}
		\label{fig:ground_truth_final}
	\end{subfigure}
	\hfill
	\begin{subfigure}[t]{0.18\linewidth}
		\includegraphics[width=\linewidth]{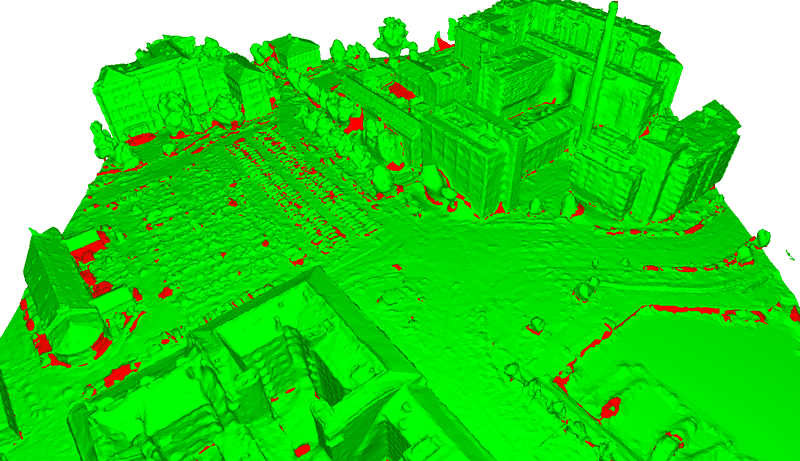}\vspace{0.1em}
		\includegraphics[width=\linewidth]{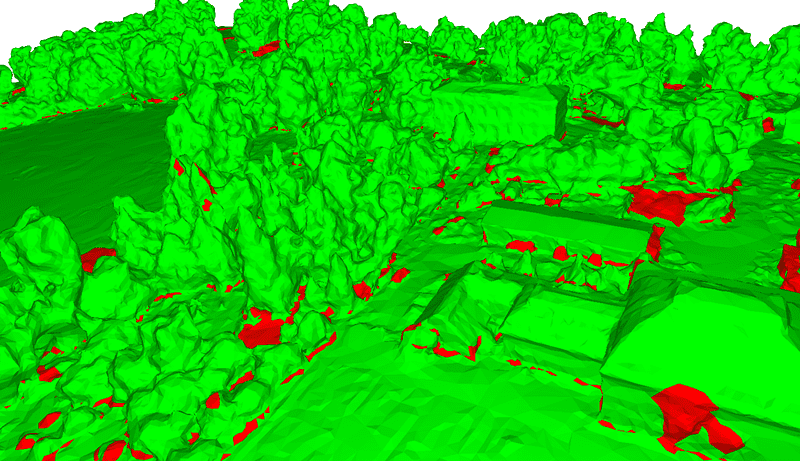}
		\caption{Error (red)}
		\label{fig:error_maps_final}
	\end{subfigure}
	\caption{Semantic segmentation results of our method on two tiles from the SUM dataset~\cite{gao2021sum}.}
	\label{fig:semantic_qualitive1}
\end{figure*}

\begin{figure}[htbp] %
\begin{adjustwidth}{-3cm}{-3cm}
	\centering
	\begin{tabular}{ccccc}	
		\includegraphics[width=0.25\textwidth]{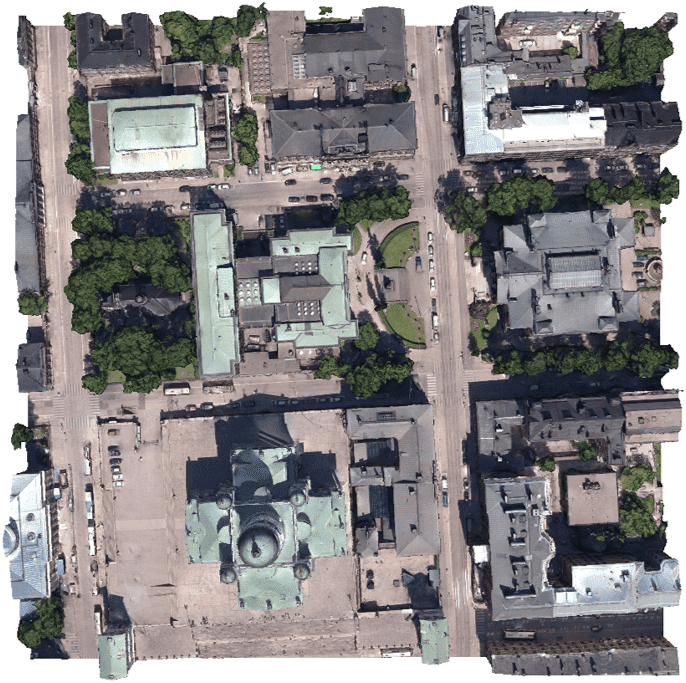}&
		\includegraphics[width=0.25\textwidth]{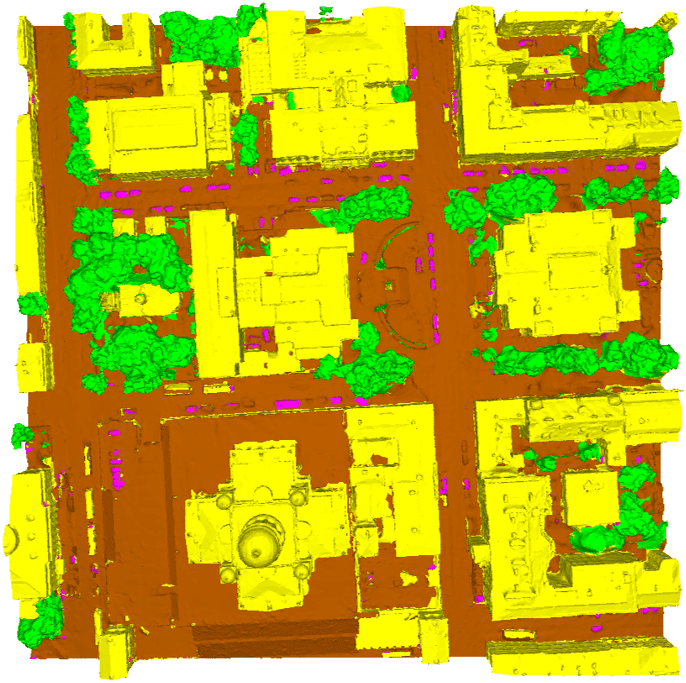}&
		\includegraphics[width=0.25\textwidth]{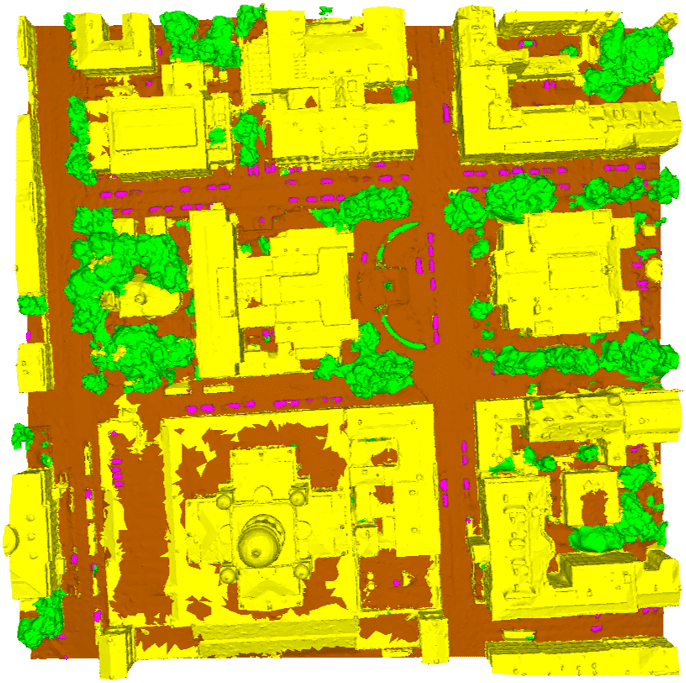}&
		\includegraphics[width=0.25\textwidth]{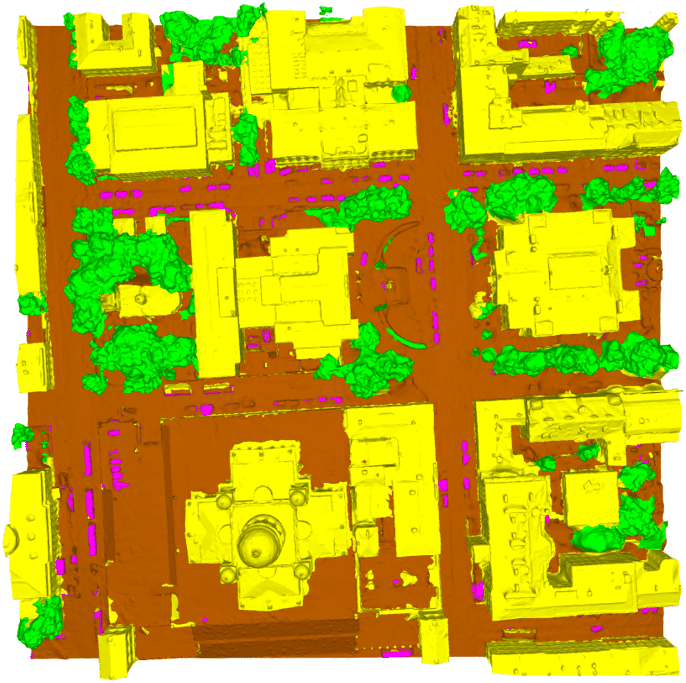}&
		\includegraphics[width=0.25\textwidth]{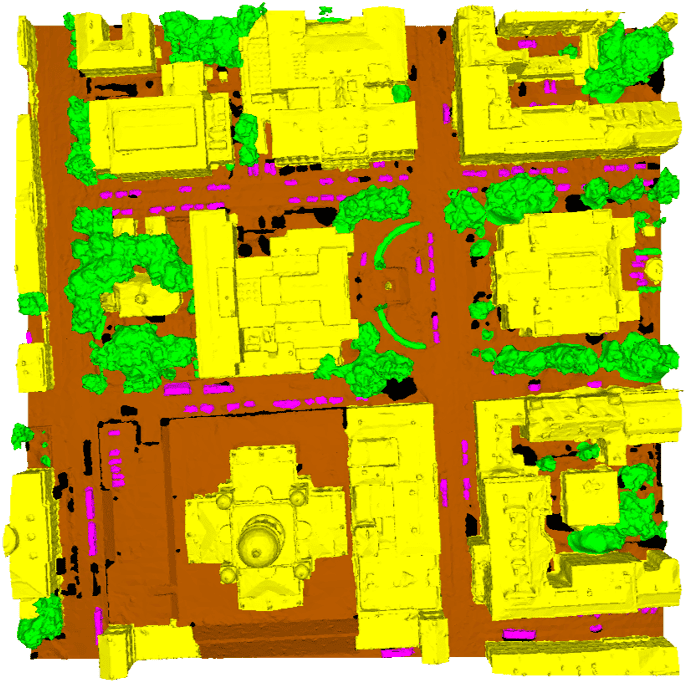}\\
		\midrule
		\includegraphics[width=0.25\textwidth]{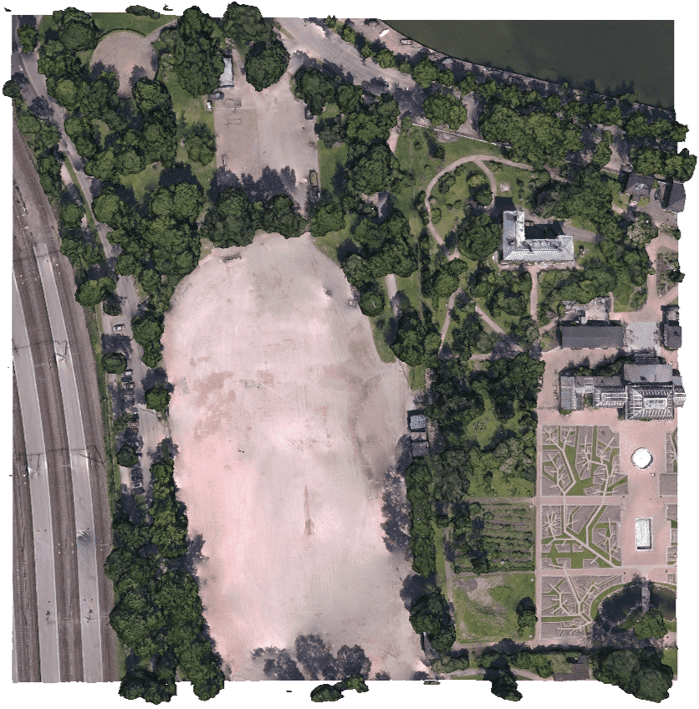}&
		\includegraphics[width=0.25\textwidth]{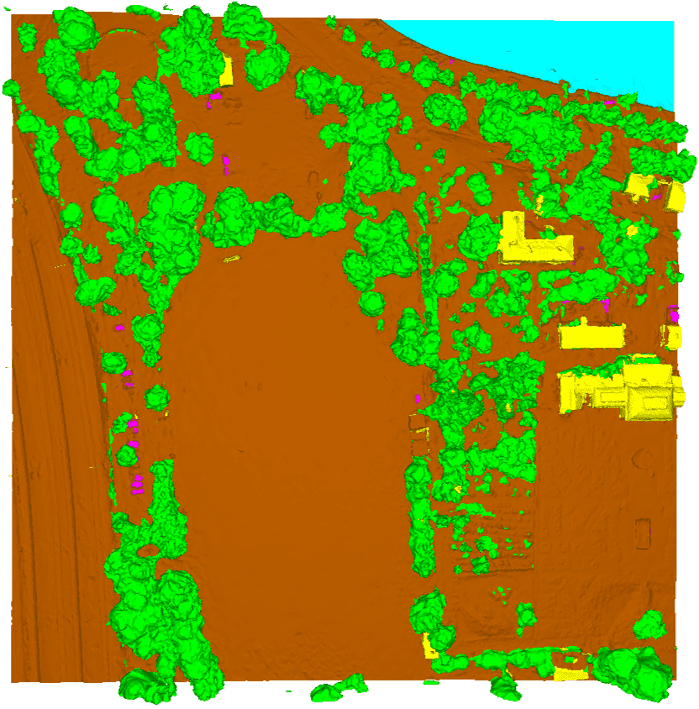}&
		\includegraphics[width=0.25\textwidth]{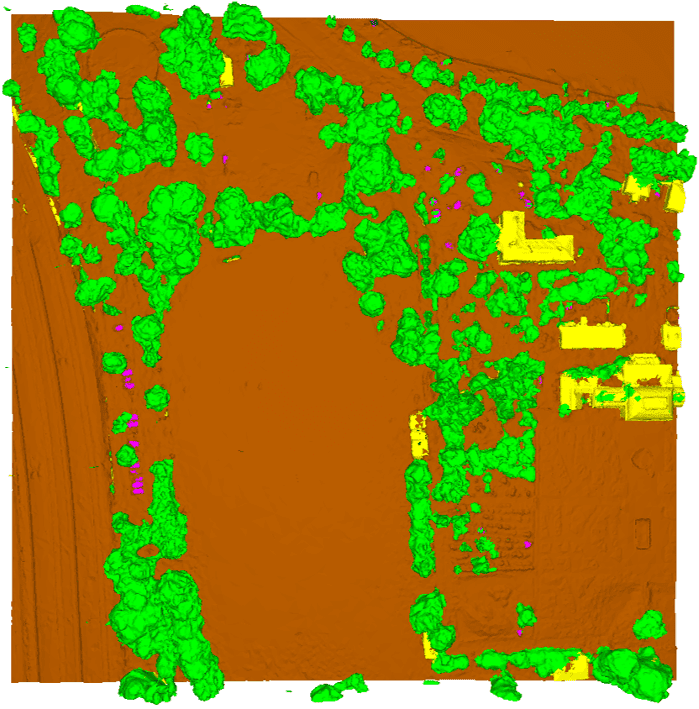}&
		\includegraphics[width=0.25\textwidth]{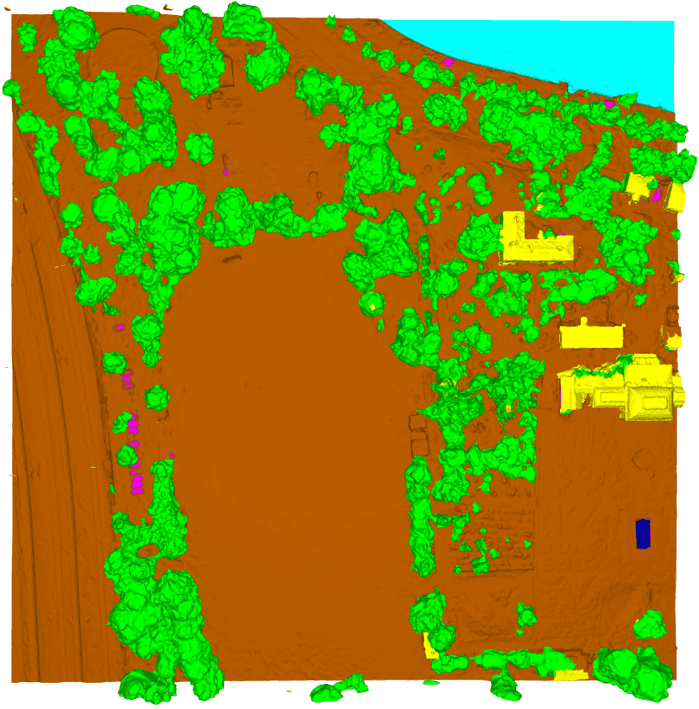}&
		\includegraphics[width=0.25\textwidth]{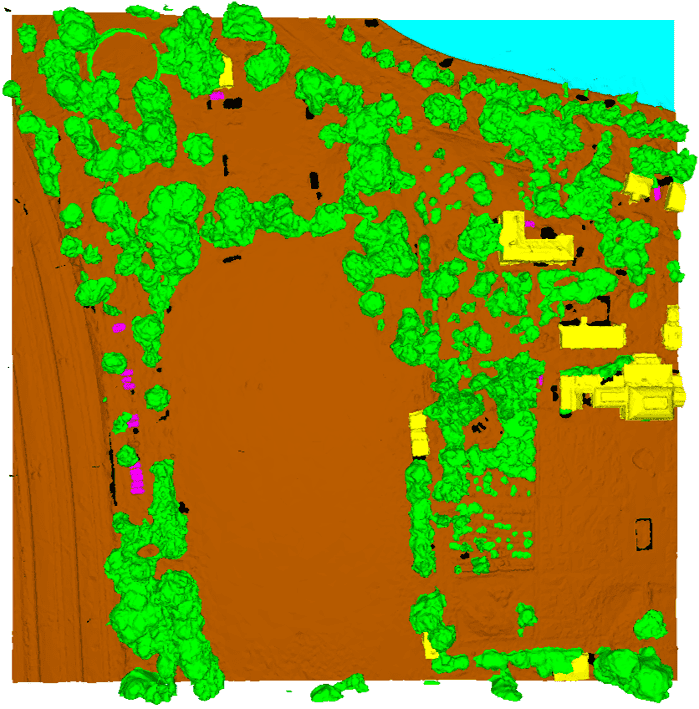}\\
		\midrule
		\includegraphics[width=0.25\textwidth]{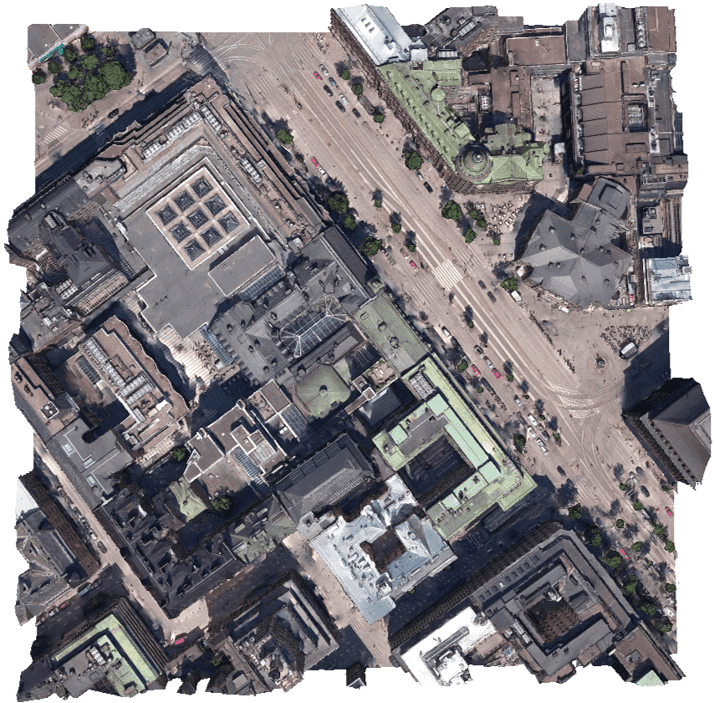}&
		\includegraphics[width=0.25\textwidth]{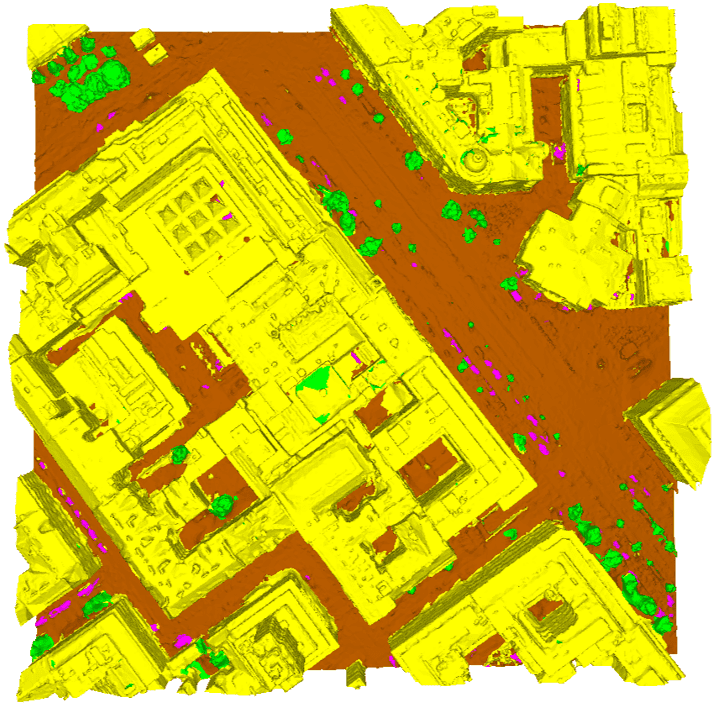}&
		\includegraphics[width=0.25\textwidth]{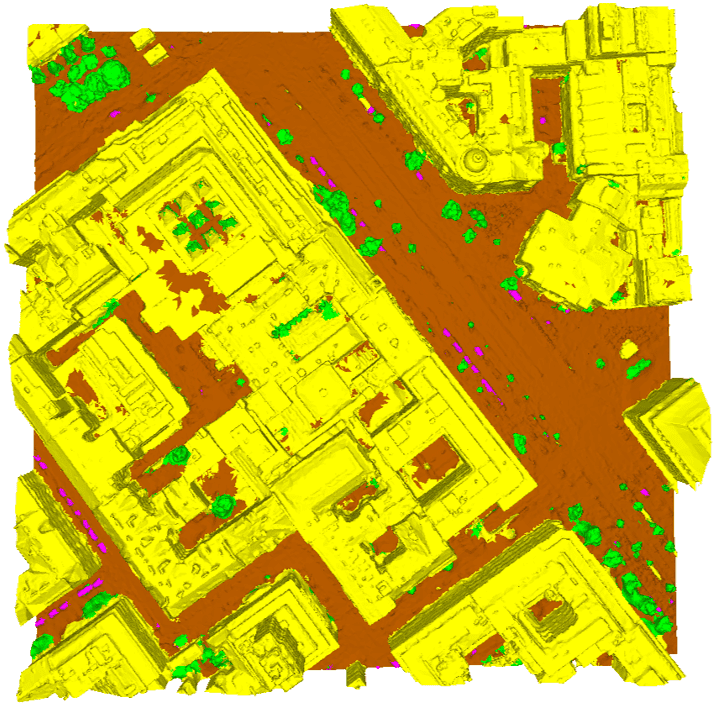}&
		\includegraphics[width=0.25\textwidth]{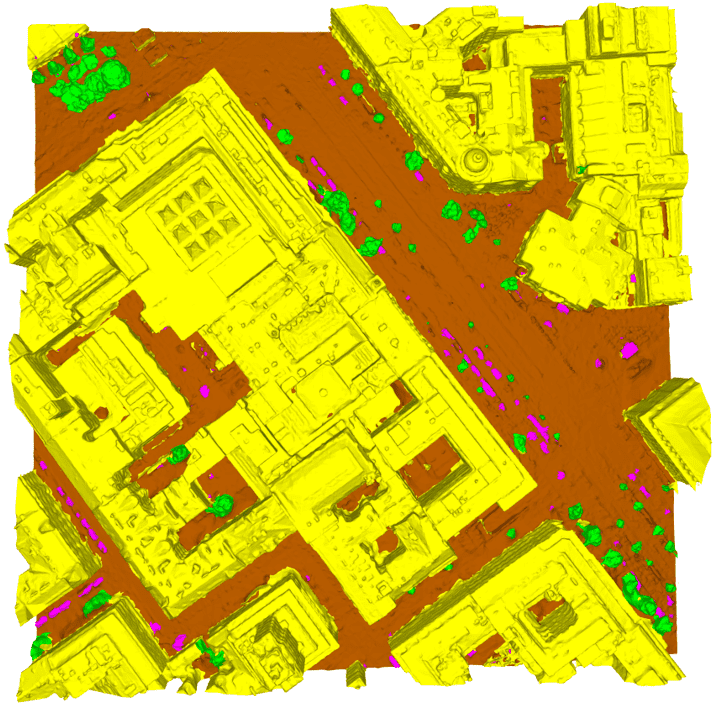}&
		\includegraphics[width=0.25\textwidth]{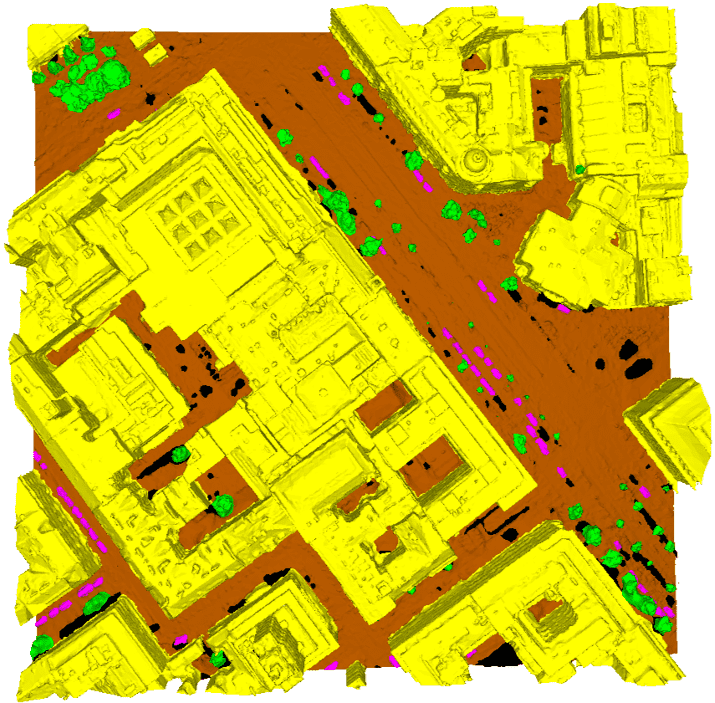}\\
		\midrule
		\includegraphics[width=0.25\textwidth]{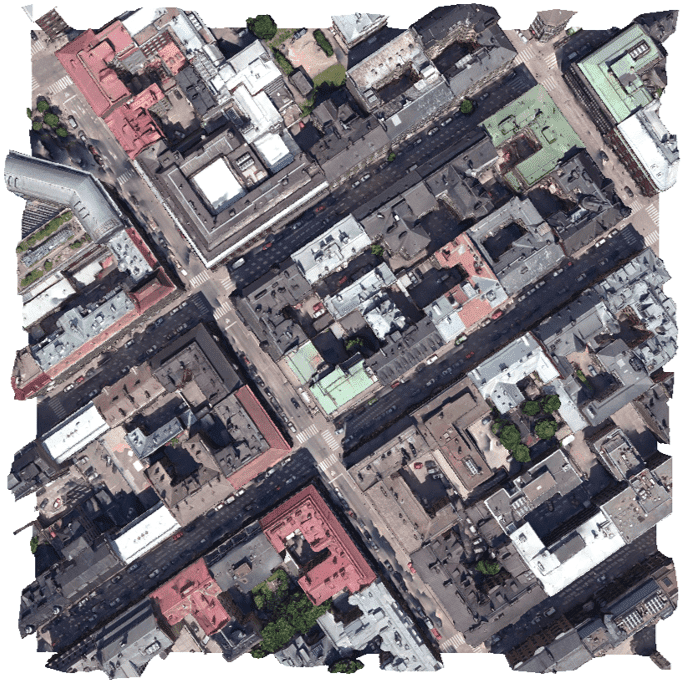}&
		\includegraphics[width=0.25\textwidth]{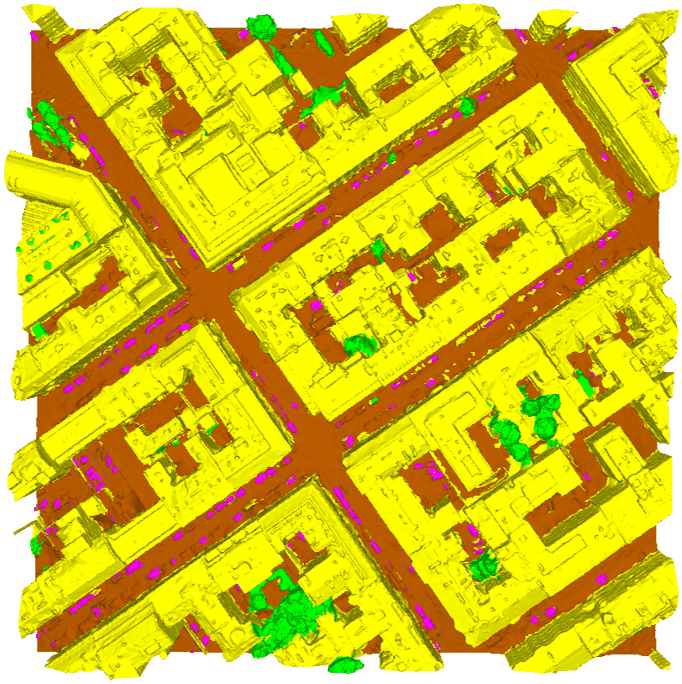}&
		\includegraphics[width=0.25\textwidth]{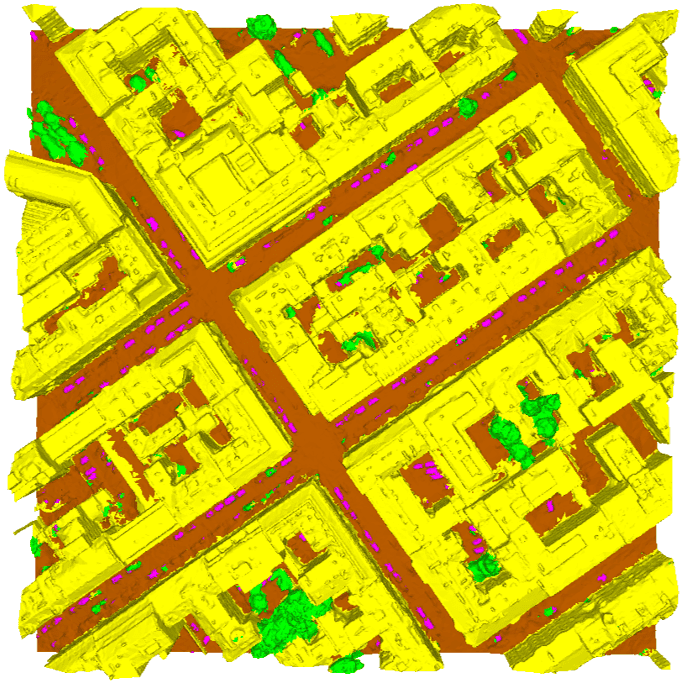}&
		\includegraphics[width=0.25\textwidth]{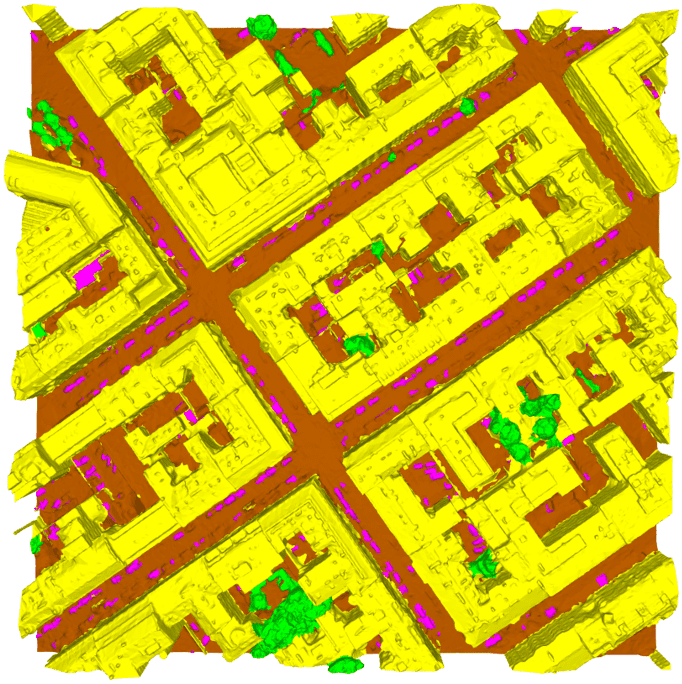}&
		\includegraphics[width=0.25\textwidth]{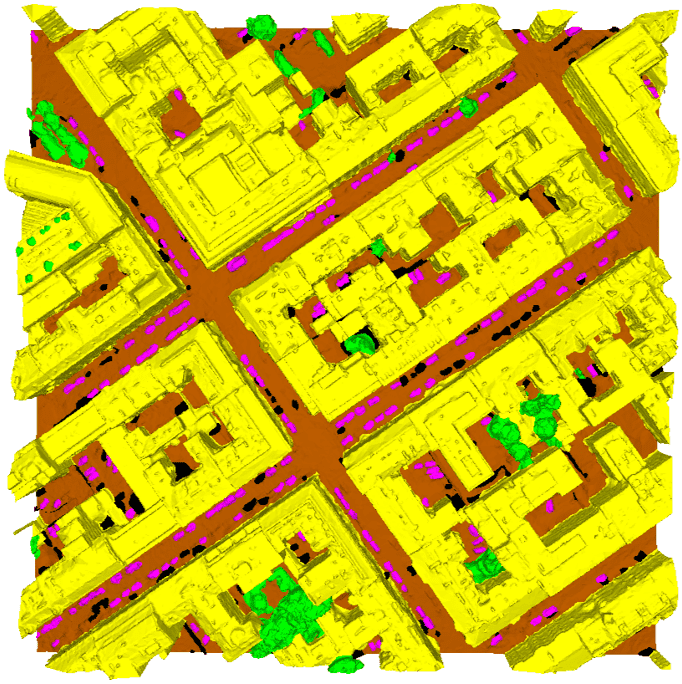}\\		
		\midrule
		\includegraphics[width=0.25\textwidth]{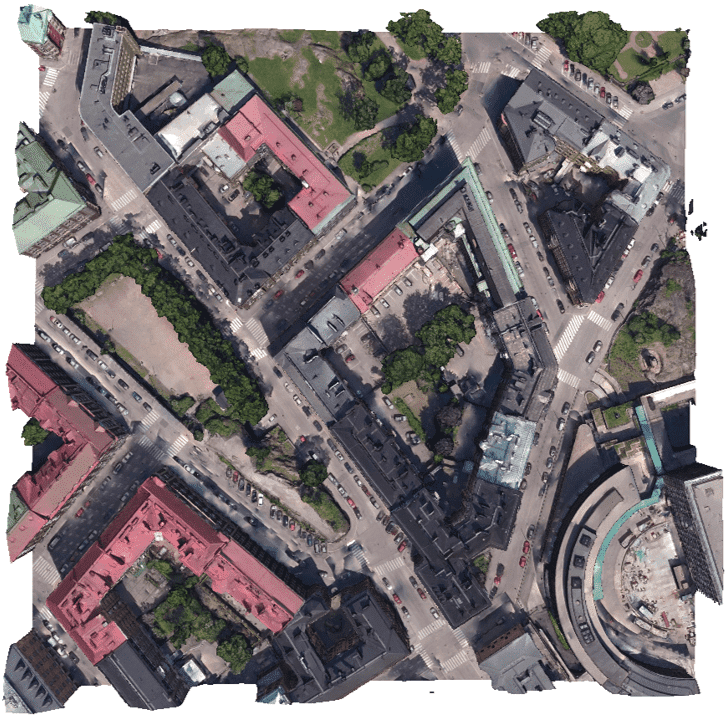}&
		\includegraphics[width=0.25\textwidth]{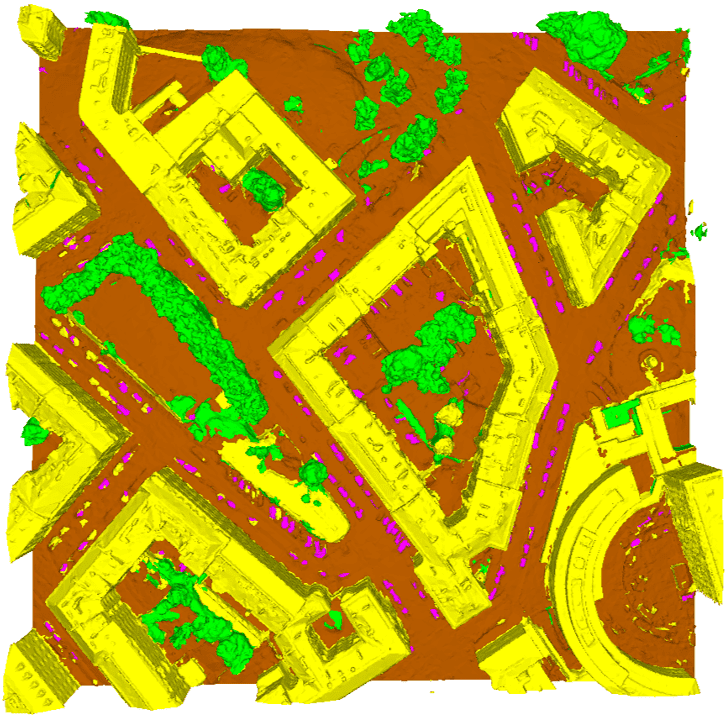}&
		\includegraphics[width=0.25\textwidth]{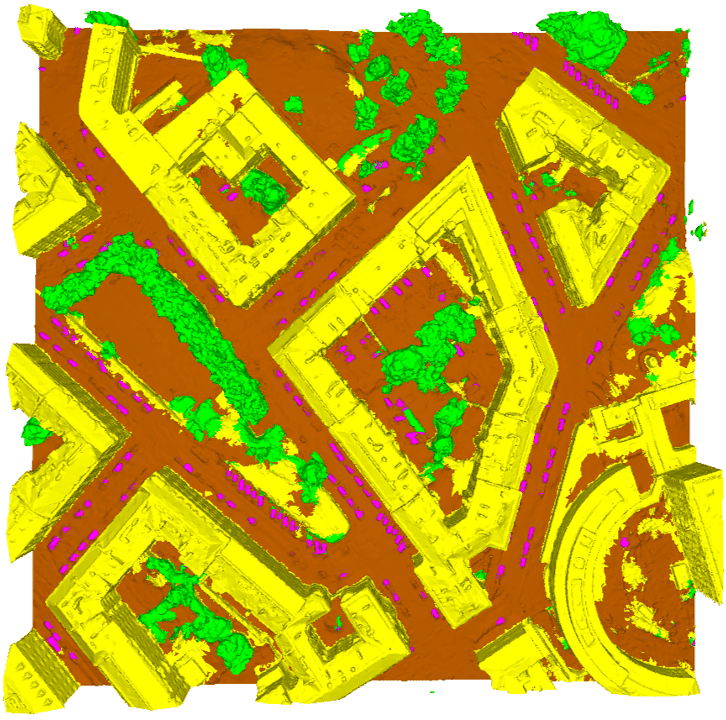}&
		\includegraphics[width=0.25\textwidth]{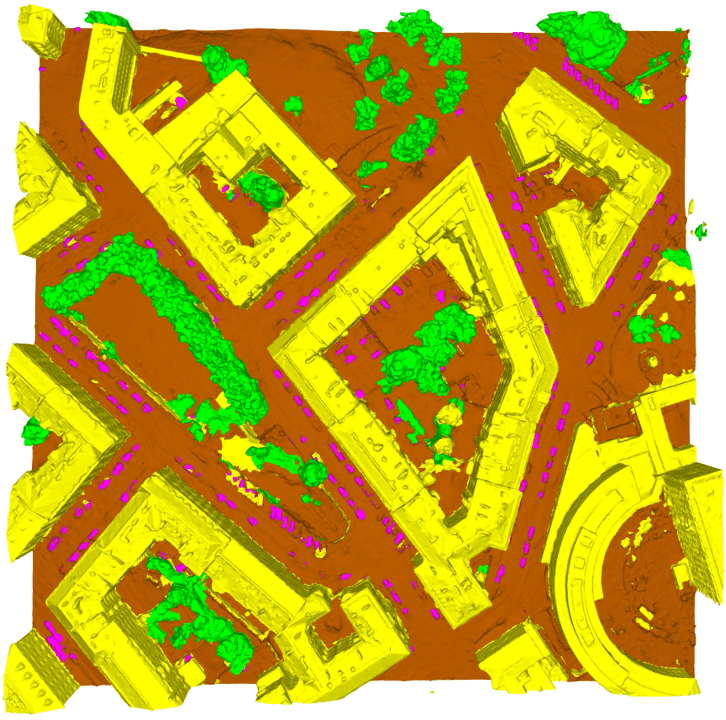}&
		\includegraphics[width=0.25\textwidth]{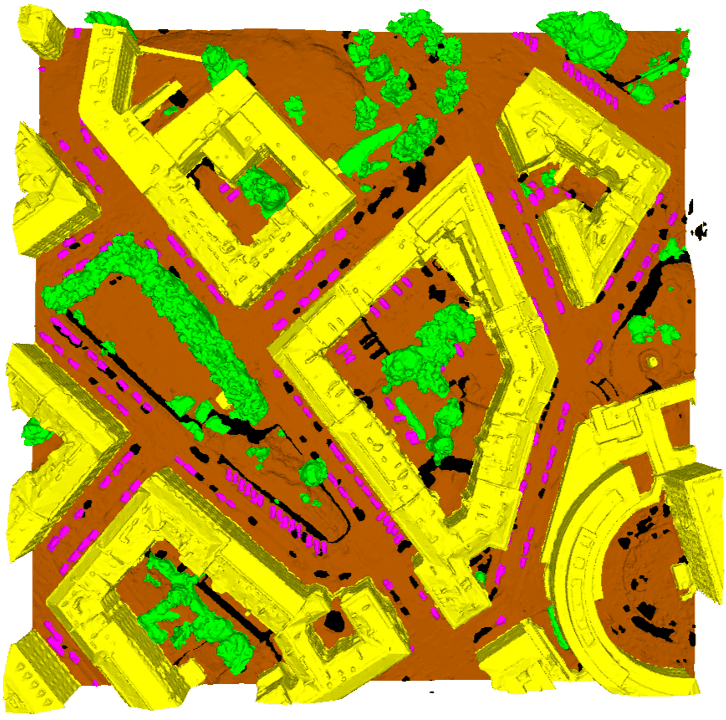}\\	
		(a) Original&
		(b) SUM-RF~\cite{gao2021sum}&
		(c) KPConv~\cite{thomas2019kpconv}&
		(d) Ours&
		(e) Truth\\	
	\end{tabular}
\end{adjustwidth}
\includegraphics[width=\textwidth]{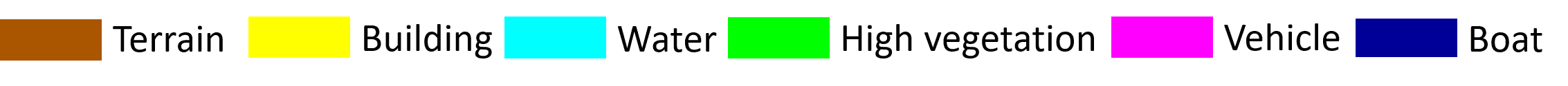}
\caption{Semantic segmentation results of SUM-RF~\cite{gao2021sum}, KPConv~\cite{thomas2019kpconv}, and our method on five tiles from the SUM dataset~\cite{gao2021sum}. }
\label{fig:semantic_qualitive2}
\end{figure}

\begin{table*}[!th]
	\centering
	\noindent\adjustbox{max width=1.0\linewidth}
	{
		\begin{tabular}{l|cccccc|cccc|cc}
			\hline
			Methods & Terra. & H-veg. & Build. & Water & Vehic. & Boat & mIoU & OA & mAcc & mF1 & \begin{tabular}[c]{@{}c@{}}Training\\($h$)\end{tabular} & \begin{tabular}[c]{@{}c@{}} Testing \\($min$)\end{tabular} \\
			\hline
			PointNet~\cite{qi2017pointnet} & 56.3 & 14.9   & 66.7 & 83.8 & 0.0 & 0.0 & 36.9 ± 2.3 & 71.4 ± 2.1 & 46.1 ± 2.6 & 44.6 ± 3.2 & 1.8 & \textbf{1}\\
			RandLaNet~\cite{hu2020randla} & 38.9 & 59.6   & 81.5 & 27.7 & 22.0 & 2.1 & 38.6 ± 4.6 & 74.9 ± 3.2 & 53.3 ± 5.1 & 49.9 ± 4.8 & 10.8 & 52\\
			SPG~\cite{landrieu2018large} & 56.4 & 61.8 & 87.4   & 36.5 & 34.4 & 6.2 & 47.1 ± 2.4  & 79.0 ± 2.8 & 64.8 ± 1.2 & 59.6 ± 1.9 & 17.8 & 26\\
			PointNet++~\cite{qi2017pointnet++} & 68.0 & 73.1   & 84.2 & 69.9 & 0.5 & 1.6 & 49.5 ± 2.1 & 85.5 ± 0.9 & 57.8 ± 1.8 & 57.1 ± 1.7 & 2.8 & 3\\
			RF-MRF~\cite{rouhani2017semantic} & 77.4 & 87.5 & 91.3 & 83.7 & 23.8 & 1.7 & 60.9 ± 0.0 & 91.2 ± 0.0 & 65.9 ± 0.0 & 68.1 ± 0.0 & \textbf{1.1} & 15\\
			SUM-RF~\cite{gao2021sum}  & 83.3 & 90.5 & 92.5 & \textbf{86.0} & 37.3 & 7.4 & 66.2 ± 0.0 & 93.0 ± 0.0 & 70.6 ± 0.0 & 73.8 ± 0.0 & 1.2 & 18\\
			KPConv~\cite{thomas2019kpconv} & \textbf{86.5} & 88.4 & 92.7 & 77.7 & \textbf{54.3} & 13.3 & 68.8 ± 5.7 & 93.3 ± 1.5 & 73.7 ± 5.4 & 76.7 ± 5.8 & 23.5 & 42\\
			Ours  & 84.9 & \textbf{90.6} & \textbf{93.9} & 84.3 & 50.9 & \textbf{32.3} & \textbf{72.8} ± 2.0 & \textbf{93.8} ± 0.4 & \textbf{79.2} ± 3.0 & \textbf{81.6} ± 2.3 & 16.3 & 62\\
			\hline
		\end{tabular}
	}
	\caption{Comparison with state-of-the-art semantic segmentation methods on the SUM dataset~\cite{gao2021sum}.
	Per-class IoU (\%), mean IoU (mIoU, \%), Overall Accuracy (OA, \%), mean class Accuracy (mAcc, \%), mean F1 score (mF1, \%), and the running times for training (over-seg: 1.8 hours, graph: 2.7 hours, classification: 11.8 hours, total: 16.3 hours) and testing (over-seg: 15 minutes, graph: 40 minutes, classification: 7 minutes, total: 62 minutes) are included. Note that each method was run ten times and the mean performance is reported here.}
	\label{tab:semanticcomp}
\end{table*}

We have tested our semantic classification method on both the SUM~\cite{gao2021sum} and the H3D~\cite{kolle2021h3d} datasets.

\subsubsection{Evaluation on SUM}

\begin{figure}[!th]
	\begin{subfigure}{0.48\linewidth}
		\includegraphics[width=\linewidth]{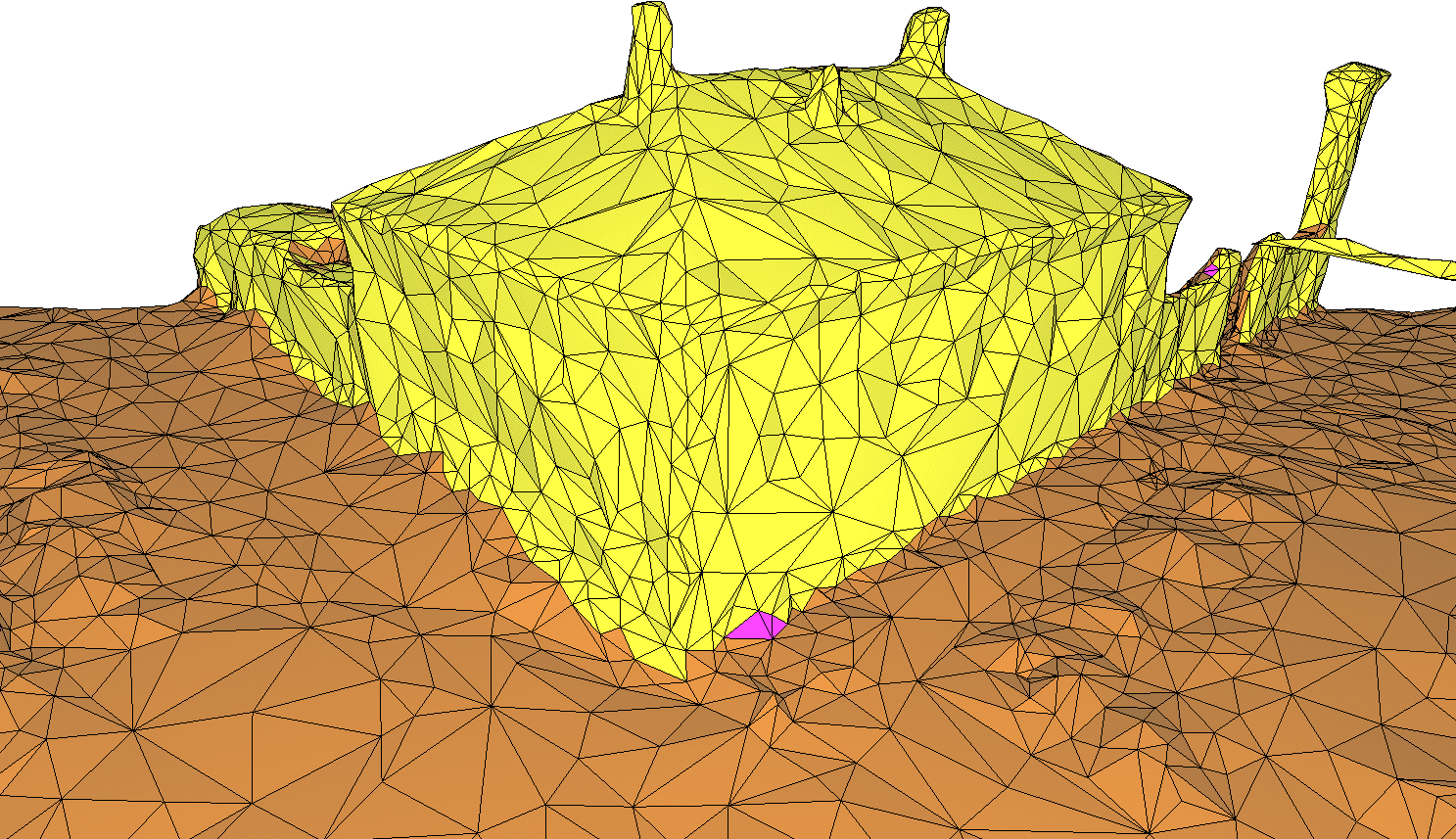}
		\caption{Near-uniform triangulation}
		\label{fig:asa}
	\end{subfigure}
	\hfill
	\begin{subfigure}{0.48\linewidth}		
		\includegraphics[width=\linewidth]{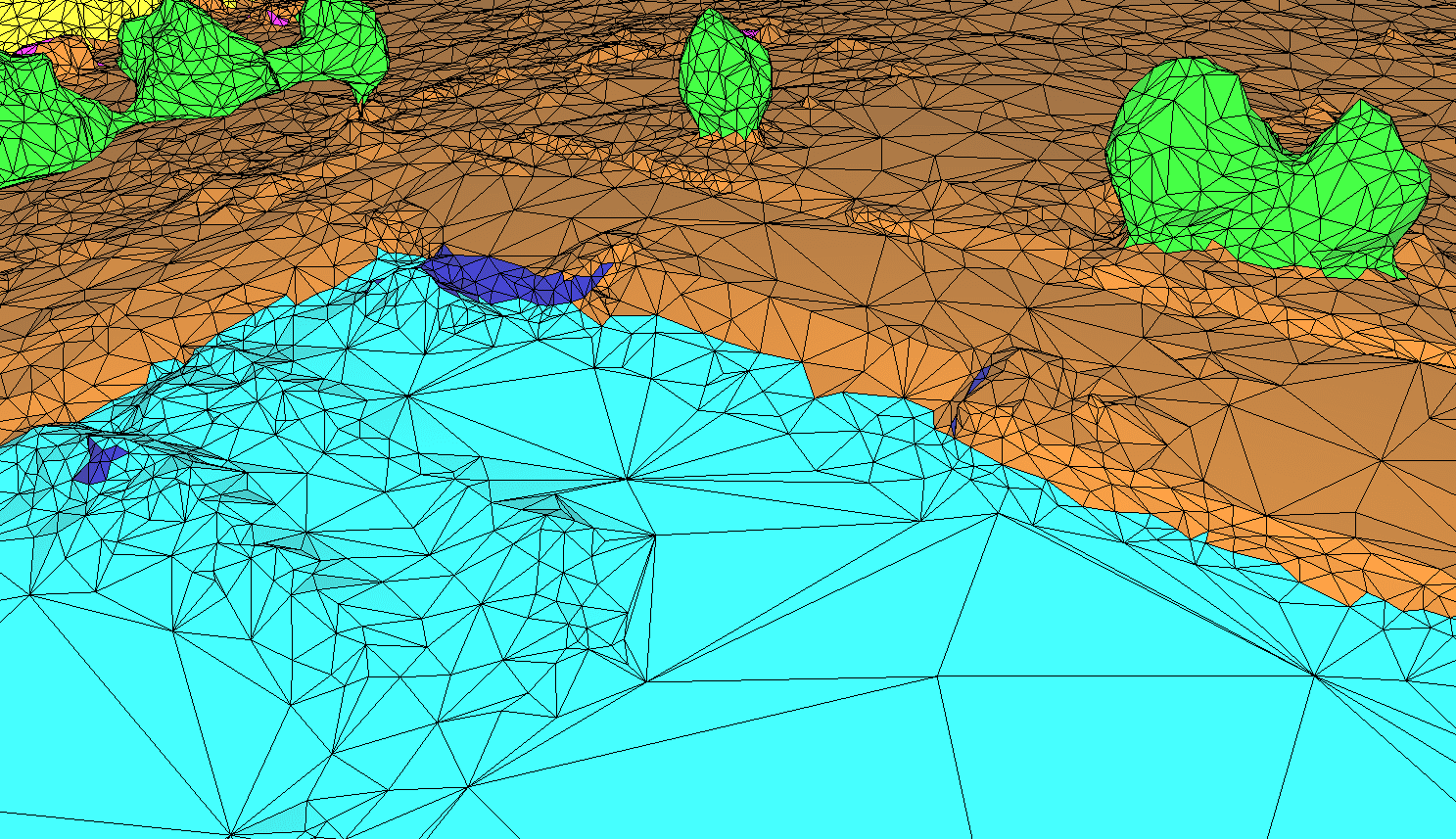}
		\caption{Non-uniform triangulation}
		\label{fig:br}
	\end{subfigure}
	\caption{Robustness against triangulation.}
	\label{fig:triangulation}
\end{figure}

\Cref{fig:semantic_qualitive1} shows the results for two tiles from the SUM dataset.
From the extensive experiments, we also observed that our method is robust against non-uniform triangulations of mesh, for which an example is shown in~\Cref{fig:triangulation}.
We have also compared our method with several state-of-the-art semantic segmentation approaches, among which RF-MRF~\cite{rouhani2017semantic} and SUM-RF~\cite{gao2021sum} directly consume meshes. 
To compare with methods originally developed for semantic segmentation of point clouds, e.g., PointNet~\cite{qi2017pointnet}, PointNet++~\cite{qi2017pointnet++}, SPG~\cite{landrieu2018large}, KPConv~\cite{thomas2019kpconv}, and RandLA-Net~\cite{hu2020randla}, we sampled points from the meshes by following~\cite{gao2021sum}.
For each deep learning method designed for point clouds, we feed it with the colored point clouds densely sampled from the texture meshes. We tune the hyper-parameters starting with their default setting. 
For a fair comparison, we use the same weight for all the competing methods involved in the comparison.
Due to the randomness of deep learning, we ran each method ten times with the same settings to record its average performance. 
We report the results in~\Cref{tab:semanticcomp}, and we can see the results of the top three best methods in~\Cref{fig:semantic_qualitive2}.
Our method achieves the highest per-class IoU on the majority of object classes, and it outperforms all other methods in all overall metrics, with a margin from $4\%$ to $35.9\%$ in terms of mIoU.
Compared to KPConv and SPG, our method requires less training time, and our results are more stable (i.e., with smaller standard deviations).

\begin{figure*}[!th]
	\centering
	\begin{subfigure}[t]{0.49\linewidth}
		\includegraphics[width=\linewidth]{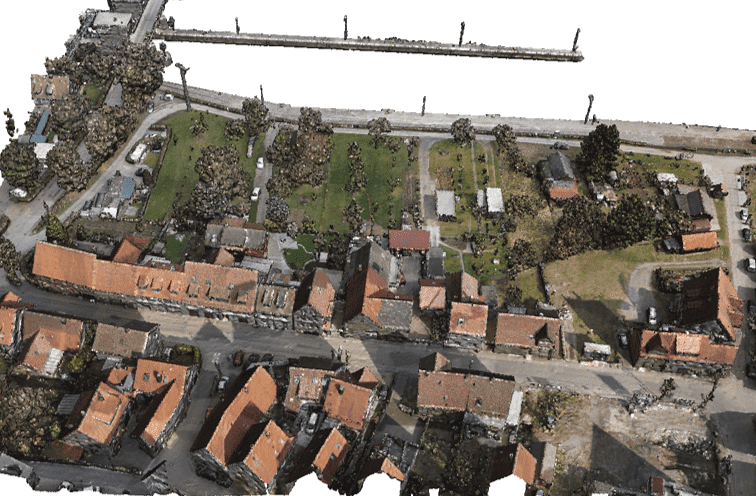}\vspace{0.1em}
		\caption{Input}
		\label{fig:tex_mesh_h3d}
	\end{subfigure}
	\hfill
	\begin{subfigure}[t]{0.49\linewidth}		
		\includegraphics[width=\linewidth]{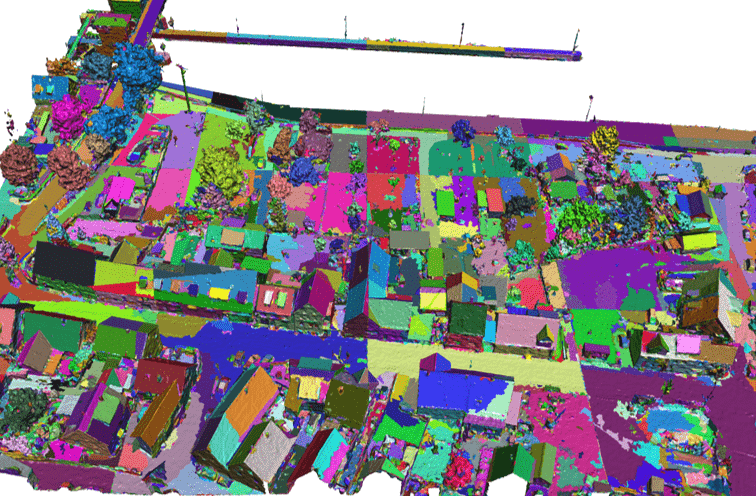}\vspace{0.1em}
		\caption{Segments}
		\label{fig:segments_h3d}
	\end{subfigure}
	\begin{subfigure}[t]{0.49\linewidth}
		\includegraphics[width=\linewidth]{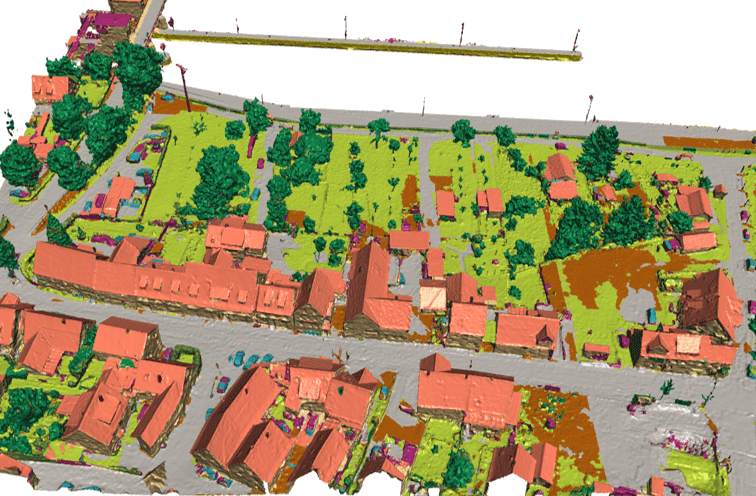}\vspace{0.1em}
		\caption{Semantic}
		\label{fig:predict_h3d}
	\end{subfigure}
	\hfill
	\begin{subfigure}[t]{0.40\linewidth}
	\hspace{-0.8cm}
	\includegraphics[width=\linewidth]{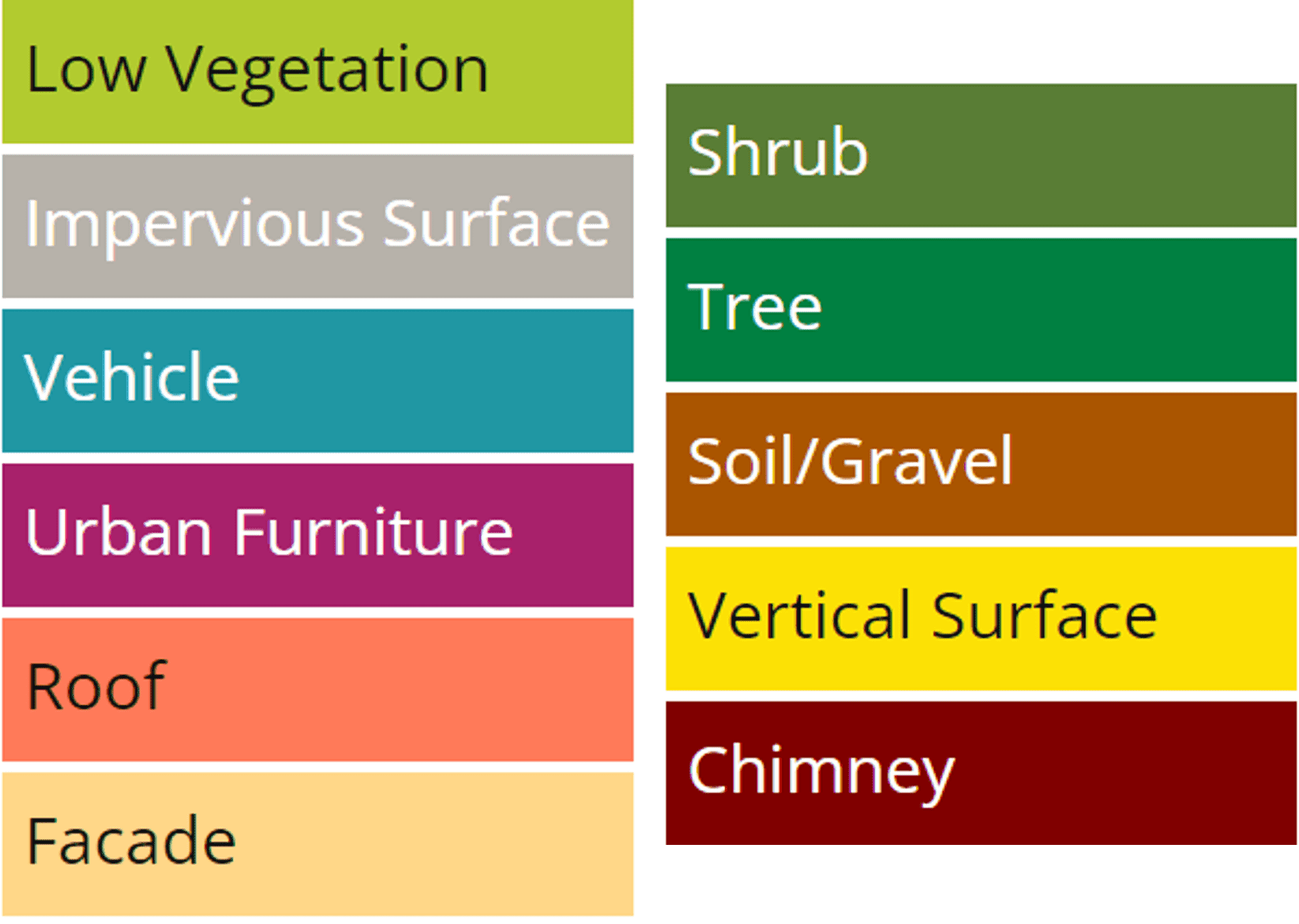}\vspace{0.1em}
	\end{subfigure}
	\caption{Semantic segmentation results of our method on the test area of the H3D dataset~\cite{kolle2021h3d}.}
	\label{fig:semantic_qualitive3}
\end{figure*}

\subsubsection{Evaluation on H3D}
We also attempted to train and test on the H3D dataset~\cite{kolle2021h3d} (see for an overview of the results in~\Cref{fig:semantic_qualitive3}).
It should be noted that the H3D dataset is much smaller (0.19 $km^2$) than SUM (4 $km^2$). 
In addition, 40\% of the area of the mesh in H3D is unlabeled and many triangles have incorrect labels, which was due to the limitation in their cloud-to-mesh labeling process where the correspondence was either not completed or has ambiguities when transferring the labels from the points to the mesh faces.
This prevents H3D from being an ideal training dataset for us, as both our method and other deep learning methods require a large amount of labeled data.
Besides, to distinguish between different classes that are geometrically coplanar in H3D (e.g., Low Vegetation, Impervious Surface, Soil, and Gravel), the \textit{planar} class is divided into sub-classes and used as a prior for over-segmentation in our approach. 
The test results show that our method achieves about 52.2\% mIoU, outperforming the KPConv (45.5\% mIoU), SPG (29.9\% mIoU), and PointNet++(15.6\% mIoU).  

\subsection{Ablation study}

To understand the effect of several design choices made in the graph construction, and the contributions of the hand-crafted features and the learned features, we have conducted an ablation study on the SUM dataset~\cite{gao2021sum}. 

\Cref{tab:ablation} summarizes the ablation result of the graphs and the features.
The upper part of~\Cref{tab:ablation} reveals the impact of removing each type of graph edges (i.e., connections between segments) on the final semantic segmentation. 
It reveals that every type of graph edges contributes to the performance, and removing any of them results in a drop in mIoU in the range $[2.9\%, 4.3\%]$.   
This implies that both local and global interactions between segments provide useful information and play an important role in semantic segmentation.
To understand the effectiveness of the designed graph, we compared it to using a graph with random connections (with the same number of edges as in our designed graph).
Although these connections may overlap with the edges we have designed, the presence of randomness significantly reduces the capability of the network. 
This is because the random set of edges does not convey the effective features captured by our carefully designed edges.
Besides, the orthogonal edges were also tested (see~\Cref{tab:ablation}), from which we can see that they are less useful than the other types of edges. This is because the orthogonal relationship is more often incident to man-made structures within the same object (such as building parts) than between segments from different objects.

The lower part of~\Cref{tab:ablation} details the ablation analysis of different features. 
We have evaluated the importance of each feature by removing it from the experiment and recording the performance of the semantic segmentation.  
These experiments show that the combination of all the features outperforms all degraded features with a margin of mIoU from $2.4\%$ to $7.8\%$, which means every feature contributes to the performance.
It is also interesting to notice that the results are more stable without RGB information as input to PointNet for feature learning, which indicates the low quality (e.g., distortion, shadow) of mesh textures in our training datasets.

\begin{table}[!tb]
	\centering
	\noindent\adjustbox{max width=1.0\linewidth}
	{
		\begin{tabular}{lllcccc}
			\hline
			&& Model & OA (\%) & mAcc (\%) & mIoU (\%) & $\Delta$mIoU (\%) \\
			\hline
			\multirow{4}{*}{\textbf{Graph Edges}} && No \textit{parallelism} & 92.9 & 75.3 & 68.5 ± 1.9 & -4.3 \\
			&& No \textit{ExMAT} & 93.1 & 76.0 & 69.3 ± 1.3 & -3.5 \\
			&& No \textit{spatial-proximity} & 93.1 & 76.3 & 69.4 ± 1.9 & -3.4 \\						
			&& No \textit{connecting-ground} & 93.3 & 76.3 & 69.9 ± 1.6 & -2.9 \\
			&&Only \textit{random} & 92.5 & 74.9 & 67.8 ± 2.3 & -5.0 \\
			&& With \textit{orthogonal} & 93.0 & 75.3 & 68.9 ± 1.8 & -3.9 \\
			\hline
			\multirow{8}{*}{\textbf{Features}} && No \textit{All Handcrafted} & 89.7 & 75.9 & 65.0 ± 3.8 & -7.8\\
			&& No \textit{Offset} & 92.4 & 75.0 & 67.2 ± 1.5 & -5.6\\
			&& No \textit{PointNet} & 92.8 & 74.5 & 68.2 ± 1.3 & -4.6 \\
			&& No \textit{Eigen} & 93.0 & 75.3 & 68.3 ± 2.3 & -4.5 \\
			&& No \textit{Color} & 93.1 & 75.0 & 68.9 ± 1.4 & -3.9 \\
			&& No \textit{Density} & 93.1 & 76.4 & 69.4 ± 2.0 & -3.4 \\
			&& No \textit{Scale} & 93.0 & 76.1 & 69.7 ± 0.9 & -3.1 \\
			&& No \textit{Shape} & 93.2 & 75.6 & 69.7 ± 0.8 & -3.1 \\
			&& No \textit{RGB with PointNet} & 93.5 & 76.4 & 70.4 ± 0.4 & -2.4 \\
			\hline
			&& Ours & \textbf{93.8} & \textbf{79.2} & \textbf{72.8} ± 2.0 & - \\
			\hline
		\end{tabular}
	}
	\caption{Ablation study of graph edges and features on the SUM dataset~\cite{gao2021sum}.
	\textit{PointNet} denotes features learned from PointNet~\cite{qi2017pointnet} with XYZ and RGB as input, and `` No \textit{RGB with PointNet}" means RGB is not used as input.
	}
	\label{tab:ablation}
\end{table}

\subsection{Generalization ability}
\label{sec:generaliza_abi}
We have conducted experiments to test the generalization ability of our method and the competing methods.
Such tests can indicate the applicability of models trained by different methods on practical test datasets.
SUM~\cite{gao2021sum} and H3D~\cite{kolle2021h3d} are well qualified for generalization ability tests as they both represent urban scenes.
As the original classes of these two datasets do not match, we merged them into four common classes that are typical of urban scenarios for testing, i.e., terrain (including water, low vegetation, impervious surface, soil, and gravel), high-vegetation (including shrub and tree), building (including roof, facade, and chimney), and vehicle (including car and boat).
For all methods, we trained the model on the SUM dataset and perform the testing on the H3D dataset.
In particular, for training and validation, we have used the training and validation splits of the SUM dataset as they cover a larger area than H3D. For testing, we have used training and validation splits of the H3D dataset that has publicly available ground truth labels.
To compare the difference in results, we also tested the same model on the test area of SUM.

\begin{table*}[!th]
	\centering
	\noindent\adjustbox{max width=1.0\linewidth}
	{
		\begin{tabular}{lllrrrrr}
			\hline
			&& Methods & \multicolumn{1}{l}{Terra.} & \multicolumn{1}{l}{ H-veg.} & \multicolumn{1}{l}{Build.} & \multicolumn{1}{l}{Vehic.} & \multicolumn{1}{l}{mIoU} \\
			\hline
			\multirow{8}{*}{\textbf{SUM}} && PointNet~\cite{qi2017pointnet} & 66.8  & 13.8  & 65.7  & 0.0  & 36.6 \\
			&&PointNet++~\cite{qi2017pointnet++} & 77.7  & 76.7  & 86.3  & 1.3  & 60.5 \\
			&&SPG~\cite{landrieu2018large} & 86.0  & 73.9  & 88.5  & 13.9  & 65.6 \\
			&&RF-MRF~\cite{rouhani2017semantic} & 86.8  & 86.7  & 90.5  & 20.7  & 71.2 \\
			&&RandLaNet~\cite{hu2020randla} & 83.0  & 91.6  & 90.1 & 22.0  & 71.7  \\
			&&KPConv~\cite{thomas2019kpconv} & 89.4 & 84.7  & 91.5 & \textbf{34.1} & 75.0 \\
			&&SUM-RF~\cite{gao2021sum} & 88.0  & 90.2  & 92.3  & 30.4  & 75.2  \\
			&&Ours & \textbf{89.5 }& \textbf{92.0} & \textbf{93.9} & 33.0  & \textbf{77.1} \\
			\hline
			\multirow{8}{*}{\textbf{H3D}} && PointNet++~\cite{qi2017pointnet++} & 0.0  & 0.0  & 0.0  & 1.1  & 0.3 \\
			&& KPConv~\cite{thomas2019kpconv} & 0.0  & 0.0  & 0.0  & 1.1  & 0.3 \\
			&& SPG~\cite{landrieu2018large} & 0.0  & 0.0  & 18.3  & 0.0  & 4.6 \\
			&& RandLaNet~\cite{hu2020randla} & 0.0  & 0.0  & 20.7  & 0.0  & 5.2 \\
			&& PointNet~\cite{qi2017pointnet} & 56.9  & 24.1  & 0.0  & 0.0  & 20.3 \\
			&& RF-MRF~\cite{rouhani2017semantic} & 73.9  & 46.0  & 40.0  & 4.7  & 41.1 \\
			&& SUM-RF~\cite{gao2021sum} & 8\textbf{0.2} & 44.0  & 41.6  & 9.2  & 43.8 \\
			&& Ours & 74.2 & \textbf{66.2} & \textbf{44.5} & \textbf{13.4} & \textbf{49.6} \\
			\hline
		\end{tabular}
	}
	\caption{Generalization ability comparison. All methods are trained on four classes of the SUM dataset~\cite{gao2021sum}. The top eight rows show the scores on the testing area of the SUM dataset, while the bottom eight records show the testing results on the four classes H3D dataset.
		Per-class IoU (\%) and mean IoU (mIoU, \%) are reported here.}
	\label{tab:generalization_ability}
\end{table*}

As shown in the top eight rows of~\Cref{tab:generalization_ability}, the mIoU of all methods has improved compared to the previous six classes in the SUM test area because the task of classification has become relatively easy due to the reduction in the number of classes. 
The bottom eight records of~\Cref{tab:generalization_ability} demonstrate the generalization ability of different methods. 
We can see that our method outperforms all competing methods with a margin from $5.8\%$ to $49.3\%$ in terms of mIoU.
Except for our approach, almost all other deep learning-based methods (i.e. PointNet++~\cite{qi2017pointnet++}, SPG~\cite{landrieu2018large}, KPConv~\cite{thomas2019kpconv}, and RandLA-Net~\cite{hu2020randla}) failed to predict the classes in a new urban scene.
This is because these methods all learn global features by having a large receptive field, and these global features can lead to overfitting of the model to the training data and result in degradation of generalization ability.
In contrast, the global features of PointNet~\cite{qi2017pointnet} are based on aggregated local features and do not correspond to a larger receptive field, which does not have the overfitting problem.
In other words, training with only local features avoids the degradation of model generalization ability, which is better illustrated by RF-MRF~\cite{rouhani2017semantic} and SUM-RF~\cite{gao2021sum} as they only use features extracted on local segments.
Whereas our approach includes global features derived from the local features, our proposed graph is based on the spatial distribution of the object components in the urban scene, which facilitates the generalization of the global features.

From~\Cref{tab:ablation} and~\Cref{tab:generalization_ability}, we can conclude that compared with features learned by the neural network, the proposed handcrafted features and edges lead to better semantic segmentation results and contribute to a stronger generalization ability, especially for data with domain gaps.  
In particular, the domain gaps are attributed to the differences between training data and test data (i.e., different feature distributions), and the reasons for these differences can be grouped into two main categories: 1) same data acquisition and processing pipeline,  but covering different urban scenarios (e.g., from dense urban area to rural or forest area); 
2) covering the same urban scenarios but with different data acquisition (e.g., using different sensors or different parameters for data collection) and processing pipelines (e.g., using different approaches or parameter configurations for generating the 3D data). 
Nevertheless, the features learned by the existing neural network architectures cannot cope with such differences without adding new training samples from the test area. 
Our segment-based handcrafted features and edges capture the intrinsic characteristics of the object (e.g., the facade is usually perpendicular to the ground or the surface of the tree is undulating and non-planar) and can better cope with the variance in feature distribution.

\subsection{Limitations}
Our method is based on the observation that urban scenes consist of objects demonstrating both planar and non-planar regions, and that object boundaries lie in the connections between the planar and non-planar regions. 
In special cases where adjacent objects contain only non-planar regions (e.g., vehicles underneath trees), our method will not be able to differentiate them.  
In addition, our approach generates segments for semantic classification, which reduces memory consumption as well as the number of samples. 
Specifically, if the training data covers a small area (e.g., H3D~\cite{kolle2021h3d}) and only a few segments are generated after over-segmentation, the number of samples may not be sufficient to train a competent model.  
Possible solutions include data augmentation of segments and graph connections or adding more labeled data.
Besides, our method requires adjacency information of the mesh for incremental region growing.
Additional preprocessing is necessary for meshes that are non-manifold or contain duplicated vertices, as they destroy the topological adjacency information.
Potential solutions can be to split non-manifold vertices or to reconstruct adjacency information of duplicated vertices. 
In our experiments, the meshes in the SUM dataset~\cite{gao2021sum} are 2-manifold, while the meshes in the H3D dataset~\cite{kolle2021h3d} are not.

%% file: source/conclusion.tex
\section{Conclusion}\label{sec:Conclusion}

We have presented a two-stage supervised framework for semantic segmentation of large-scale urban meshes.
Our planarity-sensible over-segmentation algorithm favors generating segments largely aligned with object boundaries, closer to semantically meaningful objects, can deliver descriptive features, and can represent urban scenes with a smaller number of segments.
A thorough analysis reveals that our planarity-sensible over-segmentation plays a key role in achieving superior performance in semantic segmentation.
We have also shown that exploiting multi-scale contextual information better facilitates semantic segmentation.
Furthermore, we have demonstrated that our proposed approach achieves better generalization abilities in comparison with other methods, owing to the segment-based local features and unique connections in graphs.
Our proposed new metrics are effective for evaluating mesh over-segmentation methods dedicated to semantic segmentation. We believe the proposed metrics will further stimulate improving other over-segmentation techniques. 
In future work, we would like to extend our framework to part-level (e.g., dormers, balconies, roofs, and facades of buildings) urban mesh segmentation. In addition, we will also investigate how the semantics learned from multi-view images can be used for semantic segmentation of urban meshes.

%% file: PSSNet.bbl
\begin{thebibliography}{10}
\expandafter\ifx\csname url\endcsname\relax
  \def\url#1{\texttt{#1}}\fi
\expandafter\ifx\csname urlprefix\endcsname\relax\def\urlprefix{URL }\fi
\expandafter\ifx\csname href\endcsname\relax
  \def\href#1#2{#2} \def\path#1{#1}\fi

\bibitem{Helsinki3d}
C.~of~Helsinki, Helsinki's {3D} city models,
  \url{https://www.hel.fi/helsinki/en/administration/information/general/3d},
  accessed: 2020-11-25 (Dec. 2019).

\bibitem{Google3d}
Google, {3D} imagery in google earth, \url{https://earth.google.com/web/},
  accessed: 2021-01-16 (Dec. 2012).

\bibitem{gao2021sum}
W.~Gao, L.~Nan, B.~Boom, H.~Ledoux, {SUM}: A {B}enchmark {D}ataset of
  {S}emantic {U}rban {M}eshes, ISPRS Journal of Photogrammetry and Remote
  Sensing 179 (2021) 108--120.
\newblock \href {http://dx.doi.org/10.1016/j.isprsjprs.2021.07.008}
  {\path{doi:10.1016/j.isprsjprs.2021.07.008}}.

\bibitem{biljecki2015applications}
F.~Biljecki, J.~Stoter, H.~Ledoux, S.~Zlatanova, A.~{\c{C}}{\"o}ltekin,
  Applications of {3D} city models: State of the art review, ISPRS
  International Journal of Geo-Information 4~(4) (2015) 2842--2889.

\bibitem{saran2015citygml}
S.~Saran, P.~Wate, S.~Srivastav, Y.~Krishna~Murthy, City{GML} at semantic level
  for urban energy conservation strategies, Annals of GIS 21~(1) (2015) 27--41.

\bibitem{besuievsky2018skyline}
G.~Besuievsky, B.~Beckers, G.~Patow, Skyline-based geometric simplification for
  urban solar analysis, Graphical Models 95 (2018) 42--50.

\bibitem{demantke2011dimensionality}
J.~Demantk{\'e}, C.~Mallet, N.~David, B.~Vallet, Dimensionality based scale
  selection in {3D} {LiDAR} point clouds, 2011.

\bibitem{hackel2016fast}
T.~Hackel, J.~D. Wegner, K.~Schindler, Fast semantic segmentation of {3D} point
  clouds with strongly varying density, ISPRS annals of the photogrammetry,
  remote sensing and spatial information sciences 3 (2016) 177--184.

\bibitem{qi2017pointnet}
C.~R. Qi, H.~Su, K.~Mo, L.~J. Guibas, Point{N}et: Deep learning on point sets
  for {3D} classification and segmentation, in: Proceedings of the IEEE
  conference on computer vision and pattern recognition, 2017, pp. 652--660.

\bibitem{qi2017pointnet++}
C.~R. Qi, L.~Yi, H.~Su, L.~J. Guibas, Point{N}et++: Deep hierarchical feature
  learning on point sets in a metric space, Advances in neural information
  processing systems 30 (2017) 5099--5108.

\bibitem{thomas2018semantic}
H.~Thomas, F.~Goulette, J.-E. Deschaud, B.~Marcotegui, Semantic classification
  of {3D} point clouds with multiscale spherical neighborhoods, in: 2018
  International Conference on {3D} Vision ({3DV}), IEEE, 2018, pp. 390--398.

\bibitem{thomas2019kpconv}
H.~Thomas, C.~R. Qi, J.-E. Deschaud, B.~Marcotegui, F.~Goulette, L.~J. Guibas,
  K{PC}onv: Flexible and {D}eformable {C}onvolution for {P}oint {C}louds, in:
  Proceedings of the IEEE International Conference on Computer Vision, 2019,
  pp. 6411--6420.

\bibitem{hanocka2019meshcnn}
R.~Hanocka, A.~Hertz, N.~Fish, R.~Giryes, S.~Fleishman, D.~Cohen-Or, Meshcnn: a
  network with an edge, ACM Transactions on Graphics (TOG) 38~(4) (2019) 1--12.

\bibitem{gao2019sdm}
L.~Gao, J.~Yang, T.~Wu, Y.-J. Yuan, H.~Fu, Y.-K. Lai, H.~Zhang, Sdm-net: Deep
  generative network for structured deformable mesh, ACM Transactions on
  Graphics (TOG) 38~(6) (2019) 1--15.

\bibitem{selvaraju2021buildingnet}
P.~Selvaraju, M.~Nabail, M.~Loizou, M.~Maslioukova, M.~Averkiou, A.~Andreou,
  S.~Chaudhuri, E.~Kalogerakis, Buildingnet: Learning to label 3d buildings,
  in: Proceedings of the IEEE/CVF International Conference on Computer Vision,
  2021, pp. 10397--10407.

\bibitem{fu20213d}
H.~Fu, R.~Jia, L.~Gao, M.~Gong, B.~Zhao, S.~Maybank, D.~Tao, 3d-future: 3d
  furniture shape with texture, International Journal of Computer Vision (2021)
  1--25.

\bibitem{verdie2015lod}
Y.~Verdie, F.~Lafarge, P.~Alliez, {LOD} {G}eneration for {U}rban {S}cenes,
  {ACM} Transactions on Graphics 34~(3) (2015) 1--14.
\newblock \href {http://dx.doi.org/10.1145/2732527}
  {\path{doi:10.1145/2732527}}.

\bibitem{rouhani2017semantic}
M.~Rouhani, F.~Lafarge, P.~Alliez, Semantic segmentation of {3D} textured
  meshes for urban scene analysis, ISPRS Journal of Photogrammetry and Remote
  Sensing 123 (2017) 124--139.

\bibitem{weinmann2013feature}
M.~Weinmann, B.~Jutzi, C.~Mallet, Feature relevance assessment for the semantic
  interpretation of {3D} point cloud data, ISPRS Annals of the Photogrammetry,
  Remote Sensing and Spatial Information Sciences 5~(W2) (2013) 1.

\bibitem{weinmann2015contextual}
M.~Weinmann, A.~Schmidt, C.~Mallet, S.~Hinz, F.~Rottensteiner, B.~Jutzi,
  Contextual classification of point cloud data by exploiting individual {3D}
  neigbourhoods, ISPRS Annals of the Photogrammetry, Remote Sensing and Spatial
  Information Sciences II-3 (2015), Nr. W4 2~(W4) (2015) 271--278.

\bibitem{huang2019texturenet}
J.~Huang, H.~Zhang, L.~Yi, T.~Funkhouser, M.~Nie{\ss}ner, L.~J. Guibas,
  Texture{N}et: Consistent local parametrizations for learning from
  high-resolution signals on meshes, in: Proceedings of the IEEE Conference on
  Computer Vision and Pattern Recognition, 2019, pp. 4440--4449.

\bibitem{li2019cross}
S.~Li, Z.~Luo, M.~Zhen, Y.~Yao, T.~Shen, T.~Fang, L.~Quan, Cross-atlas
  {C}onvolution for {P}arameterization {I}nvariant {L}earning on {T}extured
  {M}esh {S}urface, in: Proceedings of the IEEE Conference on Computer Vision
  and Pattern Recognition, 2019, pp. 6143--6152.

\bibitem{schult2020dualconvmesh}
J.~Schult, F.~Engelmann, T.~Kontogianni, B.~Leibe, Dual{C}onv{M}esh-{N}et:
  Joint geodesic and euclidean convolutions on {3D} meshes, in: Proceedings of
  the IEEE/CVF Conference on Computer Vision and Pattern Recognition, 2020, pp.
  8612--8622.

\bibitem{lin2018toward}
Y.~Lin, C.~Wang, D.~Zhai, W.~Li, J.~Li, Toward better boundary preserved
  supervoxel segmentation for {3D} point clouds, ISPRS journal of
  photogrammetry and remote sensing 143 (2018) 39--47.

\bibitem{landrieu2018large}
L.~Landrieu, M.~Simonovsky, Large-scale point cloud semantic segmentation with
  superpoint graphs, in: Proceedings of the IEEE Conference on Computer Vision
  and Pattern Recognition, 2018, pp. 4558--4567.

\bibitem{cohen2004variational}
D.~Cohen-Steiner, P.~Alliez, M.~Desbrun, Variational shape approximation, in:
  ACM SIGGRAPH 2004 Papers, 2004, pp. 905--914.

\bibitem{lafarge2012creating}
F.~Lafarge, C.~Mallet, Creating large-scale city models from {3D}-point clouds:
  a robust approach with hybrid representation, International journal of
  computer vision 99~(1) (2012) 69--85.

\bibitem{landrieu2019point}
L.~Landrieu, M.~Boussaha, Point cloud oversegmentation with graph-structured
  deep metric learning, in: Proceedings of the IEEE Conference on Computer
  Vision and Pattern Recognition, 2019, pp. 7440--7449.

\bibitem{Hui_2021_ICCV}
L.~Hui, J.~Yuan, M.~Cheng, J.~Xie, X.~Zhang, J.~Yang, Superpoint {N}etwork for
  {P}oint {C}loud {O}versegmentation, in: Proceedings of the IEEE/CVF
  International Conference on Computer Vision (ICCV), 2021, pp. 5510--5519.

\bibitem{cordts2016cityscapes}
M.~Cordts, M.~Omran, S.~Ramos, T.~Rehfeld, M.~Enzweiler, R.~Benenson,
  U.~Franke, S.~Roth, B.~Schiele, The cityscapes dataset for semantic urban
  scene understanding, in: Proceedings of the IEEE conference on computer
  vision and pattern recognition, 2016, pp. 3213--3223.

\bibitem{yang2018denseaspp}
M.~Yang, K.~Yu, C.~Zhang, Z.~Li, K.~Yang, Dense{ASPP} for {S}emantic
  {S}egmentation in {S}treet {S}cenes, in: Proceedings of the IEEE conference
  on computer vision and pattern recognition, 2018, pp. 3684--3692.

\bibitem{nan2012search}
L.~Nan, K.~Xie, A.~Sharf, A search-classify approach for cluttered indoor scene
  understanding, ACM Transactions on Graphics (TOG) 31~(6) (2012) 1--10.

\bibitem{liu2011entropy}
M.-Y. Liu, O.~Tuzel, S.~Ramalingam, R.~Chellappa, Entropy rate superpixel
  segmentation, in: CVPR 2011, IEEE, 2011, pp. 2097--2104.

\bibitem{vosselman2004recognising}
G.~Vosselman, B.~G. Gorte, G.~Sithole, T.~Rabbani, Recognising structure in
  laser scanner point clouds, International archives of photogrammetry, remote
  sensing and spatial information sciences 46~(8) (2004) 33--38.

\bibitem{schnabel2007efficient}
R.~Schnabel, R.~Wahl, R.~Klein, Efficient {RANSAC} for point-cloud shape
  detection, in: Computer graphics forum, Vol.~26, Wiley Online Library, 2007,
  pp. 214--226.

\bibitem{ben2018graph}
Y.~Ben-Shabat, T.~Avraham, M.~Lindenbaum, A.~Fischer, Graph based
  over-segmentation methods for {3D} point clouds, Computer Vision and Image
  Understanding 174 (2018) 12--23.

\bibitem{melzer2007non}
T.~Melzer, Non-parametric segmentation of {ALS} point clouds using mean shift,
  Journal of Applied Geodesy Jag 1~(3) (2007) 159--170.

\bibitem{papon2013voxel}
J.~Papon, A.~Abramov, M.~Schoeler, F.~Worgotter, Voxel cloud connectivity
  segmentation-supervoxels for point clouds, in: Proceedings of the IEEE
  conference on computer vision and pattern recognition, 2013, pp. 2027--2034.

\bibitem{vosselman2017contextual}
G.~Vosselman, M.~Coenen, F.~Rottensteiner, Contextual segment-based
  classification of airborne laser scanner data, ISPRS journal of
  photogrammetry and remote sensing 128 (2017) 354--371.

\bibitem{hu2020randla}
Q.~Hu, B.~Yang, L.~Xie, S.~Rosa, Y.~Guo, Z.~Wang, N.~Trigoni, A.~Markham,
  Rand{LA}-{N}et: Efficient semantic segmentation of large-scale point clouds,
  in: Proceedings of the IEEE/CVF Conference on Computer Vision and Pattern
  Recognition, 2020, pp. 11108--11117.

\bibitem{lei2021picasso}
H.~Lei, N.~Akhtar, A.~Mian, Picasso: A {CUDA}-based {L}ibrary for {D}eep
  {L}earning over {3D} {M}eshes, in: Proceedings of the IEEE/CVF Conference on
  Computer Vision and Pattern Recognition, 2021, pp. 13854--13864.

\bibitem{wu2021scenegraphfusion}
S.-C. Wu, J.~Wald, K.~Tateno, N.~Navab, F.~Tombari, Scene{G}raph{F}usion:
  Incremental {3D} {S}cene {G}raph {P}rediction from {RGB-D} {S}equences, in:
  Proceedings of the IEEE/CVF Conference on Computer Vision and Pattern
  Recognition, 2021, pp. 7515--7525.

\bibitem{guo2020deep}
Y.~Guo, H.~Wang, Q.~Hu, H.~Liu, L.~Liu, M.~Bennamoun, Deep learning for 3d
  point clouds: A survey, IEEE transactions on pattern analysis and machine
  intelligence 43~(12) (2020) 4338--4364.

\bibitem{ma20123d}
J.~Ma, S.~W. Bae, S.~Choi, {3D} medial axis point approximation using nearest
  neighbors and the normal field, The Visual Computer 28~(1) (2012) 7--19.

\bibitem{peters2016robust}
R.~Peters, H.~Ledoux, Robust approximation of the {M}edial {A}xis {T}ransform
  of {LiDAR} point clouds as a tool for visualisation, Computers \& Geosciences
  90 (2016) 123--133.

\bibitem{geurts2006extremely}
P.~Geurts, D.~Ernst, L.~Wehenkel, Extremely randomized trees, Machine learning
  63~(1) (2006) 3--42.

\bibitem{boykov2001fast}
Y.~Boykov, O.~Veksler, R.~Zabih, Fast approximate energy minimization via graph
  cuts, {IEEE} Transactions on Pattern Analysis and Machine Intelligence
  23~(11) (2001) 1222--1239.
\newblock \href {http://dx.doi.org/10.1109/34.969114}
  {\path{doi:10.1109/34.969114}}.

\bibitem{pearson1901liii}
K.~P. F.R.S., {LIII}. on lines and planes of closest fit to systems of points
  in space, Philosophical Magazine Series 1 2~(11) (1901) 559--572.

\bibitem{jaromczyk1992relative}
J.~Jaromczyk, G.~Toussaint, Relative neighborhood graphs and their relatives,
  Proceedings of the IEEE 80~(9) (1992) 1502--1517.

\bibitem{li2015gated}
Y.~Li, D.~Tarlow, M.~Brockschmidt, R.~Zemel, Gated graph sequence neural
  networks, in: Proceedings of the International Conference on Learning
  Representations, 2016.

\bibitem{cho2014learning}
K.~Cho, B.~{van Merrienboer}, C.~Gulcehre, F.~Bougares, H.~Schwenk, Y.~Bengio,
  Learning phrase representations using {RNN} encoder-decoder for statistical
  machine translation, in: Conference on Empirical Methods in Natural Language
  Processing (EMNLP 2014), 2014.

\bibitem{nair2010rectified}
V.~Nair, G.~E. Hinton, Rectified linear units improve restricted boltzmann
  machines, in: Icml, 2010.

\bibitem{ioffe2015batch}
S.~Ioffe, C.~Szegedy, Batch normalization: Accelerating deep network training
  by reducing internal covariate shift, in: International conference on machine
  learning, PMLR, 2015, pp. 448--456.

\bibitem{simonovsky2017dynamic}
M.~Simonovsky, N.~Komodakis, Dynamic edge-conditioned filters in convolutional
  neural networks on graphs, in: Proceedings of the IEEE conference on computer
  vision and pattern recognition, 2017, pp. 3693--3702.

\bibitem{szegedy2016rethinking}
C.~Szegedy, V.~Vanhoucke, S.~Ioffe, J.~Shlens, Z.~Wojna, Rethinking the
  inception architecture for computer vision, in: Proceedings of the IEEE
  conference on computer vision and pattern recognition, 2016, pp. 2818--2826.

\bibitem{cgal:eb-20b}
{The CGAL Project}, {CGAL} User and Reference Manual, {5.1.1} Edition, {CGAL
  Editorial Board}, 2020.

\bibitem{easy3d2021}
L.~Nan, Easy3{D}: a lightweight, easy-to-use, and efficient {C}++ library for
  processing and rendering 3{D} data, Journal of Open Source Software 6~(64)
  (2021) 3255.
\newblock \href {http://dx.doi.org/10.21105/joss.03255}
  {\path{doi:10.21105/joss.03255}}.

\bibitem{pytorch2019}
A.~Paszke, S.~Gross, F.~Massa, A.~Lerer, J.~Bradbury, G.~Chanan, T.~Killeen,
  Z.~Lin, N.~Gimelshein, L.~Antiga, A.~Desmaison, A.~Kopf, E.~Yang, Z.~DeVito,
  M.~Raison, A.~Tejani, S.~Chilamkurthy, B.~Steiner, L.~Fang, J.~Bai,
  S.~Chintala, Py{T}orch: An {I}mperative {S}tyle, {H}igh-{P}erformance {D}eep
  {L}earning {L}ibrary, in: H.~Wallach, H.~Larochelle, A.~Beygelzimer,
  F.~d\textquotesingle Alch\'{e}-Buc, E.~Fox, R.~Garnett (Eds.), Advances in
  Neural Information Processing Systems 32, Curran Associates, Inc., 2019, pp.
  8024--8035.

\bibitem{kolle2021h3d}
M.~Kölle, D.~Laupheimer, S.~Schmohl, N.~Haala, F.~Rottensteiner, J.~D. Wegner,
  H.~Ledoux, The hessigheim 3d (h3d) benchmark on semantic segmentation of
  high-resolution 3d point clouds and textured meshes from uav lidar and
  multi-view-stereo, ISPRS Open Journal of Photogrammetry and Remote Sensing 1
  (2021) 100001.
\newblock \href
  {http://dx.doi.org/https://doi.org/10.1016/j.ophoto.2021.100001}
  {\path{doi:https://doi.org/10.1016/j.ophoto.2021.100001}}.

\end{thebibliography}
